%% file: main.tex
\documentclass[11pt]{article}
\usepackage[fontsize=11.4pt]{fontsize}

\usepackage{hyperref}       
\usepackage{url}            
\usepackage{booktabs}       
\usepackage{amsfonts}       
\usepackage{nicefrac}       
\usepackage{microtype}      
\usepackage{xcolor}
\usepackage{graphicx}
\usepackage{subcaption}
\usepackage{caption}
\usepackage{hyphenat}
\usepackage{setspace}
\usepackage{amsmath}
\usepackage{amssymb}
\usepackage{mathtools}
\usepackage{amsthm}
\usepackage{longtable}
\usepackage{multirow}
\usepackage{float}
\usepackage{tabularx}
\usepackage{enumitem}
\usepackage{threeparttable}
\usepackage{algorithm}
\usepackage{algorithmic}
\usepackage{titletoc}
\usepackage{needspace}

\PassOptionsToPackage{numbers,sort&compress}{natbib}
\usepackage[preprint]{my_style}
\usepackage{fontspec}
\usepackage{unicode-math}

\usepackage{tcolorbox}
\tcbuselibrary{skins, breakable, listings}

\definecolor{brandblue}{RGB}{57,95,207}
\definecolor{linkblue}{HTML}{0064E0}
\definecolor{textgray}{HTML}{1C2B33}
\definecolor{boxbg}{HTML}{F1F4F7}
\definecolor{softblue}{HTML}{EDF4FF}
\definecolor{softgreen}{HTML}{EEF8F1}
\definecolor{softyellow}{HTML}{FFF7E8}
\definecolor{borderblue}{HTML}{B7CEF8}
\definecolor{bordergreen}{HTML}{B8DFC1}
\definecolor{borderyellow}{HTML}{F0D49A}
\definecolor{softgray}{HTML}{F7F8FA}
\definecolor{bordergray}{HTML}{D7DEE8}
\definecolor{cardtitleblue}{HTML}{DCE9FF}
\definecolor{cardbodyblue}{HTML}{F3F7FF}
\definecolor{cardbodyblueend}{HTML}{EAF2FF}
\definecolor{cardtitleblueend}{HTML}{CFE0FF}
\definecolor{cardaccentblue}{HTML}{1F6BFF}
\definecolor{levelonebg}{HTML}{43A047}
\definecolor{leveltwobg}{HTML}{2F6BFF}
\definecolor{levelthreebg}{HTML}{D34A4A}
\definecolor{levelonetext}{HTML}{2E7D32}
\definecolor{leveltwotext}{HTML}{1F5FD6}
\definecolor{levelthreetext}{HTML}{B4232A}

\hypersetup{
  colorlinks=true,
  linkcolor=brandblue,
  citecolor=brandblue,
  urlcolor=linkblue
}

\newcommand{\paperTitle}{This is a Paper Title}
\renewcommand{\paperTitle}{Quo Vadis, World Modeling?\\
{\Large Towards Interactive World Proxies for Continually Improving Agents}}

\newcommand{\paperAuthors}{KnowledgeX Lab @ Shanghai AI Laboratory\\[0.08cm]APRIL Lab @ Zhejiang University\\[0.08cm]LV-Lab @ National University of Singapore}

\newcommand{\publishDate}{\today}

\newcommand{\projectLink}{\href{https://worldbench.github.io/awesome-agentic-world-model}{~\texttt{https://worldbench.github.io/awesome-agentic-world-model}}}
\newcommand{\githubLink}{\href{https://github.com/worldbench/awesome-agentic-world-model}{~\texttt{https://github.com/worldbench/awesome-agentic-world-model}}}

\setlist[itemize]{leftmargin=*, itemsep=2pt, topsep=3pt, parsep=1pt}
\setlist[enumerate]{leftmargin=*, itemsep=2pt, topsep=3pt, parsep=1pt}
\renewcommand{\arraystretch}{1.14}
\titlecontents{section}
  [0.45em]
  {\normalsize\sffamily\bfseries\vspace{9pt}}
  {\textcolor{brandblue}{\contentslabel{1.65em}}}
  {}
  {\titlerule*[0.45pc]{.}\contentspage}
\titlecontents{subsection}
  [2.1em]
  {\normalsize\sffamily\vspace{4pt}}
  {\textcolor{textgray}{\contentslabel{2.35em}}}
  {}
  {\titlerule*[0.45pc]{.}\contentspage}

\newtcolorbox{blogbox}[2][]{
  enhanced,
  breakable,
  colback=#2,
  colframe=borderblue,
  boxrule=0.6pt,
  arc=7pt,
  left=10pt,
  right=10pt,
  top=6pt,
  bottom=6pt,
  before skip=6pt,
  after skip=6pt,
  #1
}

\newenvironment{definitionbox}{%
  \begin{blogbox}[colframe=borderblue,borderline west={2pt}{0pt}{brandblue}]{softblue}
}{%
  \end{blogbox}
}

\newenvironment{questionbox}{%
  \begin{blogbox}[colframe=borderyellow,borderline west={2pt}{0pt}{borderyellow},left=12pt,right=12pt,top=7pt,bottom=7pt,before upper={\let\texttt\questiontext\color{textgray}}]{softyellow}
}{%
  \end{blogbox}
}

\newenvironment{keypointbox}{%
  \begin{blogbox}[colframe=bordergray,borderline west={2pt}{0pt}{brandblue}]{softgray}
}{%
  \end{blogbox}
}

\newtcolorbox{highlightcard}[1]{
  enhanced,
  breakable,
  colback=cardbodyblue,
  colframe=borderblue,
  interior style={top color=cardbodyblue,bottom color=cardbodyblueend},
  boxrule=0.75pt,
  arc=6pt,
  left=10pt,
  right=10pt,
  top=7pt,
  bottom=7pt,
  before skip=5pt,
  after skip=5pt,
  before upper={\let\texttt\cardtext\color{textgray}},
  title={#1},
  coltitle=cardaccentblue,
  fonttitle=\sffamily\bfseries\large,
  colbacktitle=cardtitleblue,
  title style={left color=cardtitleblue,right color=cardtitleblueend},
  titlerule=0pt,
  toptitle=3pt,
  bottomtitle=3pt,
  lefttitle=10pt,
  righttitle=10pt
}

\newcommand{\WP}{\mathcal{WP}}
\newcommand{\blocktitle}[1]{{\sffamily\bfseries\textcolor{brandblue}{#1}}\par\vspace{2pt}}
\newcommand{\localheading}[1]{\vspace{4pt}\noindent{\sffamily\bfseries\textcolor{brandblue}{#1}}\par\vspace{2pt}}
\newcommand{\cardicon}[1]{\hspace{-2pt}\raisebox{-0.18em}{\includegraphics[height=1.25em]{#1}}\hspace{5pt}}
\newcommand{\cardstrong}[1]{\textbf{\textcolor{cardaccentblue}{#1}}}
\newcommand{\questiontext}[1]{{\rmfamily\bfseries\itshape\textcolor{textgray}{#1}}}
\newcommand{\questionkey}[1]{\textcolor{cardaccentblue}{\textbf{#1}}}
\newcommand{\cardtext}[1]{{\rmfamily\small\linespread{1.08}\selectfont\textcolor{textgray}{#1}}}
\newcommand{\cardkey}[1]{\textcolor{cardaccentblue}{\textbf{#1}}}
\newcommand{\levelbadge}[2]{%
  {\raisebox{0.9pt}{\colorbox{#2}{\scriptsize{\textbf{\textsc{\textcolor{white}{#1}}}}}}}%
}
\newcommand{\LoneBadge}{\levelbadge{L.1}{levelonebg}}
\newcommand{\LtwoBadge}{\levelbadge{L.2}{leveltwobg}}
\newcommand{\LthreeBadge}{\levelbadge{L.3}{levelthreebg}}
\newcommand{\LoneText}[1]{\textcolor{levelonetext}{\textbf{#1}}}
\newcommand{\LtwoText}[1]{\textcolor{leveltwotext}{\textbf{#1}}}
\newcommand{\LthreeText}[1]{\textcolor{levelthreetext}{\textbf{#1}}}
\newcommand{\LoneInline}{\LoneText{L1}}
\newcommand{\LtwoInline}{\LtwoText{L2}}
\newcommand{\LthreeInline}{\LthreeText{L3}}

\newcommand{\renderFrontBox}{%
    \tcbset{
    enhanced, frame hidden,
    colback=boxbg,
    left=0.5cm, right=0.5cm, top=0.42cm, bottom=0.48cm,
    arc=16pt,
    before skip=0pt,
    grow to left by=1.5pt, grow to right by=1.5pt,
    overlay={
    \node[anchor=north east, at=(frame.north east), xshift=-0.5cm, yshift=-0.42cm]
        {\includegraphics[height=0.90cm]{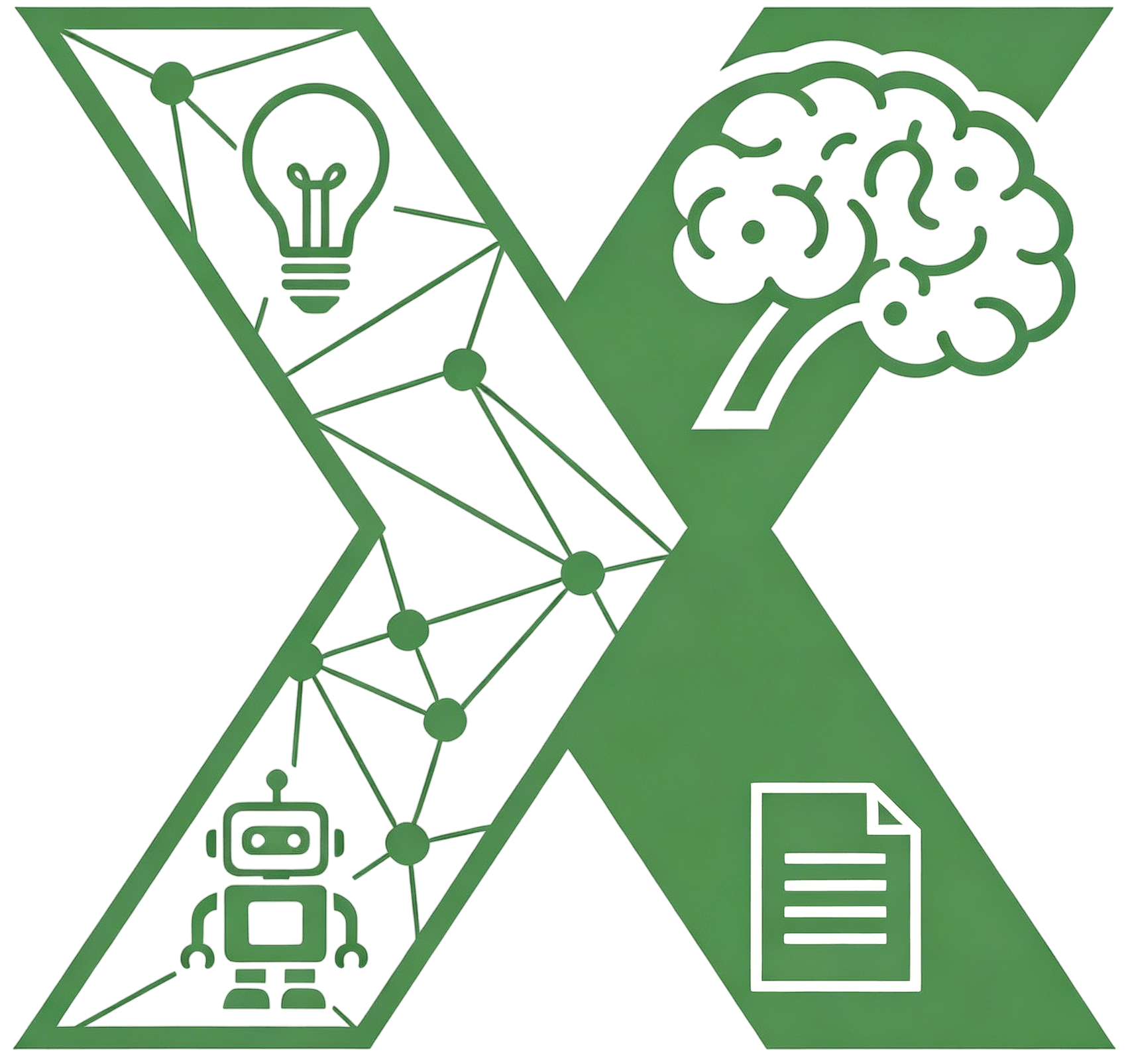}\hspace{0.18cm}%
         \includegraphics[height=0.90cm]{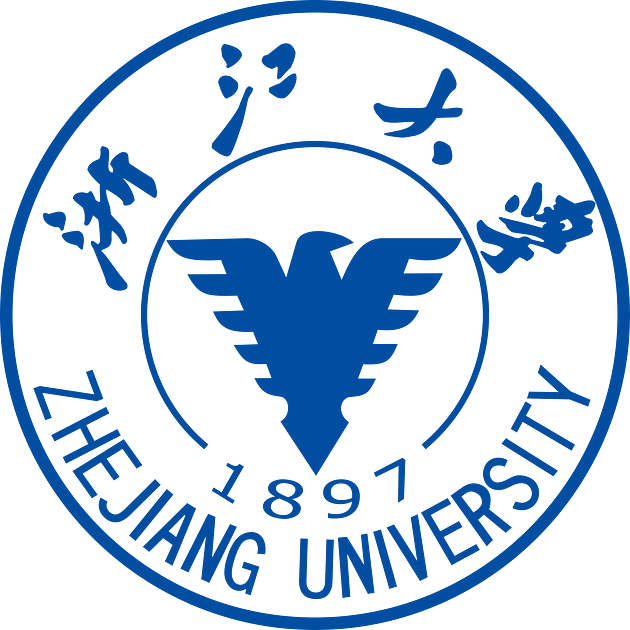}\hspace{0.18cm}%
         \includegraphics[height=0.90cm]{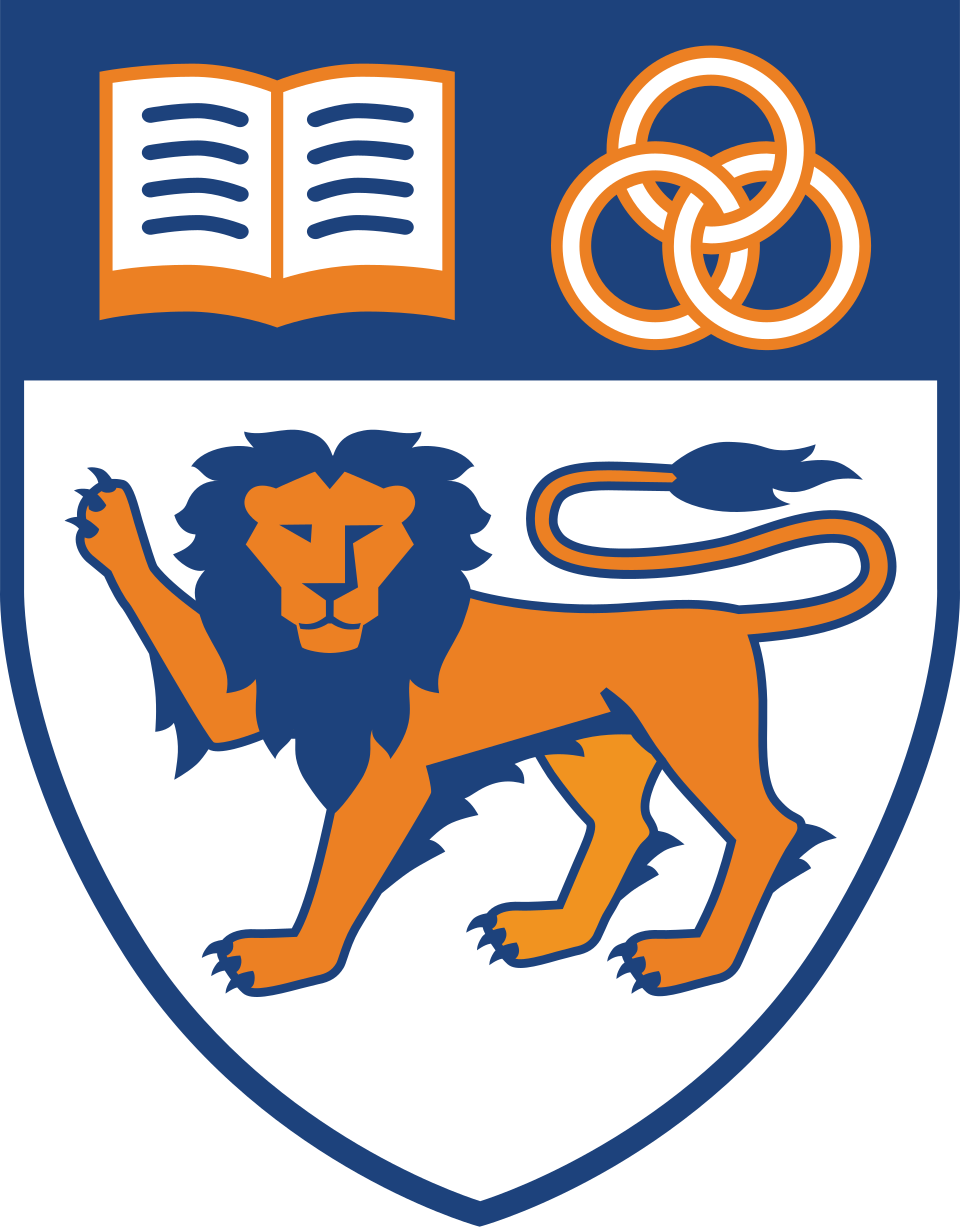}\hspace{0.18cm}%
         \includegraphics[height=0.90cm]{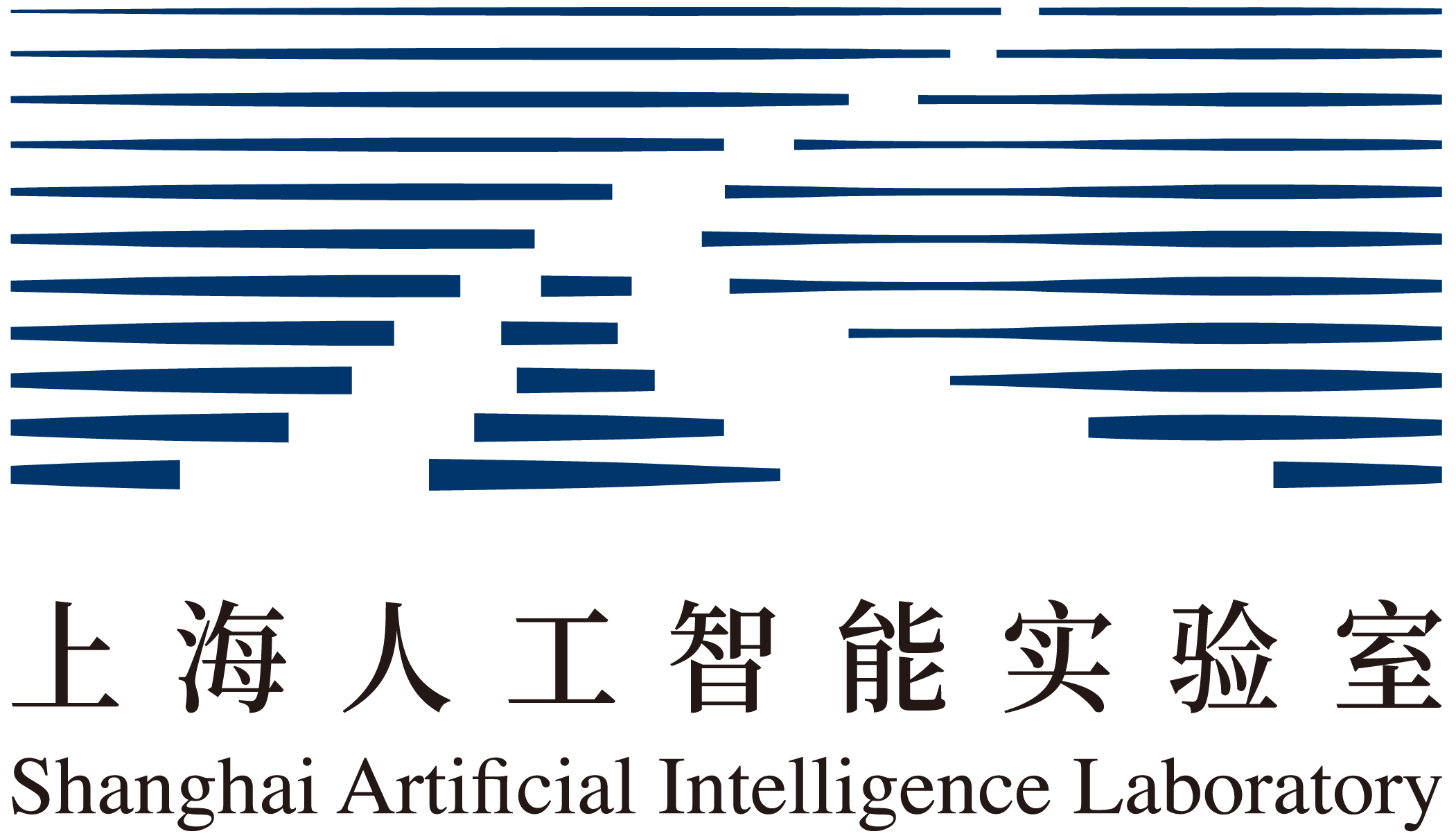}};
    }
  }%
  \begin{tcolorbox}
    \setlength{\parindent}{0cm}
    \setlength{\parskip}{0.5cm}
    {
      \setlength{\parskip}{0cm}
      \raggedright
      \nohyphens
      {
        \vskip 0.34cm
        \setstretch{1.5}
        {\huge\sffamily\bfseries\textcolor{black}{\paperTitle}}\par
      }
      \vskip 0.22cm
      \paperAuthors\par
    }
    \vskip -0.22cm
    {\color{textgray}\fontsize{10.2}{12.5}\selectfont%
    \input{sections/abstract}
\par}
    \vskip 0.34cm
    {
      \setlength{\parskip}{0cm}
      {\normalsize {\sffamily\bfseries \raisebox{-0.2em}{\includegraphics[width=0.025\linewidth]{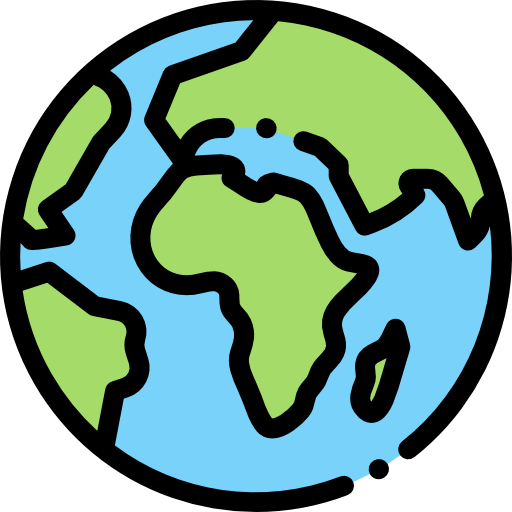}}~~Project Page:} \projectLink}\par%
      \vskip 0.2cm%
      {\normalsize {\sffamily\bfseries \raisebox{-0.2em}{\includegraphics[width=0.025\linewidth]{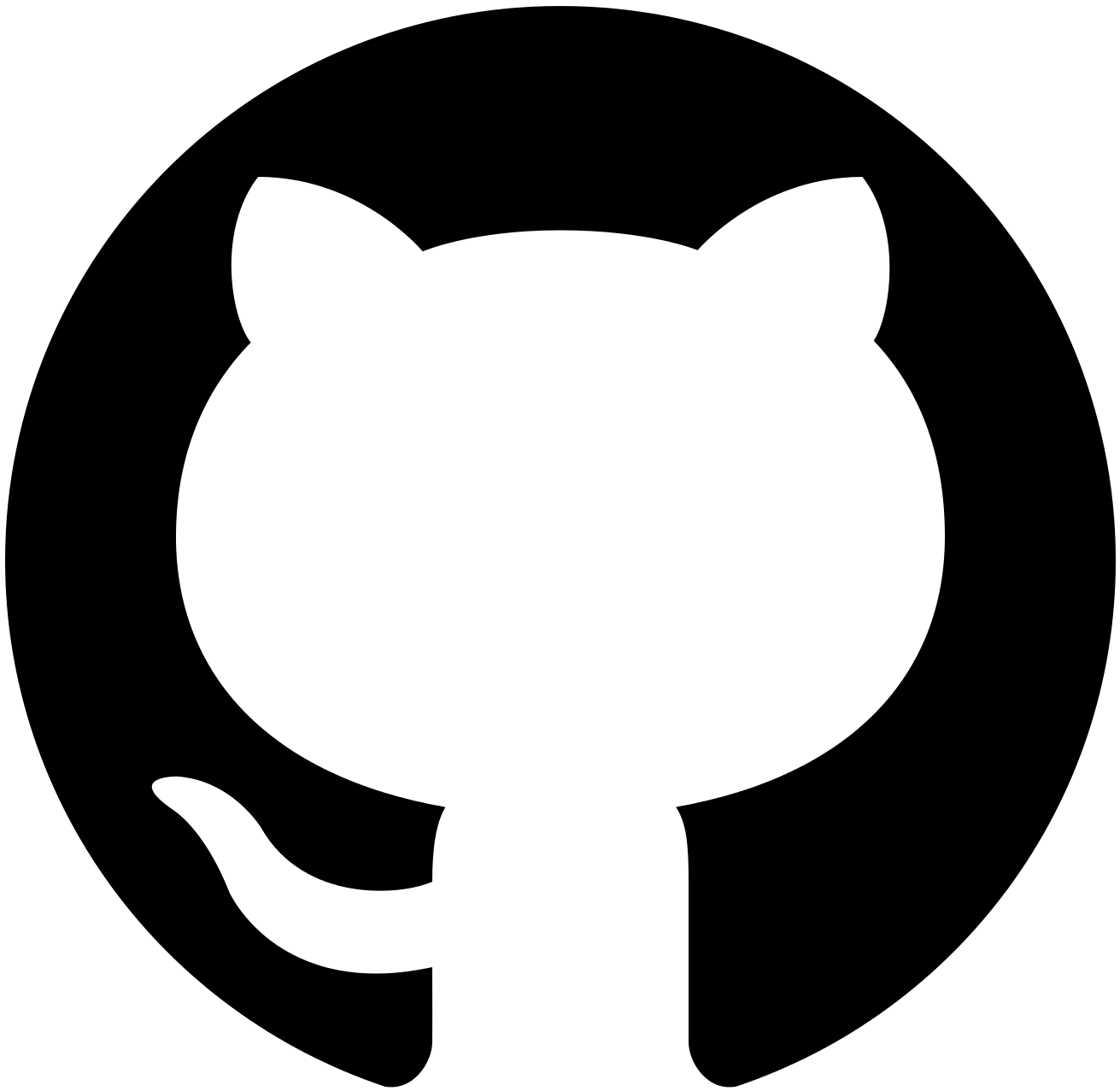}}~~GitHub Repo:} \githubLink}\par%
      \vskip 0.2cm%
      {\normalsize {\sffamily\bfseries \raisebox{-0.2em}{\includegraphics[width=0.025\linewidth]{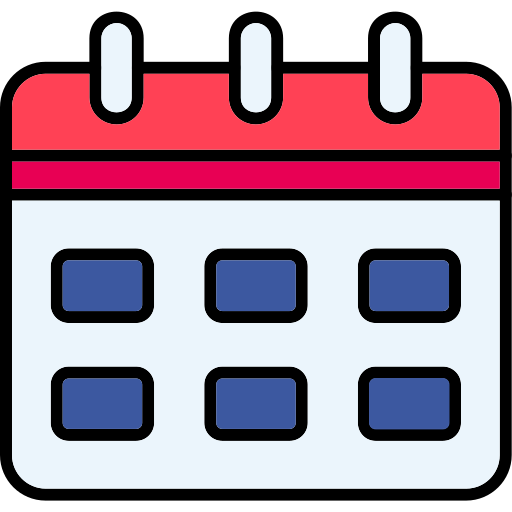}}~~Date:} \publishDate}\par%
    }
  \end{tcolorbox}
  \tcbset{reset}
}

\newcommand{\renderTeaserFigure}{%
  \vspace{-0.18cm}
  \begin{center}
    \captionsetup{skip=0pt,font=small}
    \includegraphics[width=0.999\linewidth]{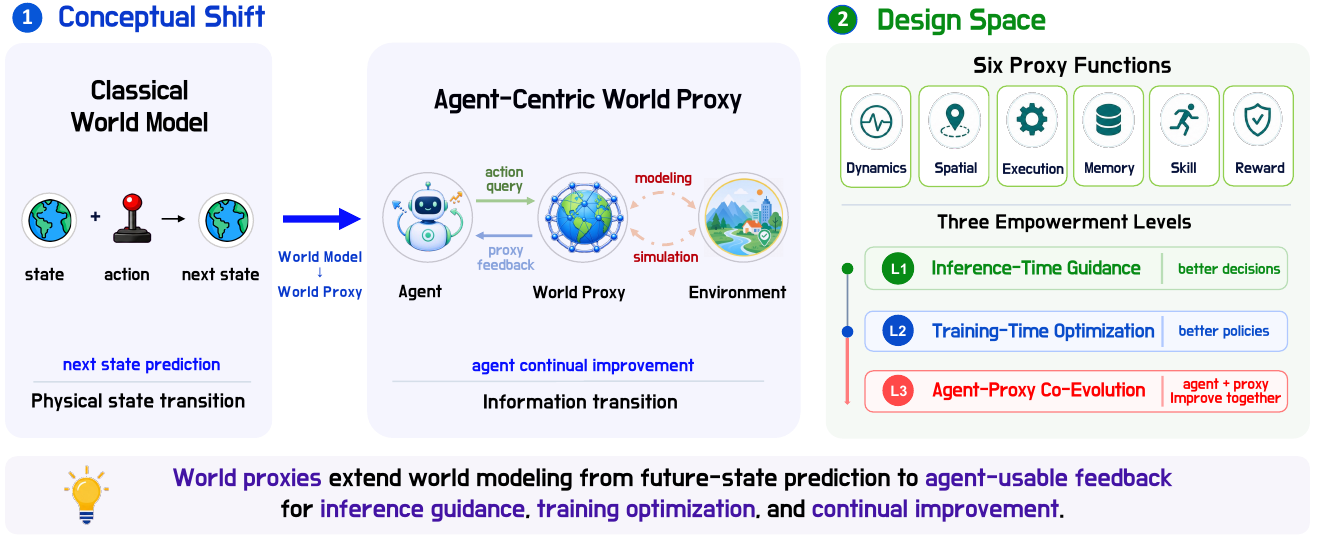}\par
    \vspace{-0.2em}
    \begin{minipage}{0.95\linewidth}
    \captionof{figure}{\textbf{Conceptual shift and design space of Agent-Centric World Proxies.} We shift from world models for physical state prediction to interactive world proxies for information transition prediction, enabling continual agent improvement through six proxy functions and three empowerment levels: \LoneBadge~inference-time guidance, \LtwoBadge~training-time optimization, and \LthreeBadge~Agent-Proxy co-evolution.}
      \label{fig:teaser}
    \end{minipage}
  \end{center}
}

\newcommand{\renderContents}{%
  \vspace{0.25cm}
  \begin{tcolorbox}[
    enhanced,
    breakable,
    frame hidden,
    colback=boxbg,
    left=0.55cm,
    right=0.45cm,
    top=0.25cm,
    bottom=0.2cm,
    arc=12pt,
    before skip=0pt,
    after skip=0pt,
    grow to left by=1.5pt,
    grow to right by=1.5pt
  ]
    \begingroup
    \normalsize
    \setstretch{1.08}
    \setlength{\parskip}{0pt}
    \setlength{\parindent}{0pt}
    \renewcommand{\contentsname}{Contents}
    \tableofcontents
    \endgroup
  \end{tcolorbox}
}

\begin{document}

\newgeometry{top=0.75in,bottom=0.5in,textwidth=6.3in,footskip=20pt}
\thispagestyle{empty}
\renderFrontBox
\renderTeaserFigure
\clearpage
\thispagestyle{empty}
\renderContents
\clearpage

\newgeometry{
  textheight=9in,
  textwidth=6.3in,
  top=1in,
  headheight=12pt,
  headsep=25pt,
  footskip=30pt
}

\input{sections/motivation}
\input{sections/definition}
\input{sections/empowerment}
\input{sections/instantiations}

\input{sections/conclusion}
\input{sections/contributors}

\clearpage\clearpage
\providecommand{\bibfont}{}
\renewcommand{\bibfont}{\fontsize{9.5}{11.4}\selectfont}
\bibliographystyle{plainnat}
\bibliography{preprint}

\end{document}

%% file: sections/abstract.tex
Continually improving agents require dynamic interaction feedback beyond static supervision, yet direct real-environment interaction is costly, slow, unsafe, and hard to parallelize. World modeling offers a natural intermediate proxy that allows agents to query lower-cost, more controllable feedback before committing to real actions. Classical world models instantiate this proxy primarily through future physical-state prediction, a formulation useful yet narrow for agents that require actionable feedback beyond raw state transitions. In this work, we conceptualize \textbf{Agent-Centric Interactive World Proxies}, shifting the fundamental paradigm from physical state transitions to \textbf{agent-usable information transitions}, such as execution outcomes, retrieved experiences or skills, and verification signals, broadening the scope of world modeling to provide versatile feedback for continually improving agents. To systematically map this design space, we organize world proxies into six functional forms based on their feedback modalities: \textit{dynamics}, \textit{spatial}, \textit{execution}, \textit{memory/experience}, \textit{skill}, and \textit{reward/verification} proxies, which together characterize the primary ways world modeling serves agent improvement. We further analyze how these proxies empower agents across three progressive levels: \LoneBadge~\LoneText{Inference-Time Guidance}, where proxy outputs enrich in-context information for superior decisions; \LtwoBadge~\LtwoText{Training-Time Optimization}, where proxy outputs yield rewards, critiques, or synthetic rollouts for policy learning; and \LthreeBadge~\LthreeText{Agent-Proxy Co-Evolution}, where real-environment evidence continuously updates both the proxy and the agent for co-evolution. Ultimately, this work recasts world modeling into an agent-centric paradigm, establishing a roadmap for building world proxies that empower agents to plan better, learn faster, and evolve continually.

%% file: sections/motivation.tex
\clearpage
\section{Motivation: Why Improving Agents Need World Modeling}
\label{sec:motivation}
\enlargethispage{1.0\baselineskip}

\subsection{From Static Supervision to Continual Improvement}

A truly capable agent does more than finish the task in front of it. It explores the unfamiliar, draws feedback from the world, and turns every success and failure into momentum for the next attempt~\cite{zhang2025earlyexp,wang2023voyager}. Capability of this kind is not granted once at training time; it is earned, again and again, through interaction.

Yet most agents today are still taught the way students cram for an exam: from static, offline data such as expert trajectories, human annotations, or supervised fine-tuning corpora~\cite{ouyang2022instructgpt,bai2022hh}. Such data captures patterns that already exist, but it cannot answer back when the agent strikes out on its own. The agent's competence is therefore \textbf{bounded by the distribution it was trained on}, and the \textbf{new information} that only active trial and error can reveal stays out of reach~\cite{vandeven2024continuallearning}. This is the ceiling that genuine self-improvement must break through.

\vspace{0.2cm}
\begin{questionbox}
\questiontext{How can an agent move beyond \questionkey{static supervision}, gather \questionkey{useful feedback} through \questionkey{active interaction}, and use it to \questionkey{improve without end}?}
\end{questionbox}
\vspace{0.2cm}

The most direct answer is to set the agent loose in the real environment and let it learn from whatever comes back:
\begingroup
\setlength{\belowdisplayskip}{3pt}
\[
\texttt{Agent}
~\rightarrow~
\texttt{Action}
~\rightarrow~
\texttt{Real Environment}
~\rightarrow~
\texttt{Feedback}
~\rightarrow~
\texttt{Agent Improvement}
\]
\endgroup

\begin{figure}[H]
  \centering
  \vspace{0.12cm}
  \captionsetup{skip=5pt,font=small}
  \includegraphics[width=0.66\columnwidth]{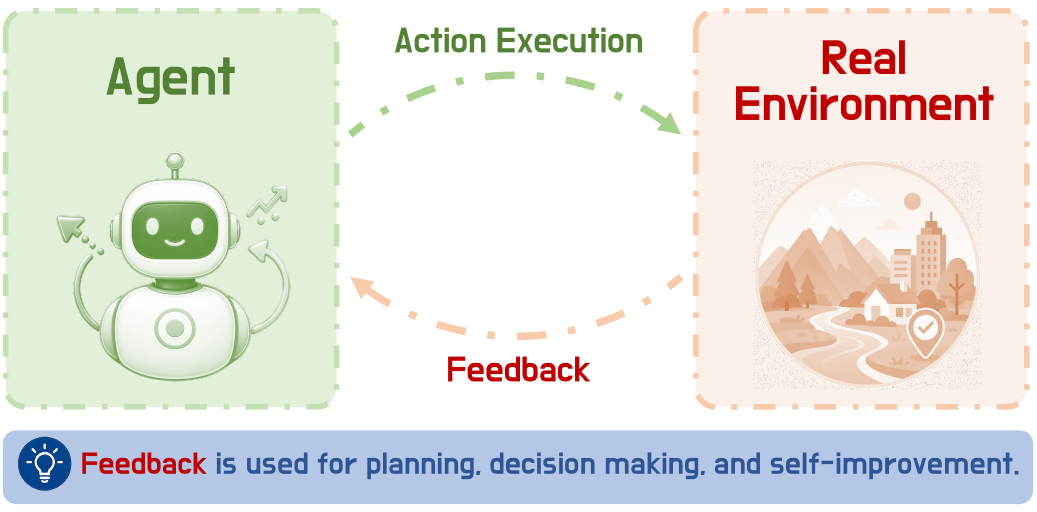}
  \captionof{figure}{\textbf{Basic Agent-Environment Interaction Loop.} The agent executes real actions and receives feedback for planning, learning, and improvement.}
  \label{fig:agent-env}
  \vspace{-0.08cm}
\end{figure}

The environment answers with observations, execution results, rewards, or errors, and the agent folds those signals back into its planning, decision making, policy learning, and continual improvement~\cite{sutton1991dyna,schrittwieser2020muzero,hafner2019dreamer}. It is a clean loop, and for a long time it was the whole story.

\subsection{Bottlenecks of Direct Real-Environment Interaction}

The real environment offers the most faithful feedback, and yet it makes a poor sole training ground once an agent needs to err thousands of times, branch across many possibilities at once, and reason several steps ahead before committing. The cracks show up along four recurring axes, collected in Table~\ref{tab:real-env-bottlenecks}: it is \textbf{costly, risky, stubbornly backward-looking, and hard to parallelize}.

In short, the real world is indispensable for grounding but ill-suited to scale. To improve efficiently, an agent needs something in between: a place to try, to predict, and to reason about what-ifs without paying the full price of real execution.

\begin{table}[t]
\centering
\vspace{-0.2cm}
\footnotesize
\setlength{\tabcolsep}{3.0pt}
\renewcommand{\arraystretch}{1.02}
\newcommand{\bottleneck}[1]{\textcolor{brandblue}{\textbf{#1}}}
\newcommand{\agentexamples}[2]{%
  \vspace{-0.85\baselineskip}
  \begin{itemize}[leftmargin=1.0em,itemsep=0pt,topsep=0pt,parsep=0pt,partopsep=0pt]
    \item \textbf{Embodied Agents:} #1
    \item \textbf{Digital Agents:} #2
  \end{itemize}
  \vspace{-0.95\baselineskip}
}
\captionsetup{skip=2pt}
\caption{Bottlenecks of direct real-environment interaction.}
\label{tab:real-env-bottlenecks}
\begin{tabularx}{\columnwidth}{
  >{\raggedright\arraybackslash}p{0.17\columnwidth}
  >{\raggedright\arraybackslash}p{0.36\columnwidth}
  >{\raggedright\arraybackslash}X
}
\toprule
\textbf{Bottleneck} & \textbf{Explanation} & \textbf{Examples}\\
\midrule
\bottleneck{Low Training Efficiency and High Cost}
& The agent needs a large amount of trial and error, while real interactions are usually expensive and slow.
& \agentexamples
  {robotic trial and error incurs device wear and safety costs~\cite{wu2023daydreamer}.}
  {web, GUI, or game agents consume computation and interaction resources during large-scale sampling~\cite{zhou2023webarena,xie2024osworld,liu2023agentbench,jimenez2023swebench}.}\\
\midrule
\bottleneck{Non-Rollbackable and Risky Interaction}
& Incorrect operations in the real environment are often difficult to undo and may even cause irreversible consequences~\cite{zeng2024wmsafety,li2026embodiedsafety}.
& \agentexamples
  {robots may collide with or damage objects, and autonomous vehicles may hit pedestrians.}
  {web agents may incorrectly submit, delete, or send information.}\\
\midrule
\bottleneck{Limited Forward-Looking Feedback}
& Real-environment feedback is usually \textbf{passive and after-the-fact}: it reveals only the outcome of an executed action, making future prediction or multi-step reasoning difficult~\cite{qian2026foresight}.
& \agentexamples
  {robots cannot easily know in advance whether an action will cause a collision.}
  {web or code agents often need to execute before knowing the result.}\\
\midrule
\bottleneck{Difficult to Parallelize}
& Real-environment instances are limited and difficult to replicate at scale like simulated environments~\cite{makoviychuk2021isaacgym,todorov2012mujoco,savva2019habitat,tao2024maniskill3,li2024behavior1k,nasiriany2024robocasa}.
& \agentexamples
  {physical robot platforms are limited in number.}
  {online services and user environments are difficult to replicate in batches.}\\
\bottomrule
\end{tabularx}
\end{table}

\subsection{World Modeling as an Intermediate Proxy}
\enlargethispage{0.5\baselineskip}

That something is a \textbf{proxy} that sits between the agent and the world~\cite{ha2018worldmodels,sutton1991dyna}:
\begingroup
\setlength{\belowdisplayskip}{3pt}
\[
\texttt{Agent}
~\leftrightarrow~
\texttt{World Modeling}
~\leftrightarrow~
\texttt{Real Environment}.
\]
\endgroup

\begin{figure}[H]
  \centering
  \vspace{0.15cm}
  \captionsetup{skip=5pt,font=small}
  \includegraphics[width=0.9\columnwidth]{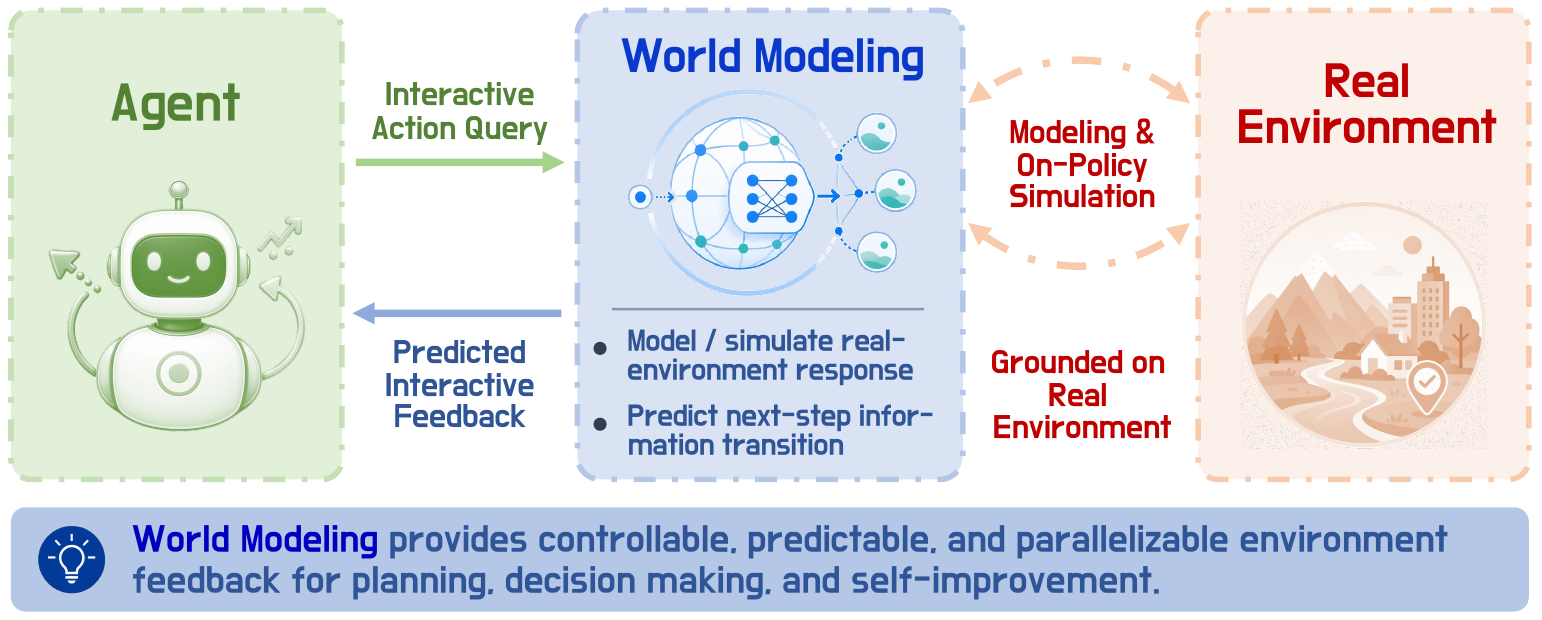}
  \captionof{figure}{\textbf{Agent-Centric World Modeling as an Intermediate Proxy.} World Modeling provides lower-cost feedback between the Agent and the Real Environment.}
  \label{fig:agent-wm-env}
  \vspace{-0.1cm}
\end{figure}

Crucially, World Modeling here is not a stand-in for reality, nor a contest to render the most photorealistic future. Its job is humbler and more useful: to act as an \textbf{intermediate proxy, grounded in the real environment}, that hands the agent interaction feedback in a \textbf{cheaper, more controllable, and more predictable} form.

Concretely, such a proxy lets the agent probe its options before committing to any of them. \emph{What might happen if I take this action? What would I see from another viewpoint? What would this command or tool call return? Have I faced something like this before, and is the plan safe to run?} Each is a question the proxy can answer cheaply, and each spares the agent a costly or irreversible move in the real world. Seen this way, World Modeling for continually improving agents is no longer confined to next-state prediction; it becomes a broader, agent-facing proxy, grounded in real-environment evidence, that can simulate, retrieve, guide, and verify.

\localheading{What makes a useful proxy?}

What separates such a proxy from a mere simulator or database is what it is optimized for. To genuinely serve a continually improving agent, Agent-Centric World Modeling should meet at least three requirements:

\begin{highlightcard}{\cardicon{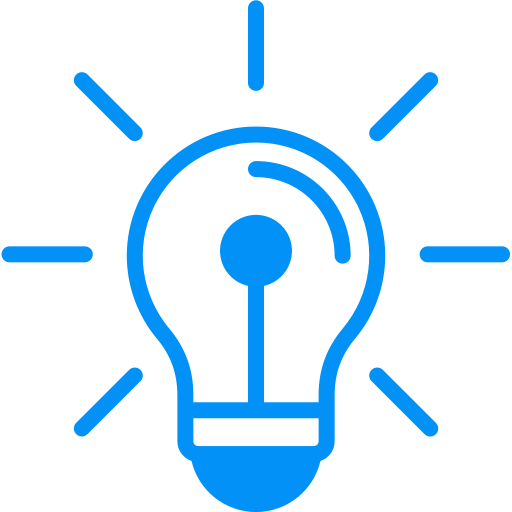} Key Requirements}
\begin{enumerate}[leftmargin=1.3em,itemsep=5pt,topsep=2pt,parsep=1pt,partopsep=0pt,label=\textcolor{brandblue}{\textbf{\arabic*.}}]
  \item \cardicon{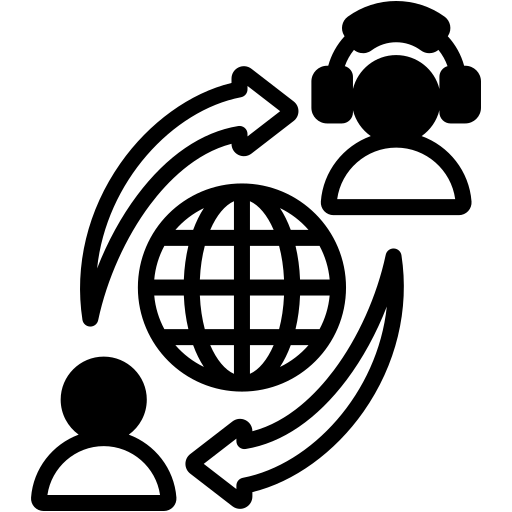}\cardstrong{Agent-Facing Closed Loop:}\\\cardtext{Support \cardkey{agent-initiated queries, actions, or interventions} and return conditioned feedback for decision making and improvement.}
  
  \item \cardicon{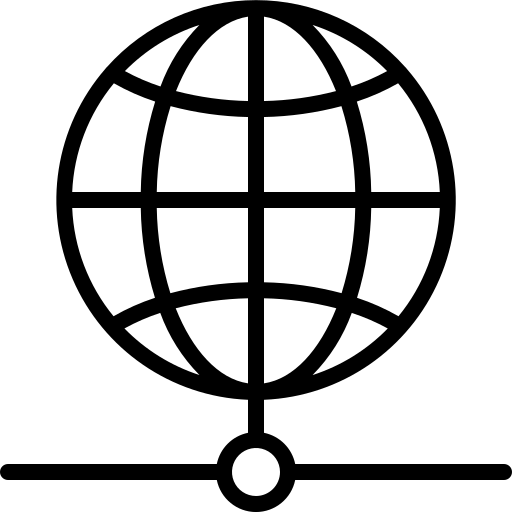}\cardstrong{Real-Environment Grounding:}\\\cardtext{Learn from \cardkey{real-environment} data, rules, trajectories, or interaction evidence, approximating the outcomes and feedback relevant to agent decisions.}
  
  \item \cardicon{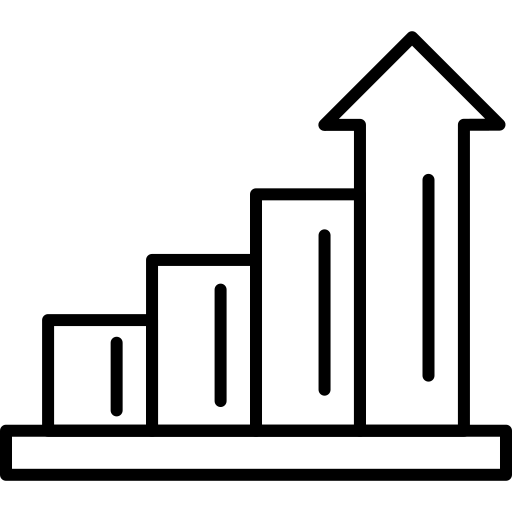}\cardstrong{Actionable Information Gain:}\\\cardtext{Optimize for \cardkey{better context}, \cardkey{safer decisions}, more \cardkey{effective exploration}, or \cardkey{higher-quality training signals}, not only visual realism or prediction accuracy.}
\end{enumerate}
\end{highlightcard}

%% file: sections/definition.tex
\vspace{0.3cm}
\section{Definition: From World Models to Agent-Centric World Proxies}
\label{sec:definition}

\subsection{Rethinking World Models}

To ask where world modeling should go, it helps to recall where it began. A classical World Model is usually defined as a state transition model~\cite{ha2018worldmodels,hafner2019dreamer,lecun2022path}:

\begingroup
\setlength{\abovedisplayskip}{3pt}
\setlength{\belowdisplayskip}{3pt}
\[
\hat{s}_{t+1}=\mathcal{WM}(s_t,a_t).
\]
\endgroup

That is, at physical time step $t$, the system is in state $s_t$, executes action $a_t$, and the World Model predicts the next physical state $\hat{s}_{t+1}$.

This form is suitable for robotics, model-based RL, video prediction, and related settings~\cite{moerland2023mbrl,wu2023daydreamer,brooks2024sora}. Its core focus is:
\begingroup
\setlength{\belowdisplayskip}{3pt}
\[
\texttt{Current State} + \texttt{Action}
\rightarrow
\texttt{Future State}.
\]
\endgroup

In other words, a traditional World Model traces a single thread of state transitions along physical time. Even modern self-supervised variants that predict in a learned latent space \cite{yu2026latent} rather than in raw pixels~\cite{assran2023ijepa,meta2025vjepa2} inherit this backbone: a state goes in, an action is applied, and a future comes out.

For a \textbf{continually improving agent}, however, the feedback worth having reaches well past the next state. Over the course of solving a task, such an agent may need to:

\begin{itemize}[leftmargin=1.35em,itemsep=1pt,topsep=2pt,parsep=0pt,partopsep=0pt,label=\textcolor{brandblue}{\textbullet}]
  \item \cardstrong{\texttt{simulate}} the result of an action, command, API call, or tool call~\cite{gu2024webdreamer,tang2024worldcoder,lu2024toolsandbox};
  \item \cardstrong{\texttt{retrieve}} relevant information from memory, experience, or failure cases~\cite{park2023generative,shinn2023reflexion,packer2023memgpt};
  \item \cardstrong{\texttt{query}} reusable skills or sub-policies for the current task~\cite{wang2023voyager,ahn2022saycan,liang2022codeaspolicies};
  \item \cardstrong{\texttt{verify}} whether a plan, trajectory, or action is safe and feasible~\cite{cobbe2021verifiers,lightman2023letsverify,mcaleese2024criticgpt};
  \item \cardstrong{\texttt{obtain}} reward, critique, preference, or error-diagnosis signals for training~\cite{christiano2017deeprlpreferences,ouyang2022instructgpt,rafailov2023dpo}.
\end{itemize}

What unites these is the \emph{kind} of answer they return. None is simply a next state; each is a different species of information, a consequence, a memory, a skill, a judgment, that the agent can immediately act on.

Thus the classical World Model is an indispensable starting point, but only a starting point: it captures one mechanism of Agent-World interaction, not the full repertoire a continually improving agent relies on.

\subsection{From World Model to World Proxy}

To embrace this wider set of interactions, we generalize the \textbf{World Model} into a \textbf{World Proxy}. The move is less a replacement than a broadening of the same core idea: an intermediate mechanism, grounded in the real environment, that supplies feedback the agent would otherwise have to win through direct execution. What changes is the reach. A World Proxy need not stop at predicting the next state; it may also simulate execution results, retrieve experience, offer skill guidance, or verify and evaluate behavior~\cite{gu2024webdreamer,park2023generative,wang2023voyager,lightman2023letsverify}.

\begin{figure}[H]
  \centering
  \vspace{0.35cm}
  \captionsetup{skip=5pt,font=small}
  \begin{minipage}{0.999\columnwidth}
    \centering
    \includegraphics[width=\linewidth]{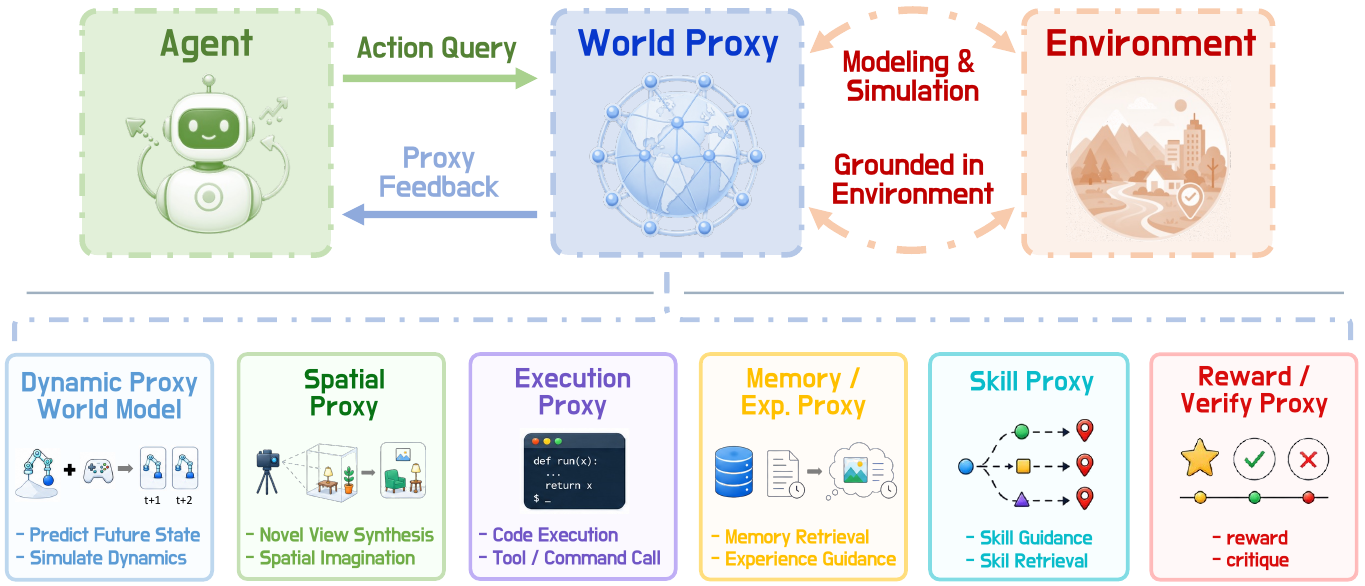}
    \caption{\textbf{From World Model to World Proxy.} World Proxy is an agent-facing proxy layer that may instantiate as dynamics prediction, spatial synthesis, execution simulation, memory retrieval, skill guidance, or reward / verification feedback.}
    \label{fig:world-proxy}
  \end{minipage}
  \vspace{-0.12cm}
\end{figure}

Concretely, this broader view stretches the classical definition along three axes, summarized in Table~\ref{tab:world-proxy-rethinking}:

\begin{table}[H]
\centering
\footnotesize
\setlength{\tabcolsep}{3.0pt}
\renewcommand{\arraystretch}{1.08}
\newcommand{\wprethink}[2]{{\scriptsize\textbf{\textcolor{brandblue}{#1}}}\newline\textbf{\textcolor{brandblue}{$\rightarrow$ #2}}}
\newcommand{\wptag}[1]{\fcolorbox{borderblue}{softblue}{\strut\hspace{2pt}\textcolor{brandblue}{\textbf{#1}}\hspace{2pt}}}
\newcommand{\wpplaintag}[1]{\fcolorbox{borderblue}{softblue}{\strut\hspace{2pt}\textcolor{brandblue}{#1}\hspace{2pt}}}
\captionsetup{skip=4pt}
\caption{From environment-centric prediction to agent-centric interaction modeling.}
\label{tab:world-proxy-rethinking}
\begin{tabularx}{\columnwidth}{
  >{\raggedright\arraybackslash}p{0.18\columnwidth}
  >{\raggedright\arraybackslash}p{0.13\columnwidth}
  >{\raggedright\arraybackslash}p{0.13\columnwidth}
  >{\raggedright\arraybackslash}X
}
\toprule
{\sffamily\bfseries\textcolor{brandblue}{Rethinking}}
& {\sffamily\bfseries\textcolor{brandblue}{From}}
& {\sffamily\bfseries\textcolor{brandblue}{To}}
& {\sffamily\bfseries\textcolor{brandblue}{Key Idea}}\\
\midrule
\wprethink{Physical Time Step}{Interaction Step}
& \wptag{$t \rightarrow t+1$}
& \wptag{$\ell \rightarrow \ell+1$}
& Agent interaction does not always advance physical time. It may query a viewpoint, simulate execution, retrieve experience, or verify a plan. We use \cardkey{interaction step $\ell$} to describe each Agent-World interaction.\\
\addlinespace[2pt]
\wprethink{State Transition}{Information Transition}
& \wptag{$s_t \rightarrow \hat{s}_{t+1}$}
& \wptag{$s_\ell \rightarrow \hat{s}_{\ell+1}$}
& The output of a World Proxy is not limited to next-state prediction. It can be viewed as an \cardkey{information transition}, such as a new observation, execution result, retrieved memory or skill, reward, verification feedback, or other agent-usable information.\\
\addlinespace[2pt]
\wprethink{External Condition}{Agent-Initiated Interaction}
& \wpplaintag{external}
& \wptag{$u_\ell^{\mathcal{F}}$}
& The interaction condition should be \cardkey{agent-initiated}: a query, action, or intervention actively proposed by the Agent according to its goal and context, rather than only an externally given condition.\\
\bottomrule
\end{tabularx}
\vspace{3pt}
\noindent
\begin{minipage}{\columnwidth}
\vspace{2pt}
\small\raggedright
\textit{Note.} $\mathcal{F}$ denotes different \textbf{proxy functions}, such as dynamics prediction, spatial rendering, execution simulation, memory retrieval, skill guidance, and reward / verification.
\end{minipage}
\end{table}

Therefore, the key question for an Agent-Centric World Proxy is not only whether the prediction is accurate, but:

\vspace{0.2cm}
\begin{questionbox}
\questiontext{What \questionkey{new information} will the agent obtain after this \questionkey{interaction}?}
\end{questionbox}
\vspace{0.2cm}

In this sense, the output of a \textbf{World Proxy} should be organized as \textbf{agent-usable feedback}, rather than only as an environmental state. The shift is subtle but freeing: a memory lookup, a code execution, and a reward estimate can all be treated as one move of the same game, each delivering information the agent did not possess an instant earlier~\cite{shinn2023reflexion,tang2024worldcoder,christiano2017deeprlpreferences}.

\subsection{Agent-Centric World Proxy Definition}

Pulling these threads together, we can now state the idea precisely. We define an Agent-Centric World Proxy as:

\vspace{0.2cm}
\begin{definitionbox}
{\rmfamily\small\linespread{1.08}\selectfont\raggedright
An Agent-Centric World Proxy is an \cardkey{environment-grounded proxy} that models or predicts \mbox{\cardkey{information transitions}} conditioned on \mbox{\cardkey{agent-initiated interactions}}, aiming to provide \cardkey{information gain} for agent improvement.\par}
\end{definitionbox}
\vspace{0.2cm}

Formally, it is expressed as:

\begingroup
\setlength{\abovedisplayskip}{3pt}
\setlength{\belowdisplayskip}{3pt}
\begin{center}
\tcbox[
  colback=softblue,
  colframe=borderblue,
  boxrule=0.4pt,
  arc=3pt,
  left=6pt,
  right=6pt,
  top=4pt,
  bottom=4pt
]{\textcolor{brandblue}{$\displaystyle
\hat{s}_{\ell+1}
=\WP
\left(
s_\ell,
u_\ell^{\mathcal{F}}
\right),
\qquad
s_\ell \in \mathcal{S}.
$}}
\end{center}
\endgroup

where:

\begin{itemize}[leftmargin=1.35em,itemsep=1pt,topsep=2pt,parsep=0pt,partopsep=0pt,label=\textcolor{brandblue}{\textbullet}]
  \item \textcolor{brandblue}{\textbf{$\ell$}}: \cardstrong{\texttt{interaction step}}, representing the $\ell$-th interaction between the Agent and the World Proxy, not limited to physical time;
  \item \textcolor{brandblue}{\textbf{$\mathcal{S}$}}: \cardstrong{\texttt{information state space}}, a generalized space that may include physical state, observation, memory, knowledge, execution result, verification, guidance, and more;
  
  \item \textcolor{brandblue}{\textbf{$\mathcal{F}$}}: \cardstrong{\texttt{proxy function}}, such as dynamics prediction, spatial rendering, execution simulation, memory retrieval, skill guidance, or reward / verification;
  
  \item \textcolor{brandblue}{\textbf{$u_\ell^{\mathcal{F}}$}}: the \cardstrong{\texttt{query}, \texttt{action}, \texttt{or intervention}} actively proposed by the Agent under a specific proxy function;
  
  \item \textcolor{brandblue}{\textbf{$\hat{s}_{\ell+1}$}}: the \cardstrong{\texttt{feedback}} returned by the World Proxy, such as a future state, novel observation, execution result, retrieved memory / skill, reward, or verification result.
\end{itemize}

A single prediction, though, is not yet improvement. To serve a \textbf{continually improving agent}, the proxy cannot remain a one-shot oracle; it must close the loop with the agent~\cite{wang2023voyager,shinn2023reflexion,webevolver2025}:
\begingroup
\setlength{\belowdisplayskip}{3pt}
\[
\texttt{Agent}
~\rightarrow~
u_\ell^{\mathcal{F}}
\rightarrow
\WP
\rightarrow
\hat{s}_{\ell+1}
~\rightarrow~
\texttt{Agent}.
\]
\endgroup

\begin{figure}[H]
  \centering
  \vspace{0.12cm}
  \captionsetup{skip=5pt,font=small}
  \includegraphics[width=0.95\columnwidth]{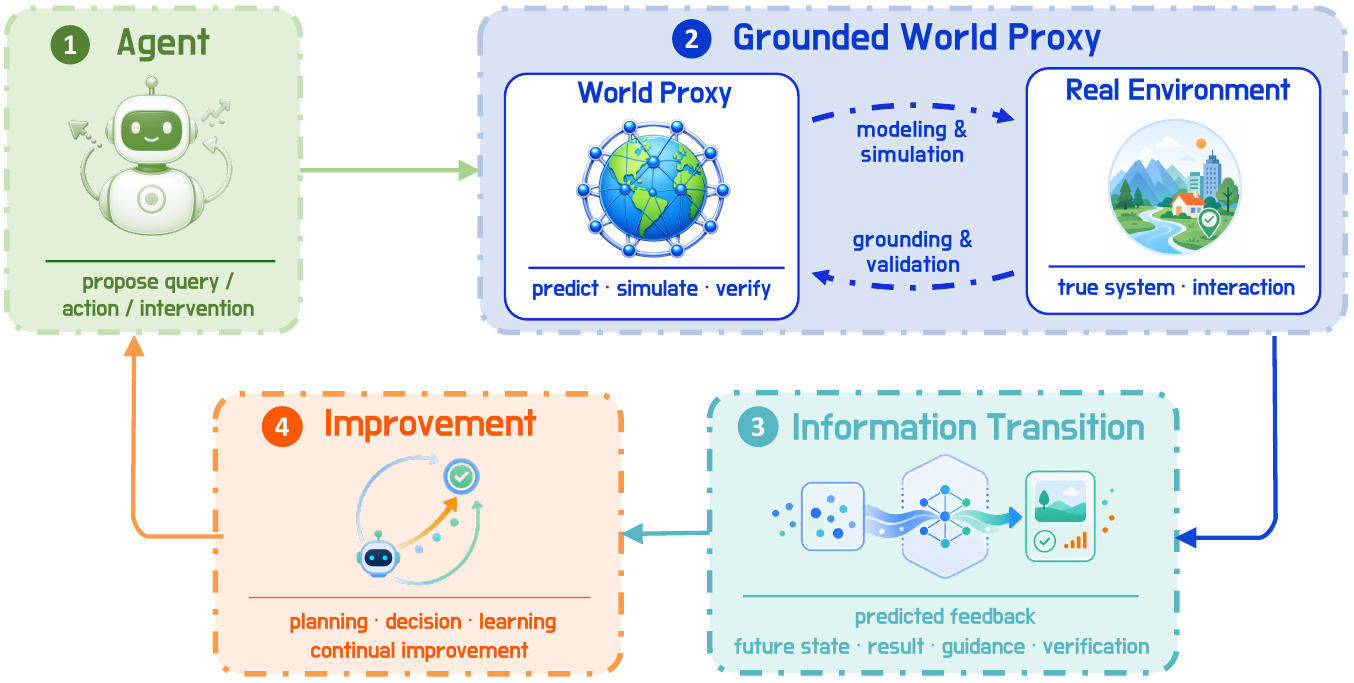}
  \caption{\textbf{Agent-in-the-Loop World Proxy.} The agent queries the World Proxy, which predicts, simulates, retrieves, or verifies the resulting information transition and returns feedback for agent improvement.}
  \label{fig:agent-in-the-loop}
  \vspace{-0.08cm}
\end{figure}

At the $\ell$-th interaction step, this loop can be divided into four steps:

\begin{itemize}[leftmargin=1.35em,itemsep=1pt,topsep=1pt,parsep=0pt,partopsep=0pt,label=\textcolor{brandblue}{\textbullet}]
  \item \cardstrong{Step \textbf{\texttt{1}}: Agent proposes an interaction}\\
  The Agent actively proposes an interaction request $u_\ell^{\mathcal{F}}$ according to its current goal, context state, and other conditions.

  \item \cardstrong{Step \textbf{\texttt{2}}: World Proxy predicts information transition}\\
  The World Proxy predicts or generates feedback conditioned on the current information state $s_\ell$ and the agent interaction $u_\ell^{\mathcal{F}}$.

  \item \cardstrong{Step \textbf{\texttt{3}}: Proxy feedback is returned to the Agent}\\
  The predicted $\hat{s}_{\ell+1}$ is returned as new feedback or information gain, such as a future state, predicted observation, execution result, retrieved guidance, or verification result.

  \item \cardstrong{Step \textbf{\texttt{4}}: Agent uses feedback for improvement}\\
  The Agent uses the feedback for planning, decision making, policy learning, or continual improvement.
\end{itemize}

Run once, this is a single prediction; iterated over many steps, it becomes a trajectory of improvement, since each answer quietly reshapes the next question the Agent thinks to ask.

\subsection{What Makes a Good World Proxy?}

Not every intermediate module is an effective World Proxy. For continually improving agents, a good World Proxy should provide feedback that is \textbf{grounded, controllable, actionable, scalable, and forward-looking} within the agent's closed loop.

\begin{table}[H]
\centering
\footnotesize
\setlength{\tabcolsep}{4pt}
\renewcommand{\arraystretch}{1.10}
\captionsetup{skip=4pt}
\caption{Criteria for effective World Proxies.}
\label{tab:good-world-proxy}
\begin{minipage}{0.94\columnwidth}
\begin{tabularx}{\linewidth}{
  >{\raggedright\arraybackslash}p{0.21\linewidth}
  >{\raggedright\arraybackslash}X
}
\toprule
{\sffamily\bfseries\textcolor{brandblue}{Criterion}}
& {\sffamily\bfseries\textcolor{brandblue}{Meaning}}\\
\midrule
\textcolor{brandblue}{\textbf{Groundedness}}
& Relies on real environment data, rules, trajectories, or interaction evidence, not detached generation.\\
\addlinespace[2pt]
\textcolor{brandblue}{\textbf{Controllability}}
& Supports agent-initiated queries, actions, or interventions and returns corresponding feedback.\\
\addlinespace[2pt]
\textcolor{brandblue}{\textbf{Feedback Usefulness}}
& Improves planning, decision making, policy learning, or continual improvement.\\
\addlinespace[2pt]
\textcolor{brandblue}{\textbf{Cost and Scalability}}
& Reduces real-world interaction cost while scaling safely across tasks and settings.\\
\addlinespace[2pt]
\textcolor{brandblue}{\textbf{Forward-Looking Ability}}
& Anticipates outcomes, counterfactuals, or risks before real execution.\\
\bottomrule
\end{tabularx}
\end{minipage}
\end{table}

In practice, useful World Proxies must balance real-environment grounding, scalability, and actionable feedback; proxies that drift too far from grounding soon stop being useful, and confidently wrong ones can be worse than none~\cite{vafa2024evaluating}.

\begin{highlightcard}{\cardicon{figs/idea.png} Takeaway}
\begin{enumerate}[leftmargin=1.45em,itemsep=0pt,topsep=0pt,parsep=0pt,partopsep=0pt,label=\textcolor{brandblue}{\textbf{\arabic*.}}]
  \item \cardicon{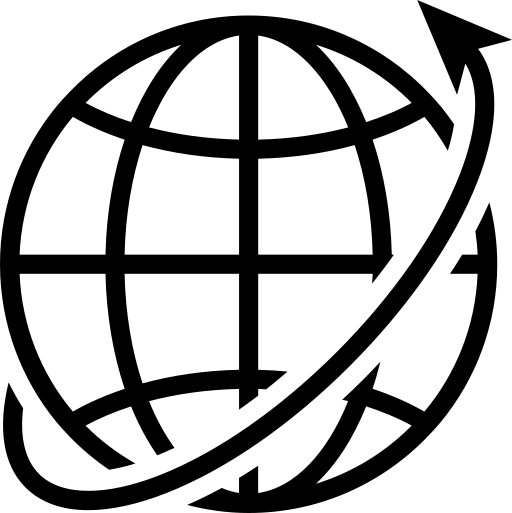}\cardstrong{From World Model to World Proxy:} 
  \\
  \cardtext{Extending beyond future-state prediction to spatial rendering, execution simulation, memory retrieval, skill guidance, reward modeling, and verification.}
  \item \cardicon{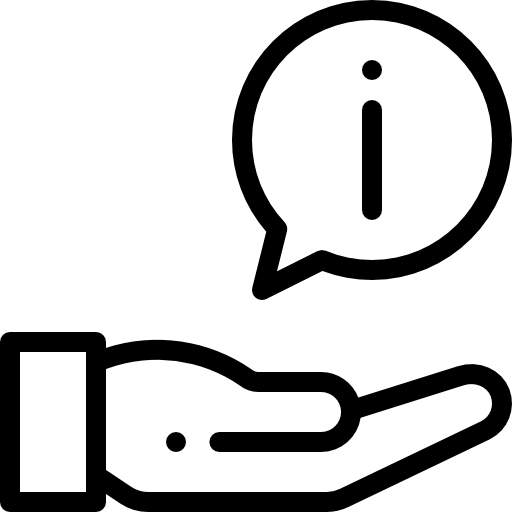}\cardstrong{From Realism to Information Gain:}
  \\
  \cardtext{Moving beyond prediction accuracy to prioritize \cardkey{agent-usable feedback} for planning, decision making, learning, and continual improvement.}
  \item \cardicon{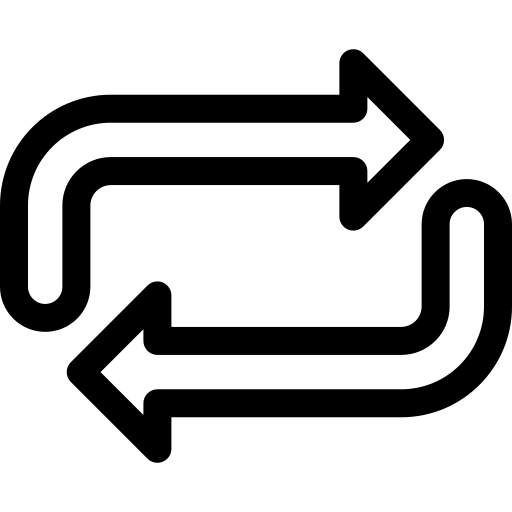}\cardstrong{From one-shot prediction to closed loop:}
  \\
  \cardtext{Use agent-initiated interaction and proxy feedback to drive \cardkey{iterative improvement}.}
\end{enumerate}
\end{highlightcard}
\vspace{0.2cm}

%% file: sections/empowerment.tex
\vspace{0.2cm}
\section{Empowerment: How World Proxies Improve Agents}
\label{sec:empowerment}

With the World Proxy defined, the obvious question is what it actually buys us:

\vspace{0.2cm}
\begin{questionbox}
\questiontext{How can World Proxies help an Agent \questionkey{plan}, \questionkey{learn}, and \questionkey{evolve}?}
\end{questionbox}
\vspace{0.2cm}

One way to answer is to grade the world model by its \textbf{own intrinsic capability}, as Chu et al.~\cite{chu2026agentic} do:

\begin{itemize}[leftmargin=1.35em,itemsep=2pt,topsep=4pt,parsep=0pt,partopsep=0pt,label=\textcolor{brandblue}{\textbullet}]
  \item \textbf{L.1~~Predictor}: one-step / local transition prediction;
  \item \textbf{L.2~~Simulator}: long-horizon, action-conditioned rollout;
  \item \textbf{L.3~~Evolver}: world model self-reflection.
\end{itemize}

\begin{figure}[H]
  \centering
  \vspace{0.12cm}
  \captionsetup{skip=5pt,font=small}
  \begin{minipage}{0.98\columnwidth}
    \centering
    \includegraphics[width=\linewidth]{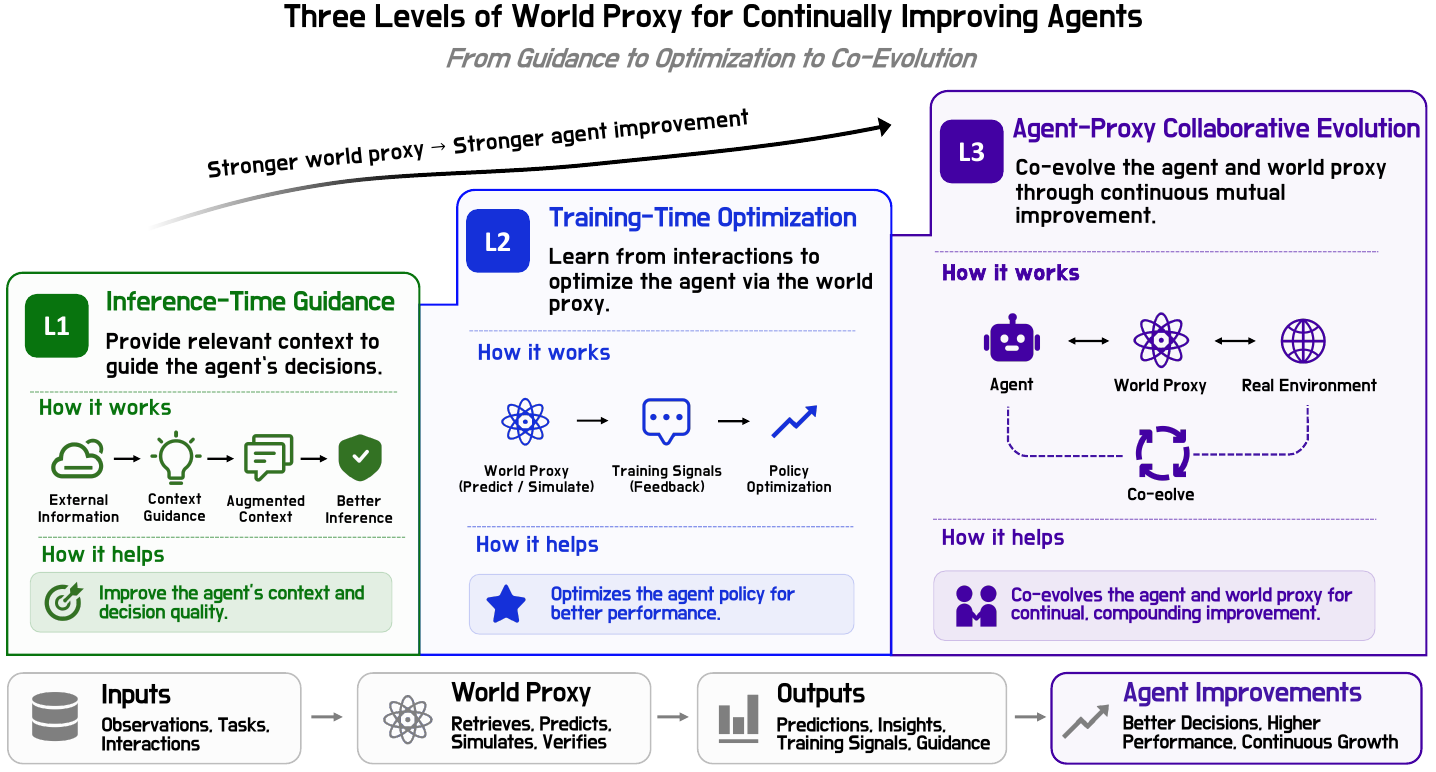}
    \caption{\textbf{L1 to L3 World Proxies for Agent Improvement.} L1 augments inference-time context, L2 provides training signals, and L3 enables Agent-Proxy co-evolution.}
    \label{fig:l1-to-l3}
  \end{minipage}
  \vspace{-0.08cm}
\end{figure}

These levels grade the world model's \emph{own} competence. Our question is orthogonal: not how capable the model is in isolation, but how much better it makes the agent. We deliberately reuse the \textbf{L1-L3} shorthand for this agent-centric axis, so the two scales rhyme without being identical; Table~\ref{tab:two-scales} sets them side by side to keep the two readings distinct.

\vspace{0.18cm}
\begin{keypointbox}
\blocktitle{World-Proxy-Driven Agent Improvement}
\small
We take an \textcolor{brandblue}{\textbf{agent-centric perspective}}: instead of asking how powerful the World Model itself is, we ask how a World Proxy drives Agent improvement. This leads to three levels:

\begin{tcolorbox}[
  enhanced,
  colback=white,
  colframe=borderblue,
  borderline west={2pt}{0pt}{levelonebg},
  boxrule=0.45pt,
  arc=4pt,
  left=8pt,right=8pt,top=5pt,bottom=5pt,
  before skip=6pt,after skip=4pt
]
\noindent{\sffamily\bfseries \LoneBadge}\hspace{0.55em}
\LoneText{Inference-Time Guidance}\par\vspace{2pt}
\begin{itemize}[leftmargin=1.3em,itemsep=1pt,topsep=1pt,parsep=0pt,partopsep=0pt,label=\textcolor{levelonetext}{\textbullet}]
    \item \textbf{Proxy Role:} augments inference-time context.
    \item \textbf{Agent Effect:} better decisions.
\end{itemize}
\end{tcolorbox}

\begin{tcolorbox}[
  enhanced,
  colback=white,
  colframe=borderblue,
  borderline west={2pt}{0pt}{leveltwobg},
  boxrule=0.45pt,
  arc=4pt,
  left=8pt,right=8pt,top=5pt,bottom=5pt,
  before skip=4pt,after skip=4pt
]
\noindent{\sffamily\bfseries \LtwoBadge}\hspace{0.55em}
\LtwoText{Training-Time Optimization}\par\vspace{2pt}
\begin{itemize}[leftmargin=1.3em,itemsep=1pt,topsep=1pt,parsep=0pt,partopsep=0pt,label=\textcolor{leveltwotext}{\textbullet}]
    \item \textbf{Proxy Role:} provides reward, verification, or simulation signals.
    \item \textbf{Agent Effect:} optimizes the policy.
\end{itemize}
\end{tcolorbox}

\begin{tcolorbox}[
  enhanced,
  colback=white,
  colframe=borderblue,
  borderline west={2pt}{0pt}{levelthreebg},
  boxrule=0.45pt,
  arc=4pt,
  left=8pt,right=8pt,top=5pt,bottom=5pt,
  before skip=4pt,after skip=1pt
]
\noindent{\sffamily\bfseries \LthreeBadge}\hspace{0.55em}
\LthreeText{Agent-Proxy Co-Evolution}\par\vspace{2pt}
\begin{itemize}[leftmargin=1.3em,itemsep=1pt,topsep=1pt,parsep=0pt,partopsep=0pt,label=\textcolor{levelthreetext}{\textbullet}]
    \item \textbf{Proxy Role:} closes the Agent-Proxy-Environment loop.
    \item \textbf{Agent Effect:} continual co-evolution.
\end{itemize}
\end{tcolorbox}
\end{keypointbox}
\vspace{0.14cm}

\begin{table}[H]
\centering
\footnotesize
\setlength{\tabcolsep}{5pt}
\renewcommand{\arraystretch}{1.18}
\captionsetup{skip=4pt,font=small}
\caption{Two orthogonal readings of the \textbf{L1-L3} shorthand: a world model's \emph{intrinsic} capability~\cite{chu2026agentic} versus the \emph{agent-centric} empowerment used in this article. The rungs align in spirit, not in definition.}
\label{tab:two-scales}
\begin{tabularx}{\columnwidth}{
  >{\centering\arraybackslash}p{0.06\columnwidth}
  !{\color{bordergray}\vrule width 0.45pt}
  >{\raggedright\arraybackslash}X
  !{\color{bordergray}\vrule width 0.45pt}
  >{\raggedright\arraybackslash}X
}
\toprule
{\sffamily\bfseries\textcolor{brandblue}{Level}}
& {\sffamily\bfseries\textcolor{brandblue}{Intrinsic WM Capability (Chu et al.)}}
& {\sffamily\bfseries\textcolor{brandblue}{Agent-Centric Empowerment (Ours)}}\\
\midrule
\LoneBadge
& \textbf{Predictor}: one-step / local transition prediction
& \LoneText{Inference-Time Guidance}: enrich context for better decisions\\
\addlinespace[2.5pt]
\LtwoBadge
& \textbf{Simulator}: long-horizon, action-conditioned rollout
& \LtwoText{Training-Time Optimization}: reshape the agent policy\\
\addlinespace[2.5pt]
\LthreeBadge
& \textbf{Evolver}: world-model self-reflection
& \LthreeText{Agent-Proxy Co-Evolution}: continual mutual improvement\\
\bottomrule
\end{tabularx}
\end{table}

\vspace{0.18cm}
Read together, the three levels form a ladder of increasing commitment. \LoneInline{} leaves the agent untouched and merely informs its next move; \LtwoInline{} reaches into the agent's parameters and rewrites its policy; \LthreeInline{} lets agent and proxy reshape each other over time. Capability grows at every rung, and so does the burden of proof on the proxy, because the deeper its feedback reaches, the more a mistake costs.

\vspace{0.28cm}
\subsection[L1: Inference-Time Guidance]{L1: Inference-Time Guidance}
\enlargethispage{0.5\baselineskip}

\begin{figure}[H]
  \centering
  \vspace{0.12cm}
  \captionsetup{skip=5pt,font=small}
  \includegraphics[width=0.999\columnwidth]{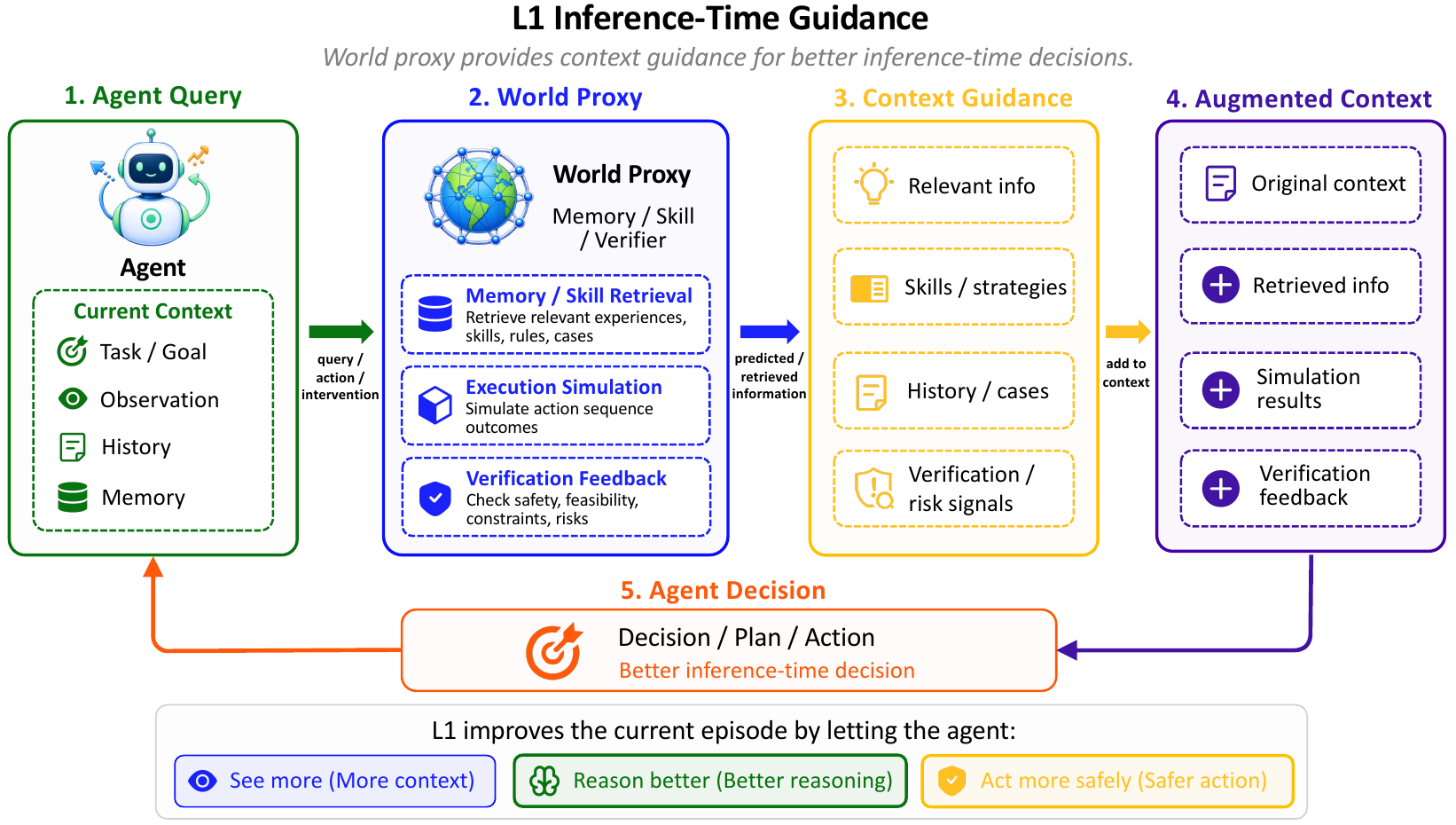}
  \caption{\textbf{L1 Inference-Time Guidance via World Proxy.} The Agent queries the World Proxy, which retrieves, simulates, or verifies relevant information. The returned guidance is added to the Agent's context for better inference-time decisions.}
  \label{fig:l1-improve}
  \vspace{-0.08cm}
\end{figure}

We begin with the lightest touch. At \LoneInline{}, the World Proxy never alters the Agent's parameters; it simply hands over extra context, experience, skill, or verification feedback at inference time, so the decision the Agent is about to make is a better-informed one~\cite{wei2022cot,kojima2022zeroshotcot,yao2022react,yao2023tot,hao2023reasoning,gu2024webdreamer}. Because nothing is retrained, \LoneInline{} is cheap and fully reversible; its ceiling, though, is the agent's existing competence, since it can only recombine what the agent already knows. Formally:

\begingroup
\setlength{\abovedisplayskip}{3pt}
\setlength{\belowdisplayskip}{3pt}
\begin{center}
\tcbox[
  colback=softblue,
  colframe=borderblue,
  boxrule=0.4pt,
  arc=3pt,
  left=6pt,
  right=6pt,
  top=4pt,
  bottom=4pt
]{\textcolor{brandblue}{$\displaystyle
\hat{s}_{\ell+1}^{guide}
=\WP
\left(
s_{\ell}^{agent},
u_{\ell}^{\mathcal{F}}
\right),
\qquad
s_{\ell}^{agent+}
=s_{\ell}^{agent}
\oplus
\hat{s}_{\ell+1}^{guide}.
$}}
\end{center}
\endgroup

\vspace{1pt}
Here, $\hat{s}_{\ell+1}^{guide}$ denotes guidance returned by the World Proxy, and $\oplus$ denotes adding it to the Agent's current context for information augmentation.

\localheading{Process}

The flow runs in one direction, from the agent's question to a richer context for its very next move; Fig.~\ref{fig:l1-improve} gives the corresponding visual layout:

\begin{center}
\begin{tcolorbox}[
  enhanced,
  width=0.92\columnwidth,
  colback=softgray,
  colframe=bordergray,
  boxrule=0.4pt,
  arc=4pt,
  left=7pt,right=7pt,top=4pt,bottom=4pt,
  before skip=2pt,after skip=4pt
]
\centering\small
\textcolor{brandblue}{\sffamily\bfseries Agent Query}
$\;\longrightarrow\;$
\textcolor{brandblue}{\sffamily\bfseries Proxy} Retrieval / Simulation / Verification
$\;\longrightarrow\;$
\par\vspace{2pt}
\textcolor{brandblue}{\sffamily\bfseries Guidance}
$\;\longrightarrow\;$
Augmented Context
$\;\longrightarrow\;$
\textcolor{brandblue}{\sffamily\bfseries Agent Decision}
\end{tcolorbox}
\end{center}

\localheading{Implementations}

\begin{itemize}[leftmargin=1.35em,itemsep=2pt,topsep=3pt,parsep=0pt,partopsep=0pt,label=\textcolor{brandblue}{\textbullet}]
  \item \cardstrong{Memory / Skill Retrieval}: retrieves historical experience, strategies, skills, or tool-use rules according to the current task~\cite{park2023generative,sumers2024coala,shinn2023reflexion,wang2023voyager,schick2023toolformer, yu2025vismem};
  \item \cardstrong{Execution Simulation}: simulates the result of an action or action sequence before real execution~\cite{gu2024webdreamer,tang2024worldcoder,rivard2025neuralos};
  \item \cardstrong{Verification Feedback}: determines whether the current plan / action is safe, feasible, and consistent with constraints~\cite{cobbe2021verifiers,lightman2023letsverify,mcaleese2024criticgpt}.
\end{itemize}

In practice, this is the agent looking before it leaps: a web agent about to click \emph{Purchase} can first ask the proxy to imagine the resulting page, then revise its plan if that page shows an error or an unintended charge, all without touching the live site~\cite{gu2024webdreamer}.
\vspace{0.12cm}
\begin{keypointbox}
\blocktitle{Key Point}
\textcolor{brandblue}{\textbf{L1}} lets the Agent \textcolor{brandblue}{\textbf{gain richer context}} within the current episode, so it can ``\textcolor{brandblue}{\textbf{see more, reason more accurately, and act more robustly}}.''
\end{keypointbox}

\vspace{0.28cm}
\subsection[L2: Training-Time Optimization]{L2: Training-Time Optimization}
\enlargethispage{1.5\baselineskip}

\LtwoInline{} raises the stakes. The World Proxy now does more than whisper context at inference time; it serves as a reward model, verifier, critic, or simulator, producing the training signals that reshape the Agent's policy itself~\cite{christiano2017deeprlpreferences,ouyang2022instructgpt,cobbe2021verifiers,lightman2023letsverify,chen2025dreamgym}. This lifts the agent's ceiling rather than merely its current context, but the gain is only as trustworthy as the signal behind it: a biased reward quietly teaches biased behavior. Formally:

\begingroup
\setlength{\abovedisplayskip}{3pt}
\setlength{\belowdisplayskip}{3pt}
\begin{center}
\tcbox[
  colback=softblue,
  colframe=borderblue,
  boxrule=0.4pt,
  arc=3pt,
  left=6pt,
  right=6pt,
  top=4pt,
  bottom=4pt
]{\textcolor{brandblue}{$\displaystyle
\hat{s}_{\ell+1}^{opt}
=\WP
\left(
s_{\ell}^{agent},
u_{\ell}^{\mathcal{F}}
\right),
\qquad
\text{agent}^{+}
=Train
\left(
\text{agent},
\hat{s}_{\ell+1}^{opt}
\right).
$}}
\end{center}
\endgroup

\vspace{1pt}
Here, $\hat{s}_{\ell+1}^{opt}$ denotes training signals generated by the World Proxy, such as reward, verification, critique, or simulated rollouts. These signals can be converted into objectives such as SFT, DPO~\cite{rafailov2023dpo}, PPO~\cite{schulman2017ppo}, or GRPO~\cite{shao2024deepseekmath}.

Compared with \LoneInline{}, the key change in \LtwoInline{} is:

\begin{center}
\small
\textcolor{brandblue}{\sffamily\bfseries Proxy Output as Context}
$\;\longrightarrow\;$
\textcolor{brandblue}{\sffamily\bfseries Proxy Output as Training Signal}
\end{center}

\begin{figure}[H]
  \centering
  \vspace{0.12cm}
  \captionsetup{skip=5pt,font=small}
  \includegraphics[width=0.999\columnwidth]{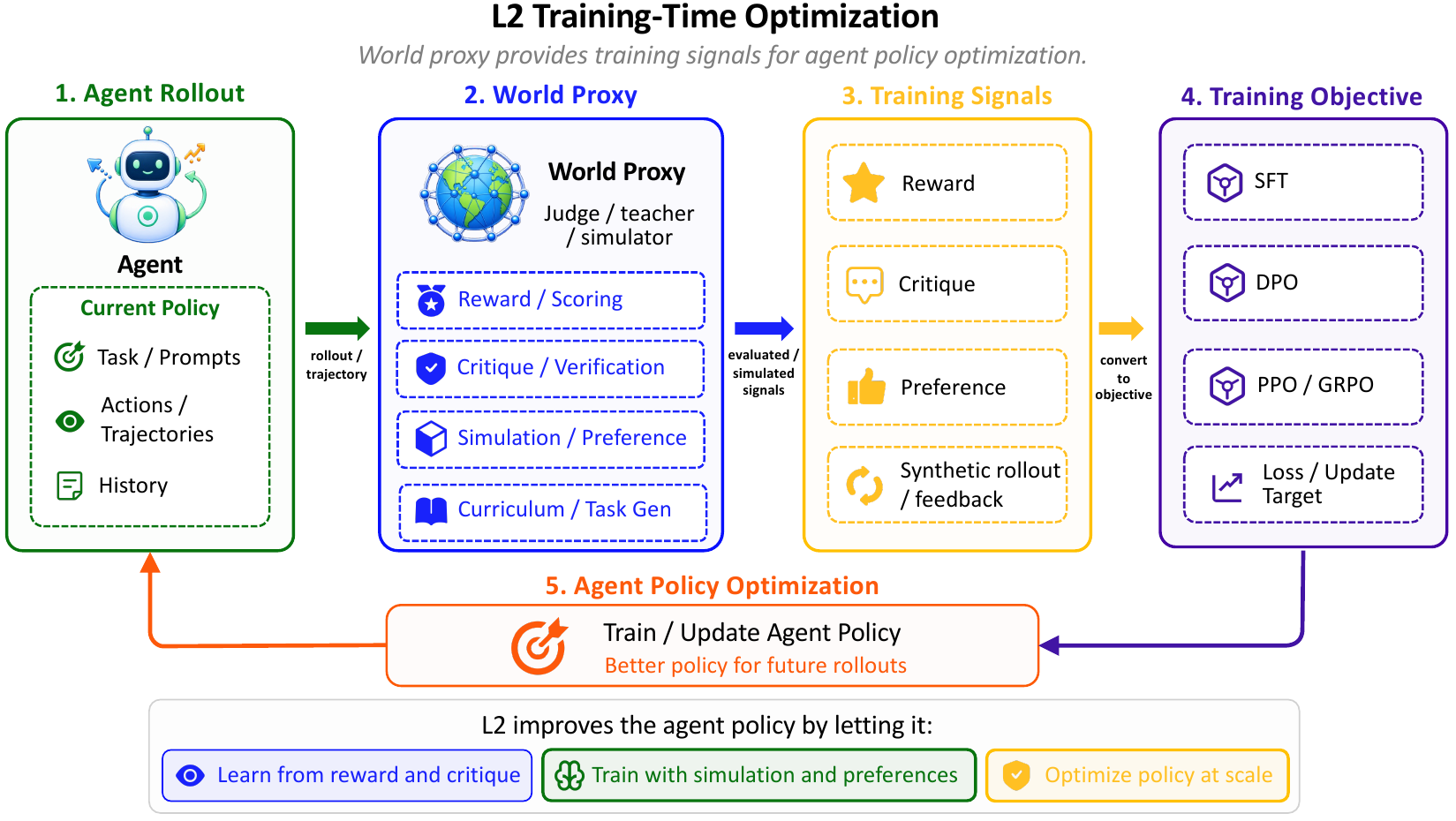}
  \caption{\textbf{L2 Training-Time Optimization Driven by World Proxy.} The Agent generates rollouts, which are evaluated, verified, or simulated by the World Proxy. The resulting reward, critique, preference, or verification signal is converted into a training objective to optimize the Agent policy.}
  \label{fig:l2-improve}
  \vspace{-0.08cm}
\end{figure}

\localheading{Process}

Now the loop bends back into learning: the proxy's verdict on a rollout becomes a gradient on the policy rather than a hint for the moment; Fig.~\ref{fig:l2-improve} gives the corresponding visual layout.

\begin{center}
\begin{tcolorbox}[
  enhanced,
  width=0.94\columnwidth,
  colback=softgray,
  colframe=bordergray,
  boxrule=0.4pt,
  arc=4pt,
  left=7pt,right=7pt,top=4pt,bottom=4pt,
  before skip=2pt,after skip=4pt
]
\centering\small
\textcolor{brandblue}{\sffamily\bfseries Agent Rollout}
$\;\longrightarrow\;$
\textcolor{brandblue}{\sffamily\bfseries Proxy} Verification / Simulation
\,$\longrightarrow$\,
Reward / Critique / Preference
\par\vspace{2pt}
$\;\longrightarrow\;$
Training Objective
$\;\longrightarrow\;$
\textcolor{brandblue}{\sffamily\bfseries Agent Policy Optimization}
\end{tcolorbox}
\end{center}

\localheading{Implementations}

\begin{itemize}[leftmargin=1.35em,itemsep=1pt,topsep=2pt,parsep=0pt,partopsep=0pt,label=\textcolor{brandblue}{\textbullet}]
  \item \cardstrong{Proxy-as-Reward}:\\The World Proxy or verifier scores trajectories to form rewards~\cite{christiano2017deeprlpreferences,ouyang2022instructgpt,bai2022hh}.
  
  \item \cardstrong{Proxy-as-Critic}:\\Identifies failure causes and outputs critiques, error diagnoses, or constraint violations~\cite{mcaleese2024criticgpt,madaan2023selfrefine,zhang2025critic}.
  \item \cardstrong{Proxy-as-Simulator}:\\Generates synthetic trajectories or constructs preference pairs for DPO / RLHF / GRPO~\cite{chen2025dreamgym,wang2025ragen,wang2025vagen}.
  \item \cardstrong{Proxy-Guided Curriculum}:\\Generates harder or more targeted tasks based on the Agent's current failure modes~\cite{wang2026awm,zhang2026searchgym}.
\end{itemize}

In practice, the agent's own rollouts become training fuel: the proxy scores, verifies, or replays them into synthetic trajectories and preference pairs, so a policy can be optimized at a scale that collecting real interactions could never reach~\cite{chen2025dreamgym,wang2025ragen}.
\vspace{0.06cm}

\begin{keypointbox}
\blocktitle{Key Point}
\textcolor{brandblue}{\textbf{L2}} upgrades the World Proxy from an ``inference-time advisor'' to a ``\textcolor{brandblue}{\textbf{training-time judge, teacher, or simulated environment}},'' directly \textcolor{brandblue}{\textbf{optimizing the Agent policy}}.
\end{keypointbox}

\vspace{0.12cm}
\subsection[L3: Agent-Proxy Co-Evolution]{L3: Agent-Proxy Co-Evolution}
\enlargethispage{1.5\baselineskip}

\begin{figure}[H]
  \centering
  \vspace{0.12cm}
  \captionsetup{skip=4pt,font=small}
  \includegraphics[width=0.999\columnwidth]{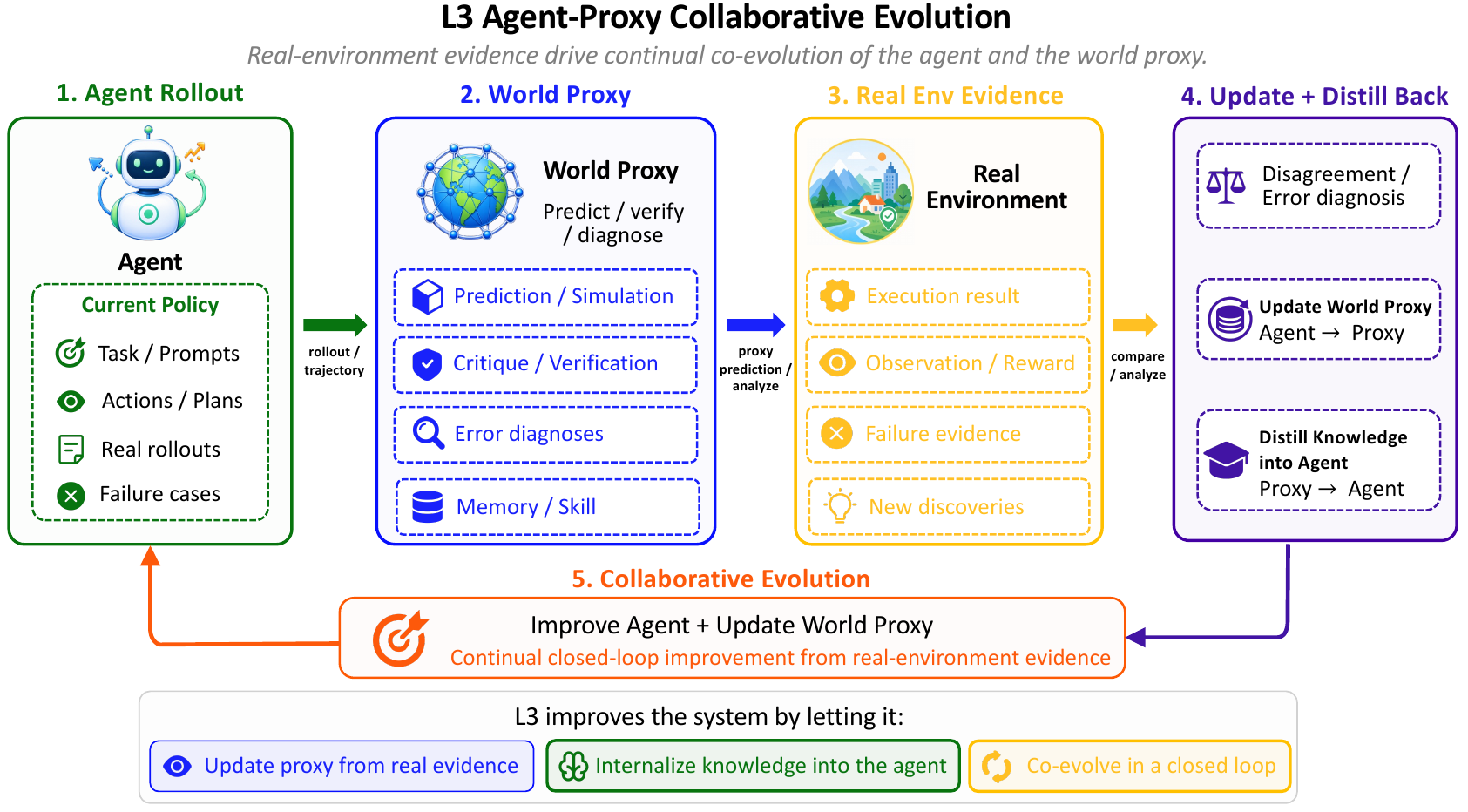}
  \caption{\textbf{L3 Agent-Proxy Co-Evolution.} Real-environment evidence updates the World Proxy, while useful proxy knowledge is distilled back into the Agent policy for continual improvement.}
  \label{fig:l3-improve}
  \vspace{-0.08cm}
\end{figure}

\LthreeInline{} completes the arc by closing a continual loop among the Agent, the World Proxy, and the real environment. Real trajectories, failures, and fresh discoveries update the proxy; the sharpened proxy then guides, verifies, and trains the Agent, so both improve together rather than one after the other~\cite{wang2023voyager,shinn2023reflexion,webevolver2025,ren2026aligning}. This is the most powerful rung, but it also requires machinery that keeps proxy and policy aligned as both change. Formally:

\begingroup
\setlength{\abovedisplayskip}{3pt}
\setlength{\belowdisplayskip}{3pt}
\begin{center}
\tcbox[
  colback=softblue,
  colframe=borderblue,
  boxrule=0.4pt,
  arc=3pt,
  left=6pt,
  right=6pt,
  top=4pt,
  bottom=4pt
]{\textcolor{brandblue}{$\displaystyle
\hat{s}_{\ell+1}^{proxy}
=\WP
\left(
s_{\ell}^{agent},
u_{\ell}^{\mathcal{F}}
\right),
\qquad
(\text{agent}^{+},\WP^{+})
=CoEvolve
\left(
\text{agent},
\WP,
\hat{s}_{\ell+1}^{proxy},
s_{\ell+1}^{env}
\right)
$}}
\end{center}
\endgroup

\vspace{0pt}
Here, $\hat{s}_{\ell+1}^{proxy}$ denotes proxy feedback, $s_{\ell+1}^{env}$ denotes real-environment evidence, and $CoEvolve(\cdot)$ updates both the Agent and the World Proxy.

\localheading{Process}

The loop now uses real-environment gaps as the update signal; Fig.~\ref{fig:l3-improve} gives the corresponding visual layout:

\begin{center}
\begin{tcolorbox}[
  enhanced,
  width=0.96\columnwidth,
  colback=softgray,
  colframe=bordergray,
  boxrule=0.4pt,
  arc=4pt,
  left=7pt,right=7pt,top=3pt,bottom=3pt,
  before skip=1pt,after skip=2pt
]
\centering\small
\textcolor{brandblue}{\sffamily\bfseries Agent Rollout}
$\;\longrightarrow\;$
\textcolor{brandblue}{\sffamily\bfseries Proxy} Prediction / Verification
$\;\longrightarrow\;$
\textcolor{brandblue}{\sffamily\bfseries Real-Environment} Evidence
$\;\longrightarrow\;$
\par\vspace{2pt}
Disagreement / Error Diagnosis
$\;\longrightarrow\;$
\textcolor{brandblue}{\sffamily\bfseries Update Proxy + Distill to Agent}
$\;\longrightarrow\;$
\textcolor{brandblue}{\sffamily\bfseries Co-Evolution}
\end{tcolorbox}
\end{center}

\newpage

\localheading{Implementations}

\begin{itemize}[leftmargin=1.35em,itemsep=1pt,topsep=2pt,parsep=0pt,partopsep=0pt,label=\textcolor{brandblue}{\textbullet}]
  \item \cardstrong{Proxy $\rightarrow$ Agent Internalization}: memory, skills, constraints, verification rules, or reward signals are distilled into Agent parameters or policies, turning external feedback into internal capability~\cite{wang2023voyager,liang2022codeaspolicies,ahn2022saycan}.
  \item \cardstrong{Agent $\rightarrow$ Proxy Update}: real interaction trajectories, effective experiences, failure cases, and environmental feedback update the World Proxy's memory, skill library, verifier, simulator, or reward model~\cite{shinn2023reflexion,webevolver2025,butt2024codeit}.
\end{itemize}

In practice, picture a web agent whose every deployment both draws on the proxy's predictions and feeds fresh trajectories back into it: proxy and policy are retrained in tandem, each cycle shrinking the gap between what the agent imagines and what the world actually returns~\cite{webevolver2025}.
\vspace{0.06cm}

\vspace{0.2cm}
\begin{keypointbox}
\blocktitle{Key Point}
\textcolor{brandblue}{\textbf{L3}} makes the World Proxy more than a static tool: it becomes a \textcolor{brandblue}{\textbf{co-evolving partner}} that is continually updated through real-environment evidence.
\end{keypointbox}

\vspace{0.2cm}

\vspace{0.2cm}
\begin{highlightcard}{\cardicon{figs/idea.png} Worked Example: One Web Agent up the Ladder}
To see the three levels as one continuum rather than three separate tricks, follow a single web agent completing a multi-step online purchase, and watch the \emph{same} execution proxy deepen its role at each rung:

\begin{itemize}[leftmargin=1.35em,itemsep=3pt,topsep=3pt,parsep=0pt,partopsep=0pt,label=\textcolor{brandblue}{\textbullet}]
  \item \LoneBadge~\LoneText{Guidance.} Before clicking \emph{Purchase}, the agent asks the proxy to imagine the resulting page; if that page shows an error or an unintended charge, it revises its plan, without ever touching the live site~\cite{gu2024webdreamer}.
  \item \LtwoBadge~\LtwoText{Optimization.} Those imagined rollouts are not discarded: the proxy scores and replays them into synthetic trajectories and preference pairs, and the agent's policy is optimized on this fabricated experience at a scale live interaction could never reach~\cite{chen2025dreamgym}.
  \item \LthreeBadge~\LthreeText{Co-Evolution.} Once deployed, the agent's real trajectories flow back to retrain the proxy, and the sharpened proxy in turn yields better guidance and training signal, each cycle shrinking the gap between what the agent imagines and what the web actually returns~\cite{webevolver2025}.
\end{itemize}

One agent, one proxy function, three escalating roles: \textbf{advisor}, \textbf{teacher}, and \textbf{partner}.
\end{highlightcard}

\vspace{0.16cm}

\begin{highlightcard}{\cardicon{figs/idea.png} Takeaway}
\cardstrong{World Proxies empower agents through three progressive levels:}

\begin{itemize}[leftmargin=1.35em,itemsep=3pt,topsep=3pt,parsep=0pt,partopsep=0pt,label=\textcolor{brandblue}{\textbullet}]
  \item \LoneBadge~\LoneText{Inference-Time Guidance.}
  The World Proxy provides memory, skill, simulation, or verification feedback to \textcolor{brandblue}{\textbf{enrich inference-time context}} and support better decisions.

  \item \LtwoBadge~\LtwoText{Training-Time Optimization.}
  The World Proxy acts as a reward model, critic, verifier, or simulator, generating \textcolor{brandblue}{\textbf{training signals}} that directly \textcolor{brandblue}{\textbf{optimize the Agent policy}}.

  \item \LthreeBadge~\LthreeText{Agent-Proxy Co-Evolution.}
  \textcolor{brandblue}{\textbf{Real-environment evidence}} updates the World Proxy, whose knowledge is \textcolor{brandblue}{\textbf{distilled back}} into the Agent for \textcolor{brandblue}{\textbf{continual improvement}}.
\end{itemize}
\end{highlightcard}

%% file: sections/instantiations.tex
\section{Instantiations: Functional Forms of Agent-Centric World Proxies}
\label{sec:instantiations}
\enlargethispage{1.0\baselineskip}

If Section~\ref{sec:empowerment} asked \emph{how} a World Proxy helps, this section asks \emph{in what form} it appears. The single symbol $\mathcal{F}$ in our definition quietly stands in for a whole family of proxy functions; here we unpack it into six concrete forms: \textbf{dynamics prediction, spatial observation, execution simulation, memory retrieval, skill guidance, and reward/verification}. Each has grown into a research area in its own right, and several recent surveys map them in depth~\cite{zhu2024sora_survey,li2025embodied_wm_survey,feng2025ad_wm_survey,tu2025wm_ad_survey}.

\begin{figure}[H]
  \centering
  \captionsetup{skip=5pt,font=small}
  \includegraphics[width=0.999\columnwidth]{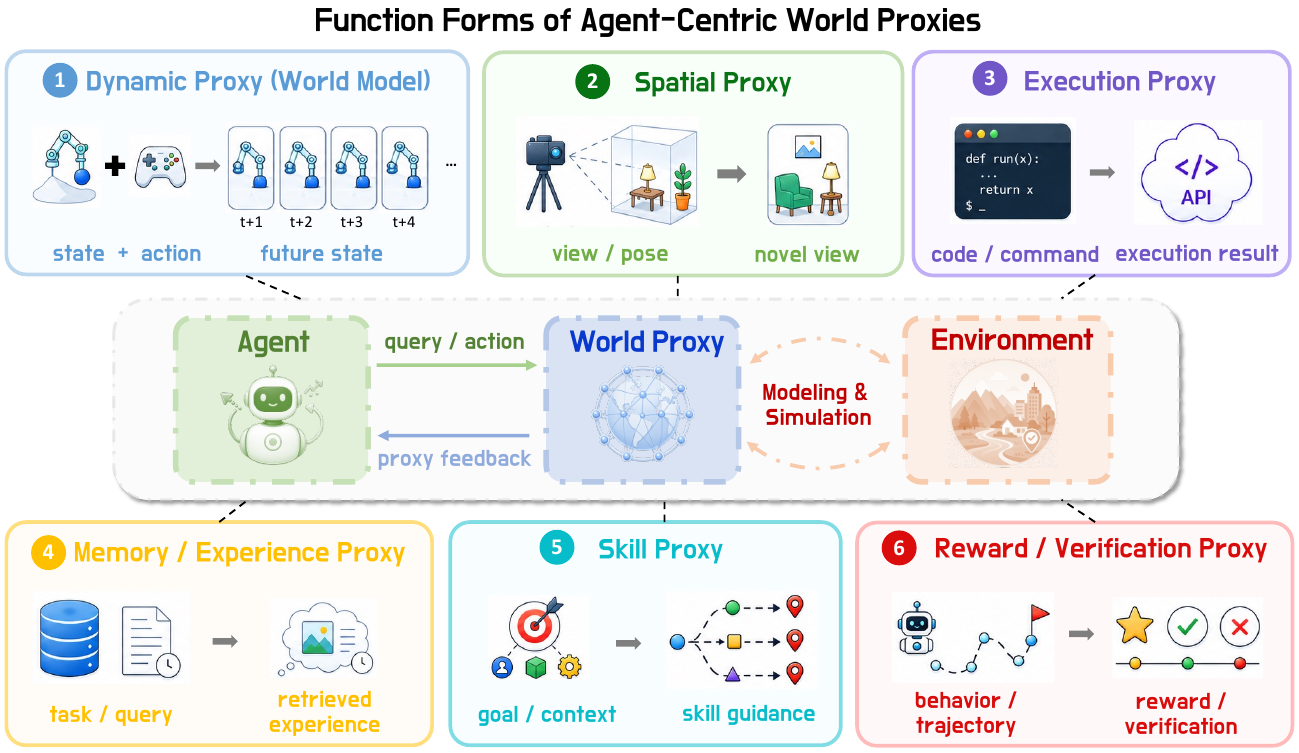}
  \caption{\textbf{Functional forms of Agent-Centric World Proxies.} A World Proxy may simulate dynamics, render spatial observations, predict execution outcomes, retrieve memory, suggest skills, or provide reward / verification feedback.}
  \label{fig:agent-wp-functions}
  \vspace{0.2cm}
\end{figure}

\subsection{Overview}
\vspace{-0.45\baselineskip}

The six forms differ less in their machinery than in the \emph{question} each one answers for the agent. Table~\ref{tab:proxy-functions} lines them up at a glance, pairing every function with the input it consumes, the slice of the world it stands in for, and what it hands back; we then take each in turn.

\newcommand{\proxyformulabox}[1]{%
\begingroup
\setlength{\abovedisplayskip}{3pt}%
\setlength{\belowdisplayskip}{3pt}%
\begin{center}
\tcbox[
  colback=softblue,
  colframe=borderblue,
  boxrule=0.4pt,
  arc=3pt,
  left=6pt,
  right=6pt,
  top=4pt,
  bottom=4pt
]{\textcolor{brandblue}{$\displaystyle #1$}}
\end{center}
\endgroup
}

\newenvironment{proxyitems}{%
\begin{itemize}[leftmargin=1.25em,itemsep=1pt,topsep=2pt,parsep=0pt,partopsep=0pt,label=\textcolor{brandblue}{\textbullet}]
}{%
\end{itemize}
}

\newcommand{\proxyheading}[1]{\Needspace{5\baselineskip}\localheading{#1}}

\subsection{Dynamics Proxy (World Model)}

\proxyheading{Formula}

\proxyformulabox{
\hat{s}_{\ell+1}, \hat{r}_{\ell+1}
=\WP^{\mathrm{dyn}}
\left(
s_\ell,
u_\ell^{\mathrm{dyn}}
\right).
}

\localheading{Meaning}

The Dynamics Proxy is the classical form of a World Model. Given the current state, history, and an agent action or future query, it predicts the future state and may also predict reward. It is the proxy in its most literal sense, a learned stand-in for the dynamics of the environment, and the closest of the six forms to the textbook world model.

\begin{table}[H]
\centering
\fontsize{7.3pt}{8.35pt}\selectfont
\setlength{\tabcolsep}{2.6pt}
\setlength{\extrarowheight}{0.6pt}
\renewcommand{\arraystretch}{1.05}
\captionsetup{skip=4pt,font=small}
\newcommand{\proxyhead}[1]{{\sffamily\bfseries\textcolor{brandblue}{#1}}}
\newcommand{\proxyfn}[1]{\textbf{\textcolor{brandblue}{#1}}}
\caption{Functional forms of Agent-Centric World Proxies.}
\label{tab:proxy-functions}
\begin{tabularx}{\columnwidth}{
  >{\raggedright\arraybackslash}p{0.175\columnwidth}
  !{\color{bordergray}\vrule width 0.45pt}
  >{\raggedright\arraybackslash}p{0.17\columnwidth}
  >{\raggedright\arraybackslash}p{0.19\columnwidth}
  >{\raggedright\arraybackslash}p{0.16\columnwidth}
  !{\color{bordergray}\vrule width 0.45pt}
  >{\raggedright\arraybackslash}X
}
\toprule
\proxyhead{Proxy Function} & \proxyhead{Agent Input} & \proxyhead{What it Proxies} & \proxyhead{Proxy Output} & \proxyhead{Typical Examples}\\
\midrule
\proxyfn{Dynamics (World Model)}
& state / history + action / future query
& real-world dynamics and temporal transitions
& future state, rollout, predicted reward
& video prediction, action-conditioned generation, robotics dynamics model, model-based RL\\
\addlinespace[2.2pt]
\proxyfn{Spatial}
& scene context + viewpoint / pose / location query
& spatial observation under alternative viewpoints
& novel observation, rendered view, spatial representation
& NeRF, 3D Gaussian Splatting, novel view synthesis, spatial imagination\\
\addlinespace[2.2pt]
\proxyfn{Execution}
& code / command / click / API call / tool call
& consequences of executable interactions in digital environments
& execution result, state change, stdout / stderr
& browser simulator, GUI simulator, code execution predictor, API response simulator\\
\addlinespace[2.2pt]
\proxyfn{Memory / Experience}
& task context + retrieval query
& reusable past interaction evidence
& retrieved experience, failure cases, constraints
& experience memory, reflection memory, failure memory\\
\addlinespace[2.2pt]
\proxyfn{Skill}
& goal / context + skill query
& reusable action knowledge or behavior priors
& skill suggestion, action prior
& skill library, reusable behavior module\\
\addlinespace[2.2pt]
\proxyfn{Reward / Verification}
& plan / trajectory / answer / action
& evaluation feedback, preferences, criteria
& reward, critique, preference, verification
& reward model, verifier, critic, trajectory evaluator, LLM-as-Judge\\
\bottomrule
\end{tabularx}
\end{table}

Here:

\begin{proxyitems}
  \item $s_\ell$: current state or historical observations;
  \item $u_\ell^{\mathrm{dyn}}$: an action, action sequence, or future query proposed by the Agent;
  \item $\hat{s}_{\ell+1}$: predicted future state;
  \item $\hat{r}_{\ell+1}$: predicted reward (optional).
\end{proxyitems}

When the interaction step $\ell$ aligns with the physical time step $t$, this form reduces to classical next-state prediction:

\[
\hat{s}_{t+1}
=\mathcal{WM}(s_t,a_t).
\]

\proxyheading{Typical Examples}

\begin{proxyitems}
  \item video prediction / future frame prediction~\cite{ho2022videodiffusion,voleti2022mcvd,hoeppe2022ramvid,yan2021videogpt,babaeizadeh2021fitvid,liu2024lwm,wu2024ivideogpt};
  \item action-conditioned video generation~\cite{bruce2024genie,yang2024unisim,brooks2024sora,nvidia2025cosmos,deepmind2025genie3,oasis2024,bai2025recammaster,mei2024dreamforge,yang2026spiral,zhang2025matrixgame,gao2026lingbotworld,wu2024ivideogpt,valevski2025gamengin,yu2025gamefactory};
  \item interactive game-world simulation~\cite{valevski2025gamengin,oasis2024,guo2025mineworld,che2024gamegenx,yu2025gamefactory,deepmind2025genie3};
  \item robotics dynamics prediction~\cite{wu2023daydreamer,hafner2019dreamer,hafner2023dreamerv3,hansen2024tdmpc2,meta2025vjepa2,ha2018worldmodels,janner2019mbpo};
  \item autonomous driving trajectory prediction~\cite{gao2024vista,hu2023gaia1,wang2024drivedreamer,zheng2024occworld,bian2025dynamiccity,yan2025ad-r1,min2024driveworld,zhang2024copilot4d,liang2026lidarcrafter,xu2025u4d,nvidia2025gaia2,yang2025driving,mei2025vision};
  \item model-based reinforcement learning~\cite{ha2018worldmodels,schrittwieser2020muzero,hafner2019dreamer,janner2019mbpo,micheli2023iris,alonso2024diamond,zhang2023storm,hafner2019recurrent,kaiser2019simple}.
\end{proxyitems}

\vspace{0.15cm}
\begin{keypointbox}
\blocktitle{Key Point}
A \textcolor{brandblue}{\textbf{Dynamics Proxy}} focuses on: how the environment would change if the Agent executed a given action.
\end{keypointbox}
\vspace{0.2cm}

\Needspace{6\baselineskip}
\subsection{Spatial Proxy}

\proxyheading{Formula}

\proxyformulabox{
\hat{o}_{\ell+1}^{\mathrm{view}}
=\WP^{\mathrm{spatial}}
\left(
s_\ell,
u_\ell^{\mathrm{spatial}}
\right).
}

\proxyheading{Meaning}

The Spatial Proxy generates observations under spatial or viewpoint conditions. The Agent queries a new location, camera pose, or viewpoint, and the World Proxy returns the corresponding observation or spatial representation. In effect it lets the agent \emph{look before it moves}, turning an expensive physical relocation into a cheap query about what a yet-unseen vantage point would reveal.

Here:

\begin{proxyitems}
  \item $s_\ell$: known visual, geometric, or spatial information;
  \item $u_\ell^{\mathrm{spatial}}$: the queried viewpoint, pose, or spatial position;
  \item $\hat{o}_{\ell+1}^{\mathrm{view}}$: the predicted or rendered observation from that viewpoint.
\end{proxyitems}

\proxyheading{Typical Examples}

\begin{proxyitems}
  \item NeRF-based neural rendering~\cite{mildenhall2020nerf,barron2021mipnerf,yu2021plenoxels,muller2022instantngp,chen2022tensorf,barron2022mipnerf360,tancik2023nerfstudio};
  \item 3D Gaussian Splatting~\cite{kerbl20233dgs,wu20244dgs,lu2023scaffoldgs,yu2023mipsplatting,chen20243dgssurvey,fei20243dgaussianera};
  \item 3D/4D scene reconstruction~\cite{kong2025survey_3d4d,wang2023dust3r,leroy2024mast3r,li2024megasam,zhang2024monst3r,wang2024spann3r,wang2025cut3r,yang2025fast3r,wang2025vggt};
  \item 3D/4D scene generation~\cite{yang2026x,sun2024dimensionx,fridman2023scenescape,hollein2023text2room,zhang2023text2nerf,xie2023citydreamer,chen2023scenedreamer,yu2024wonderworld,li2024dreamscene,worldlabs2025marble};
  \item visual imagination in navigation (Navigation WM~\cite{bar2024navigation,koh2021pathdreamer,dong2026lcvn,hu2026navthinker,dong2026uniwm});
  \item spatial reasoning and manipulation~\cite{zha2025enable,mindcube,zhen2025tesseract,lu2025gwm,chai2025gaf,huang2026pointworld,zhang2025robooccworld,huang2025particleformer,cao2025physxanything,wang2026mvista4d,zeng2020transporternets,shridhar2021cliport,shridhar2022peract,huang2023voxposer,brohan2022rt1,brohan2023rt2}.
\end{proxyitems}

\vspace{0.2cm}
\begin{keypointbox}
\blocktitle{Key Point}
A \textcolor{brandblue}{\textbf{Spatial Proxy}} focuses on: what the Agent would observe from another viewpoint or position.
\end{keypointbox}
\vspace{0.2cm}

\subsection{Execution Proxy}

\proxyheading{Formula}

\proxyformulabox{
\hat{s}_{\ell+1}^{\mathrm{exec}}, \hat{y}_{\ell+1}^{\mathrm{exec}}
=\WP^{\mathrm{exec}}
\left(
s_\ell,
u_\ell^{\mathrm{exec}}
\right).
}

\proxyheading{Meaning}

The Execution Proxy simulates executable interactions. The Agent issues code, a command, a web click, an API call, or a tool call, and the World Proxy predicts the resulting state and feedback. Where a dynamics proxy models continuous physics, an execution proxy models the discrete and often brittle logic of digital systems, in which a single misplaced character or click can flip the outcome entirely.

Here:

\begin{proxyitems}
  \item $s_\ell$: current web, code, file, program, or tool state;
  \item $u_\ell^{\mathrm{exec}}$: the executable interaction issued by the Agent;
  \item $\hat{s}_{\ell+1}^{\mathrm{exec}}$: predicted post-execution state;
  \item $\hat{y}_{\ell+1}^{\mathrm{exec}}$: predicted feedback, such as stdout, stderr, an error message, a test result, or a page change.
\end{proxyitems}

\proxyheading{Typical Examples}

\begin{proxyitems}
  \item Browser / Web interaction simulator: simulates page states and feedback after clicks, inputs, or navigation~\cite{gu2024webdreamer,chae2025wma,xiao2026webworld,feng2025wwm,gao2025websynthesis,wac2026};
  \item GUI environment simulator: predicts interface changes after button clicks, window switching, or form filling~\cite{luo2025vimo,cao2026mobiledreamer,cuwm2026,zheng2026code2world,li2025mobileworldbench,gworld2026,ai2026comact,hu2026itercad};
  \item Code execution predictor: predicts stdout, error messages, or test results after running code~\cite{tang2024worldcoder,dainese2024codewm,copet2025cwm,maimon2026selfexecution,rahmani2026debuggingcwm,butt2024codeit};
  \item Shell / file-system transition model: predicts file states, stdout, or stderr after command execution~\cite{rivard2025neuralos,cuwm2026,tang2024worldcoder,dainese2024codewm,copet2025cwm,maimon2026selfexecution,rahmani2026debuggingcwm,xie2024osworld};
  \item API / tool-call response simulator: predicts the returned result after an API or tool call~\cite{ren2025gtm,ganapavarapu2026mcpcosmos,guo2025dymo,lu2024toolsandbox,hu2026occubench,schick2023toolformer,hao2023reasoning}.
\end{proxyitems}

\vspace{0.2cm}
\begin{keypointbox}
\blocktitle{Key Point}
An \textcolor{brandblue}{\textbf{Execution Proxy}} focuses on: what the digital environment would return after the Agent performs an operation.
\end{keypointbox}
\vspace{0.2cm}

\subsection{Memory / Experience Proxy}

\proxyheading{Formula}

\proxyformulabox{
\hat{m}_{\ell+1}
=\WP^{\mathrm{mem}}
\left(
s_\ell,
u_\ell^{\mathrm{mem}}
\right).
}

\proxyheading{Meaning}

The Memory / Experience Proxy retrieves task-relevant information from past interactions, trajectories, failures, or constraints. Its output is not a physical state, but experiential feedback for planning and decision making.

Here:

\begin{proxyitems}
  \item $s_\ell$: current task context, Agent memory, or environment information;
  \item $u_\ell^{\mathrm{mem}}$: the retrieval query issued by the Agent;
  \item $\hat{m}_{\ell+1}$: retrieved experience, failure case, constraint, or risk hint.
\end{proxyitems}

An ordinary static memory store is not necessarily a World Proxy. It becomes a Memory / Experience Proxy only when a \textbf{store-retrieve dynamic system} returns environment-, task-, or decision-relevant information in response to the Agent's query and feeds it back into planning, decision making, or improvement. A generative agent recalling the most relevant slices of its past before it acts, or a reflective agent remembering exactly why its last attempt failed, both fall under this view.

\proxyheading{Typical Examples}

\begin{proxyitems}
  \item Experience-based world memory: retrieves reusable experience from successful or failed trajectories~\cite{park2023generative,chen2025dreamgym,qiao2024wkm,shinn2023reflexion,sumers2024coala,packer2023memgpt,wu2026memharness};
  \item Constraint / failure memory: retrieves constraints, risks, and avoidance strategies from past failures~\cite{ren2026aligning,shinn2023reflexion,butt2024codeit,chen2025dreamgym,wang2025ragen,webevolver2025};
  \item Reflection memory model: returns prior mistakes, corrective feedback, or improvement hints~\cite{shinn2023reflexion,sumers2024coala,park2023generative,packer2023memgpt,ren2026aligning,qiao2024wkm}.
\end{proxyitems}

\vspace{0.2cm}
\begin{keypointbox}
\blocktitle{Key Point}
A \textcolor{brandblue}{\textbf{Memory / Experience Proxy}} focuses on: the prior experience, failure cases, or constraints the Agent needs for the current decision.
\end{keypointbox}
\vspace{0.05cm}

\subsection{Skill Proxy}
\enlargethispage{0.5\baselineskip}

\proxyheading{Formula}
\vspace{-0.25\baselineskip}

\proxyformulabox{
\hat{g}_{\ell+1}^{\mathrm{skill}}
=\WP^{\mathrm{skill}}
\left(
s_\ell,
u_\ell^{\mathrm{skill}}
\right).
}
\vspace{-0.18\baselineskip}

\proxyheading{Meaning}

The Skill Proxy retrieves or recommends reusable behavioral knowledge. While the Memory / Experience Proxy emphasizes what has happened before, the Skill Proxy emphasizes what the Agent can do now.

It recommends reusable skills, tool-use routines, or action priors based on the current goal, task context, and environment state. An open-world agent, for instance, can bank a routine it once worked out, crafting a particular tool or completing a multi-step form, and later summon it whole rather than rediscovering it move by move.

Here:

\begin{proxyitems}
  \item $s_\ell$: current task, goal, environment state, or Agent context;
  \item $u_\ell^{\mathrm{skill}}$: the skill or policy query issued by the Agent;
  \item $\hat{g}_{\ell+1}^{\mathrm{skill}}$: a retrieved skill, tool-use routine, reusable behavior module, or action prior.
\end{proxyitems}

\proxyheading{Typical Examples}

\begin{proxyitems}
  \item Skill library: retrieves executable skills for the current task~\cite{wang2023voyager,huang2025cascade,ahn2022saycan,liang2022codeaspolicies,huang2022innermonologue,butt2024codeit};
  \item Tool-use routine: retrieves tool-calling workflows or operation templates~\cite{ren2025gtm,schick2023toolformer,ganapavarapu2026mcpcosmos,guo2025dymo,hao2023reasoning,liu2023agentbench};
  \item Reusable behavior module: provides a reusable behavior strategy~\cite{wang2023voyager,huang2025cascade,ahn2022saycan,liang2022codeaspolicies,huang2022innermonologue,tao2024maniskill3}.
\end{proxyitems}

\vspace{0.05cm}
\begin{keypointbox}
\blocktitle{Key Point}
A \textcolor{brandblue}{\textbf{Skill Proxy}} focuses on: the reusable skills or behavior strategies available for the current task.
\end{keypointbox}
\vspace{0.2cm}

\subsection{Reward / Verification Proxy}

\proxyheading{Formula}

\proxyformulabox{
\hat{v}_{\ell+1}^{\mathrm{eval}}
=\WP^{\mathrm{eval}}
\left(
s_\ell,
u_\ell^{\mathrm{eval}}
\right).
}

\proxyheading{Meaning}

The Reward / Verification Proxy evaluates an Agent's behavior, trajectory, answer, or plan. It acts as a reward model, verifier, critic, or evaluator. Rather than telling the agent what the world will look like next, it tells the agent how good its behavior is, collapsing an entire rollout into a single, actionable verdict.

Here:

\begin{proxyitems}
  \item $s_\ell$: current task, context, environment information, or historical trajectory;
  \item $u_\ell^{\mathrm{eval}}$: the plan, trajectory, answer, or action submitted by the Agent;
  \item $\hat{v}_{\ell+1}^{\mathrm{eval}}$: predicted reward, verification result, critique, preference, or failure reason.
\end{proxyitems}

This is a \textbf{feedback-oriented World Proxy}: it does not need to simulate the full environment, but predicts how the Agent's behavior would be judged by the environment, rules, or evaluation system.

\proxyheading{Typical Examples}

\begin{proxyitems}
  \item reward model~\cite{christiano2017deeprlpreferences,ziegler2019finetuning,stiennon2020summarize,ouyang2022instructgpt,bai2022hh,bai2022constitutional};
  \item preference model~\cite{rafailov2023dpo,azar2023ipo,yuan2023rrhf,ethayarajh2024kto,hong2024orpo,meng2024simpo,bai2022constitutional};
  \item verifier~\cite{cobbe2021verifiers,uesato2022process,lightman2023letsverify,wang2023mathshepherd,shao2024deepseekmath,mcaleese2024criticgpt};
  \item critic model~\cite{mcaleese2024criticgpt,madaan2023selfrefine,shinn2023reflexion,zhang2025critic,bai2022constitutional,liu2023geval};
  \item trajectory evaluator~\cite{fan2025worldmodelbench,qin2024worldsimbench,worldlens,zhou2023webarena,xie2024osworld,liu2023agentbench,vafa2024evaluating};
  \item LLM-as-Judge~\cite{zheng2023judge,liu2023geval,kim2023prometheus,dubois2024alpacaeval,gu2024judge,lin2025computer}.
\end{proxyitems}

\vspace{0.2cm}
\begin{keypointbox}
\blocktitle{Key Point}
A \textcolor{brandblue}{\textbf{Reward / Verification Proxy}} focuses on: whether the Agent's behavior is correct, safe, and feasible, and how it should improve.
\end{keypointbox}
\vspace{0.2cm}

\subsection{Putting It Together: Functions Meet Levels}

The two axes of this article are orthogonal: any proxy \emph{function} (Section~\ref{sec:instantiations}) can empower an agent at any \emph{level} (Section~\ref{sec:empowerment}). Reading the six functions against \LoneInline{}-\LthreeInline{} turns the design space into a simple map, with representative systems in each cell (Table~\ref{tab:functions-x-levels}). The mapping is illustrative rather than exclusive, since many systems span more than one level, and the sparser cells mark territory that remains largely open.

\begin{table}[t]
\centering
\fontsize{7.5pt}{8.15pt}\selectfont
\setlength{\tabcolsep}{2.2pt}
\setlength{\extrarowheight}{0.4pt}
\renewcommand{\arraystretch}{0.98}
\captionsetup{skip=2pt,font=small}
\newcommand{\fxlhead}[2]{{\sffamily\bfseries #1~{\scriptsize #2}}}
\newcommand{\fxlfn}[1]{\textbf{\textcolor{brandblue}{#1}}}
\caption{\textbf{Functions $\times$ Levels.} Representative ways each proxy function empowers agents across L1, L2, and L3. Sparse cells flag underexplored directions.}
\label{tab:functions-x-levels}
\begin{tabularx}{\columnwidth}{
  >{\raggedright\arraybackslash}p{0.135\columnwidth}
  !{\color{bordergray}\vrule width 0.45pt}
  >{\raggedright\arraybackslash}X
  >{\raggedright\arraybackslash}X
  >{\raggedright\arraybackslash}X
}
\toprule
\fxlhead{Function}{}
& \fxlhead{\LoneBadge}{\LoneText{Inference-Time Guidance}}
& \fxlhead{\LtwoBadge}{\LtwoText{Training-Time Optimization}}
& \fxlhead{\LthreeBadge}{\LthreeText{Agent-Proxy Co-Evolution}}\\
\midrule
\fxlfn{Dynamics}
& plan via imagined rollouts~\cite{schrittwieser2020muzero,hao2023reasoning}
& learn a policy in imagination~\cite{hafner2023dreamerv3,janner2019mbpo}
& online model learning on a real robot~\cite{wu2023daydreamer}\\
\addlinespace[1.5pt]
\fxlfn{Spatial}
& imagine views to navigate~\cite{bar2024navigation,koh2021pathdreamer}
& train on synthesized observations~\cite{zhu2025aether}
& \textit{underexplored}\\
\addlinespace[1.5pt]
\fxlfn{Execution}
& simulate actions before acting~\cite{gu2024webdreamer,chae2025wma}
& synthesize experience for RL~\cite{chen2025dreamgym}
& self-improve via a coevolving model~\cite{webevolver2025}\\
\addlinespace[1.5pt]
\fxlfn{Memory}
& retrieve experience at decision time~\cite{park2023generative,sumers2024coala}
& learn from past successes and failures~\cite{ren2026aligning,wu2026memharness}
& memory updated, then distilled back~\cite{butt2024codeit}\\
\addlinespace[1.5pt]
\fxlfn{Skill}
& reuse skills from a library~\cite{wang2023voyager}
& create and train new skills~\cite{huang2025cascade}
& skill library grows with the agent~\cite{wang2023voyager,huang2025cascade}\\
\addlinespace[1.5pt]
\fxlfn{Reward / Verif.}
& critique a plan at inference~\cite{zhang2025critic,lin2025computer}
& reward / verifier signals for RL~\cite{wang2025ragen,wang2025vagen}
& \textit{emerging}\\
\bottomrule
\end{tabularx}
\end{table}

Two patterns stand out. Reading \emph{down} a column shows that a single level admits many functional realizations; reading \emph{across} a row shows that the same function can graduate from advisor to teacher to partner as it climbs \LoneInline{} to \LthreeInline{}. The blank corners, spatial and reward proxies at the co-evolution level, are not accidents but invitations.

\vspace{0.45\baselineskip}
\begin{highlightcard}{\cardicon{figs/idea.png} Takeaway}
\cardstrong{Different proxy functions, one purpose: \cardkey{agent-usable feedback}.}

\begin{itemize}[leftmargin=1.35em,itemsep=3pt,topsep=4pt,parsep=0pt,partopsep=0pt,label=\textcolor{brandblue}{\textbullet}]
  \item \cardstrong{Beyond future-state prediction:}
  a World Proxy may predict dynamics, render spatial observations, simulate execution, retrieve experience, suggest skills, or provide reward / verification feedback.

  \item \cardstrong{Function defines feedback type:}
  dynamics, spatial, execution, memory, skill, and verification proxies differ in what \cardkey{information transition} they approximate for the Agent.

  \item \cardstrong{Actionability is the common criterion:}
  regardless of function, the output should help the Agent plan, decide, learn, verify, or \cardkey{continually improve}.
\end{itemize}
\end{highlightcard}

%% file: sections/conclusion.tex
\vspace{0.4cm}
\section{Conclusion: Quo Vadis, World Modeling?}
\label{sec:conclusion}

We opened with a question, so let us close with an answer. For most of its history, world modeling has been pursued as the art of \textbf{predicting the world}: given a state and an action, render the next frame as faithfully as possible~\cite{ding2024survey_wm}.

This article has argued for a quieter but consequential shift, from predicting the world to \textbf{serving the agent}. Once the goal becomes continual improvement, the right object is no longer a state predictor but an \textbf{Agent-Centric World Proxy}: an environment-grounded mechanism that returns the information transition an agent needs, whether a future state, rendered view, execution result, retrieved memory or skill, or verdict on a plan.

That reframing organized the rest of the story. We saw \emph{why} real environments alone cannot carry continual improvement, \emph{what} changes when physical state transitions become interactive information transitions, \emph{how} proxies empower agents across \LoneBadge~\LoneText{inference-time guidance}, \LtwoBadge~\LtwoText{training-time optimization}, and \LthreeBadge~\LthreeText{Agent-Proxy co-evolution}, and \emph{in what forms} they appear: dynamics, spatial, execution, memory, skill, and reward / verification. The unifying thread is not visual realism for its own sake, but \textbf{actionable information gain}.

The same reframing also sharpens the open problems. A proxy is only as useful as it is trustworthy, and trust remains the hardest part to guarantee.

\vspace{0.2cm}
\vspace{0.25\baselineskip}
\begin{highlightcard}{\cardicon{figs/idea.png} Open Challenges}
\begin{enumerate}[leftmargin=1.45em,itemsep=4pt,topsep=3pt,parsep=0pt,partopsep=0pt,label=\textcolor{brandblue}{\textbf{\arabic*.}}]
  \item \cardstrong{Fidelity and the limits of imagination.} Generative models can look convincing while violating the dynamics they claim to model~\cite{vafa2024evaluating,kang2025howfar,guo2026physcond,jing2026counterscene}. Errors compound over long rollouts; \cardkey{calibrated uncertainty}, not just sharper pixels, is the missing ingredient.
  \item \cardstrong{Knowing when to trust the proxy.} An agent must decide, online, whether to act on proxy feedback or return to the real environment. Today agents rarely make that judgment well~\cite{qian2026foresight}; treating the proxy as an oracle invites silent failure.
  \item \cardstrong{Reward hacking and safety.} When the proxy becomes the reward or verifier (\LtwoInline{}), the agent is incentivized to exploit blind spots. The same sandbox that makes risky exploration safe also opens a new attack surface~\cite{zeng2024wmsafety,li2026embodiedsafety,xu2026ctrlattack,liu2026jailwam,yan2026safedream}.
  \item \cardstrong{Evaluation that measures information gain.} Current benchmarks score realism, fidelity, controllability, or human-aligned quality~\cite{fan2025worldmodelbench,qin2024worldsimbench,worldlens}, but still grade the proxy \emph{in isolation}. We need \cardkey{agent-centric benchmarks}: did the feedback help the agent plan, learn, or improve?
\end{enumerate}
\end{highlightcard}
\vspace{0.3cm}

\vspace{0.35\baselineskip}
None of these are reasons for pessimism; they are the agenda. Each becomes tractable once we stop asking a world model to be a perfect mirror and start asking it to be a useful interlocutor, one whose answers are grounded, calibrated, and continually corrected by contact with reality. The \LthreeInline{} loop, where real-environment evidence keeps the proxy honest and the proxy keeps the agent improving, is as much a safety mechanism as a learning one.

So, \emph{quo vadis}? We expect the most valuable world models of the coming years to be judged less by how vividly they dream and more by how much better they make the agents that query them. If this article nudges the conversation from \textbf{building world simulators} toward \textbf{building world proxies that agents can learn from}, it will have served its purpose.

%% file: sections/contributors.tex
\vspace{0.3cm}
\section{List of Contributors}
\label{sec:contributors}

\newcommand{\contribnames}[1]{\textit{\textcolor{textgray}{#1}}}

\begin{itemize}[leftmargin=1.35em,itemsep=3pt,topsep=4pt,parsep=0pt,partopsep=0pt,label=\textcolor{brandblue}{\textbullet}]
  \item \cardstrong{Concept \& Design:}\\ \contribnames{Yu Yang, Xuemeng Yang, Licheng Wen}
  \item \cardstrong{Writing \& Editing:}\\ \contribnames{Yu Yang, Xuemeng Yang, Licheng Wen, Lingdong Kong, Xiaobin Hu, Dongyue Lu, Wei Chow}
  \item \cardstrong{Figures \& Visual Design:}\\ \contribnames{Xiyan Huang, Yuxiang Feng}
  \item \cardstrong{Discussion \& Insights:}\\ \contribnames{Yue Liao, Jianbiao Mei, Daocheng Fu, Rong Wu, Pinlong Cai, Ran Yi, Ying Tai, Jiangning Zhang}
  \item \cardstrong{Advising:}\\ \contribnames{Botian Shi, Yong Liu, Shuicheng Yan}
\end{itemize}
\vspace{0.5cm}

%% file: main.bbl
\begin{thebibliography}{216}
\providecommand{\natexlab}[1]{#1}
\providecommand{\url}[1]{\texttt{#1}}
\expandafter\ifx\csname urlstyle\endcsname\relax
  \providecommand{\doi}[1]{doi: #1}\else
  \providecommand{\doi}{doi: \begingroup \urlstyle{rm}\Url}\fi

\bibitem[Agarwal et~al.(2025)Agarwal, Ali, Bala, Balaji, et~al.]{nvidia2025cosmos}
Niket Agarwal, Arslan Ali, Maciej Bala, Yogesh Balaji, et~al.
\newblock Cosmos world foundation model platform for physical {AI}.
\newblock \emph{arXiv preprint arXiv:2501.03575}, 2025.
\newblock URL \url{https://arxiv.org/abs/2501.03575}.

\bibitem[Ahn et~al.(2022)Ahn, Brohan, Brown, Chebotar, Cortes, David, Finn, Fu, et~al.]{ahn2022saycan}
Michael Ahn, Anthony Brohan, Noah Brown, Yevgen Chebotar, Omar Cortes, Byron David, Chelsea Finn, Chuyuan Fu, et~al.
\newblock Do as i can, not as i say: Grounding language in robotic affordances.
\newblock In \emph{Conf. Robot Learn.}, 2022.
\newblock URL \url{https://arxiv.org/abs/2204.01691}.

\bibitem[Ai et~al.(2026)Ai, Hu, Yang, Zou, Zhang, Fu, Yang, Zhou, Deng, Cai, et~al.]{ai2026comact}
Jiaxin Ai, Tao Hu, Xuemeng Yang, Shu Zou, Hairong Zhang, Daocheng Fu, Yu~Yang, Hongbin Zhou, Nianchen Deng, Pinlong Cai, et~al.
\newblock {ComAct}: Reframing professional software manipulation via {COM-as-Action} paradigm.
\newblock \emph{arXiv preprint arXiv:2606.13239}, 2026.
\newblock URL \url{https://arxiv.org/abs/2606.13239}.

\bibitem[Alonso et~al.(2024)Alonso, Jelley, Micheli, Kanervisto, Storkey, Pearce, and Fleuret]{alonso2024diamond}
Eloi Alonso, Adam Jelley, Vincent Micheli, Anssi Kanervisto, Amos Storkey, Tim Pearce, and Fran{\c{c}}ois Fleuret.
\newblock Diffusion for world modeling: Visual details matter in {Atari}.
\newblock In \emph{Adv. Neural Inf. Process. Syst.}, volume~37, 2024.
\newblock URL \url{https://arxiv.org/abs/2405.12399}.

\bibitem[Assran et~al.(2023)Assran, Duval, Misra, Bojanowski, Vincent, Rabbat, LeCun, and Ballas]{assran2023ijepa}
Mahmoud Assran, Quentin Duval, Ishan Misra, Piotr Bojanowski, Pascal Vincent, Michael Rabbat, Yann LeCun, and Nicolas Ballas.
\newblock Self-supervised learning from images with a joint-embedding predictive architecture.
\newblock In \emph{IEEE/CVF Conf. Comput. Vis. Pattern Recog.}, 2023.
\newblock URL \url{https://arxiv.org/abs/2301.08243}.

\bibitem[Assran et~al.(2025)Assran, Bardes, Fan, Garrido, Howes, Muckley, Rizvi, Roberts, Sinha, et~al.]{meta2025vjepa2}
Mido Assran, Adrien Bardes, David Fan, Quentin Garrido, Russell Howes, Matthew Muckley, Ammar Rizvi, Claire Roberts, Koustuv Sinha, et~al.
\newblock {V-JEPA 2}: Self-supervised video models enable understanding, prediction and planning.
\newblock \emph{arXiv preprint arXiv:2506.09985}, 2025.
\newblock URL \url{https://arxiv.org/abs/2506.09985}.

\bibitem[Azar et~al.(2024)Azar, Rowland, Piot, Guo, Calandriello, Valko, and Munos]{azar2023ipo}
Mohammad~Gheshlaghi Azar, Mark Rowland, Bilal Piot, Daniel Guo, Daniele Calandriello, Michal Valko, and R{\'e}mi Munos.
\newblock A general theoretical paradigm to understand learning from human preferences.
\newblock In \emph{Int. Conf. Artif. Intell. Stat.}, 2024.
\newblock URL \url{https://arxiv.org/abs/2310.12036}.

\bibitem[Babaeizadeh et~al.(2021)Babaeizadeh, Saffar, Nair, Levine, Finn, and Erhan]{babaeizadeh2021fitvid}
Mohammad Babaeizadeh, Mohammad~Taghi Saffar, Suraj Nair, Sergey Levine, Chelsea Finn, and Dumitru Erhan.
\newblock {FitVid}: Overfitting in pixel-level video prediction.
\newblock \emph{arXiv preprint arXiv:2106.13195}, 2021.
\newblock URL \url{https://arxiv.org/abs/2106.13195}.

\bibitem[Bai et~al.(2025)Bai, Xia, Fu, Wang, Mu, Cao, Liu, Hu, Bai, Wan, and Zhang]{bai2025recammaster}
Jianhong Bai, Menghan Xia, Xiao Fu, Xintao Wang, Lianrui Mu, Jinwen Cao, Zuozhu Liu, Haoji Hu, Xiang Bai, Pengfei Wan, and Di~Zhang.
\newblock {ReCamMaster}: Camera-controlled generative rendering from a single video.
\newblock \emph{arXiv preprint arXiv:2503.11647}, 2025.
\newblock URL \url{https://arxiv.org/abs/2503.11647}.

\bibitem[Bai et~al.(2022{\natexlab{a}})Bai, Jones, Ndousse, Askell, Chen, DasSarma, Drain, Fort, et~al.]{bai2022hh}
Yuntao Bai, Andy Jones, Kamal Ndousse, Amanda Askell, Anna Chen, Nova DasSarma, Dawn Drain, Stanislav Fort, et~al.
\newblock Training a helpful and harmless assistant with reinforcement learning from human feedback.
\newblock \emph{arXiv preprint arXiv:2204.05862}, 2022{\natexlab{a}}.
\newblock URL \url{https://arxiv.org/abs/2204.05862}.

\bibitem[Bai et~al.(2022{\natexlab{b}})Bai, Kadavath, Kundu, Askell, Kernion, Jones, Chen, Goldie, et~al.]{bai2022constitutional}
Yuntao Bai, Saurav Kadavath, Sandipan Kundu, Amanda Askell, Jackson Kernion, Andy Jones, Anna Chen, Anna Goldie, et~al.
\newblock Constitutional {AI}: Harmlessness from {AI} feedback.
\newblock \emph{arXiv preprint arXiv:2212.08073}, 2022{\natexlab{b}}.
\newblock URL \url{https://arxiv.org/abs/2212.08073}.

\bibitem[Bar et~al.(2025)Bar, Zhou, Tran, Darrell, and LeCun]{bar2024navigation}
Amir Bar, Gaoyue Zhou, Danny Tran, Trevor Darrell, and Yann LeCun.
\newblock Navigation world models.
\newblock In \emph{IEEE/CVF Conf. Comput. Vis. Pattern Recog.}, pages 15791--15801, 2025.
\newblock URL \url{https://arxiv.org/abs/2412.03572}.

\bibitem[Barron et~al.(2021)Barron, Mildenhall, Tancik, Hedman, Martin-Brualla, and Srinivasan]{barron2021mipnerf}
Jonathan~T. Barron, Ben Mildenhall, Matthew Tancik, Peter Hedman, Ricardo Martin-Brualla, and Pratul~P. Srinivasan.
\newblock {Mip-NeRF}: A multiscale representation for anti-aliasing neural radiance fields.
\newblock In \emph{IEEE/CVF Int. Conf. Comput. Vis.}, pages 5835--5844, 2021.
\newblock URL \url{https://arxiv.org/abs/2103.13415}.

\bibitem[Barron et~al.(2022)Barron, Mildenhall, Verbin, Srinivasan, and Hedman]{barron2022mipnerf360}
Jonathan~T. Barron, Ben Mildenhall, Dor Verbin, Pratul~P. Srinivasan, and Peter Hedman.
\newblock {Mip-NeRF 360}: Unbounded anti-aliased neural radiance fields.
\newblock In \emph{IEEE/CVF Conf. Comput. Vis. Pattern Recog.}, pages 5470--5479, 2022.
\newblock URL \url{https://arxiv.org/abs/2111.12077}.

\bibitem[Bian et~al.(2025)Bian, Kong, Xie, Pan, Qiao, and Liu]{bian2025dynamiccity}
Hengwei Bian, Lingdong Kong, Haozhe Xie, Liang Pan, Yu~Qiao, and Ziwei Liu.
\newblock {DynamicCity}: Large-scale {4D} occupancy generation from dynamic scenes.
\newblock In \emph{Int. Conf. Learn. Represent.}, 2025.
\newblock URL \url{https://arxiv.org/abs/2410.18084}.

\bibitem[Brohan et~al.(2023{\natexlab{a}})Brohan, Brown, Carbajal, Chebotar, Chen, Choromanski, Ding, Driess, Dubey, Finn, Florence, Fu, Hausman, Herzog, Hsu, Ichter, Levine, Lu, Mordatch, Sermanet, Xiao, Xu, Yu, Zitkovich, et~al.]{brohan2023rt2}
Anthony Brohan, Noah Brown, Justice Carbajal, Yevgen Chebotar, Xi~Chen, Krzysztof Choromanski, Tianli Ding, Danny Driess, Avinava Dubey, Chelsea Finn, Pete Florence, Chuyuan Fu, Karol Hausman, Alex Herzog, Jasmine Hsu, Brian Ichter, Sergey Levine, Yao Lu, Igor Mordatch, Pierre Sermanet, Ted Xiao, Peng Xu, Tianhe Yu, Brianna Zitkovich, et~al.
\newblock {RT-2}: Vision-language-action models transfer web knowledge to robotic control.
\newblock In \emph{Conf. Robot Learn.}, 2023{\natexlab{a}}.
\newblock URL \url{https://arxiv.org/abs/2307.15818}.

\bibitem[Brohan et~al.(2023{\natexlab{b}})Brohan, Brown, Carbajal, Chebotar, Dabis, Finn, Gopalakrishnan, Hausman, Herzog, Hsu, Ichter, Levine, Lu, Mordatch, Nachum, Parada, Sermanet, Xiao, Xu, Yu, Zitkovich, et~al.]{brohan2022rt1}
Anthony Brohan, Noah Brown, Justice Carbajal, Yevgen Chebotar, Joseph Dabis, Chelsea Finn, Keerthana Gopalakrishnan, Karol Hausman, Alex Herzog, Jasmine Hsu, Brian Ichter, Sergey Levine, Yao Lu, Igor Mordatch, Ofir Nachum, Carolina Parada, Pierre Sermanet, Ted Xiao, Peng Xu, Tianhe Yu, Brianna Zitkovich, et~al.
\newblock {RT-1}: Robotics transformer for real-world control at scale.
\newblock In \emph{Robot. Sci. Syst.}, 2023{\natexlab{b}}.
\newblock URL \url{https://arxiv.org/abs/2212.06817}.

\bibitem[Brooks et~al.(2024)Brooks, Peebles, Holmes, DePue, Guo, Jing, Schnurr, Taylor, et~al.]{brooks2024sora}
Tim Brooks, Bill Peebles, Connor Holmes, Will DePue, Yufei Guo, Li~Jing, David Schnurr, Joe Taylor, et~al.
\newblock Video generation models as world simulators.
\newblock Technical report, OpenAI, 2024.
\newblock URL \url{https://openai.com/research/video-generation-models-as-world-simulators}.

\bibitem[Bruce et~al.(2024)Bruce, Dennis, Edwards, Parker-Holder, Shi, Hughes, Lai, Mavalankar, et~al.]{bruce2024genie}
Jake Bruce, Michael Dennis, Ashley Edwards, Jack Parker-Holder, Yuge Shi, Edward Hughes, Matthew Lai, Aditi Mavalankar, et~al.
\newblock Genie: Generative interactive environments.
\newblock In \emph{Int. Conf. Mach. Learn.}, pages 4603--4623, 2024.
\newblock URL \url{https://arxiv.org/abs/2402.15391}.

\bibitem[Butt et~al.(2024)Butt, Manczak, Wiggers, Rainone, Zhang, Defferrard, and Cohen]{butt2024codeit}
Natasha Butt, Blazej Manczak, Auke Wiggers, Corrado Rainone, David~W. Zhang, Micha{\"e}l Defferrard, and Taco Cohen.
\newblock {CodeIt}: Self-improving language models with prioritized hindsight replay.
\newblock In \emph{Int. Conf. Mach. Learn.}, pages 5013--5034, 2024.
\newblock URL \url{https://arxiv.org/abs/2402.04858}.

\bibitem[Cao et~al.(2026)Cao, Zhong, Zeng, Zheng, Huang, Qiu, Shi, Mao, and Wan]{cao2026mobiledreamer}
Yilin Cao, Yufeng Zhong, Zhixiong Zeng, Liming Zheng, Jing Huang, Haibo Qiu, Peng Shi, Wenji Mao, and Guanglu Wan.
\newblock {MobileDreamer}: Generative sketch world model for {GUI} agent.
\newblock \emph{arXiv preprint arXiv:2601.04035}, 2026.
\newblock URL \url{https://arxiv.org/abs/2601.04035}.

\bibitem[Cao et~al.(2025)Cao, Hong, Chen, Pan, and Liu]{cao2025physxanything}
Ziang Cao, Fangzhou Hong, Zhaoxi Chen, Liang Pan, and Ziwei Liu.
\newblock {PhysX-Anything}: Simulation-ready physical {3D} assets from single image.
\newblock \emph{arXiv preprint arXiv:2511.13648}, 2025.
\newblock URL \url{https://arxiv.org/abs/2511.13648}.

\bibitem[Chae et~al.(2025)Chae, Kim, Ong, Gwak, Song, Kim, Kim, Lee, and Yeo]{chae2025wma}
Hyungjoo Chae, Namyoung Kim, Kai Tzu-iunn Ong, Minju Gwak, Gwanwoo Song, Jihoon Kim, Sunghwan Kim, Dongha Lee, and Jinyoung Yeo.
\newblock Web agents with world models: Learning and leveraging environment dynamics in web navigation.
\newblock In \emph{Int. Conf. Learn. Represent.}, 2025.
\newblock URL \url{https://arxiv.org/abs/2410.13232}.

\bibitem[Chai et~al.(2025)Chai, Deng, Shao, Zhang, Lv, Xing, Li, Zhang, and Liu]{chai2025gaf}
Ying Chai, Litao Deng, Ruizhi Shao, Jiajun Zhang, Kangchen Lv, Liangjun Xing, Xiang Li, Hongwen Zhang, and Yebin Liu.
\newblock {GAF}: Gaussian action field as a {4D} representation for dynamic world modeling in robotic manipulation.
\newblock \emph{arXiv preprint arXiv:2506.14135}, 2025.
\newblock URL \url{https://arxiv.org/abs/2506.14135}.

\bibitem[Che et~al.(2025)Che, He, Liu, Jin, and Chen]{che2024gamegenx}
Haoxuan Che, Xuanhua He, Quande Liu, Cheng Jin, and Hao Chen.
\newblock {GameGen-X}: Interactive open-world game video generation.
\newblock In \emph{Int. Conf. Learn. Represent.}, 2025.
\newblock URL \url{https://arxiv.org/abs/2411.00769}.

\bibitem[Chen et~al.(2022)Chen, Xu, Geiger, Yu, and Su]{chen2022tensorf}
Anpei Chen, Zexiang Xu, Andreas Geiger, Jingyi Yu, and Hao Su.
\newblock {TensoRF}: Tensorial radiance fields.
\newblock In \emph{Eur. Conf. Comput. Vis.}, pages 333--350, 2022.
\newblock URL \url{https://arxiv.org/abs/2203.09517}.

\bibitem[Chen and Wang(2024)]{chen20243dgssurvey}
Guikun Chen and Wenguan Wang.
\newblock A survey on {3D} gaussian splatting.
\newblock \emph{ACM Comput. Surv.}, 2024.
\newblock URL \url{https://arxiv.org/abs/2401.03890}.

\bibitem[Chen et~al.(2025)Chen, Zhao, Zhang, Liu, Qi, Wu, Kalluri, Cao, et~al.]{chen2025dreamgym}
Zhaorun Chen, Zhuokai Zhao, Kai Zhang, Bo~Liu, Qi~Qi, Yifan Wu, Tarun Kalluri, Sara Cao, et~al.
\newblock Scaling agent learning via experience synthesis.
\newblock \emph{arXiv preprint arXiv:2511.03773}, 2025.
\newblock URL \url{https://arxiv.org/abs/2511.03773}.

\bibitem[Chen et~al.(2023)Chen, Wang, and Liu]{chen2023scenedreamer}
Zhaoxi Chen, Guangcong Wang, and Ziwei Liu.
\newblock {SceneDreamer}: Unbounded {3D} scene generation from {2D} image collections.
\newblock \emph{IEEE Trans. Pattern Anal. Mach. Intell.}, 2023.
\newblock URL \url{https://arxiv.org/abs/2302.01330}.

\bibitem[Christiano et~al.(2017)Christiano, Leike, Brown, Martic, Legg, and Amodei]{christiano2017deeprlpreferences}
Paul~F. Christiano, Jan Leike, Tom~B. Brown, Miljan Martic, Shane Legg, and Dario Amodei.
\newblock Deep reinforcement learning from human preferences.
\newblock In \emph{Adv. Neural Inf. Process. Syst.}, 2017.
\newblock URL \url{https://arxiv.org/abs/1706.03741}.

\bibitem[Chu et~al.(2026)Chu, Zhang, Lin, Kong, Zhang, Tu, Ma, Huang, et~al.]{chu2026agentic}
Meng Chu, Xuan~Billy Zhang, Kevin~Qinghong Lin, Lingdong Kong, Jize Zhang, Teng Tu, Weijian Ma, Ziqi Huang, et~al.
\newblock Agentic world modeling: Foundations, capabilities, laws, and beyond.
\newblock \emph{arXiv preprint arXiv:2604.22748}, 2026.
\newblock URL \url{https://arxiv.org/abs/2604.22748}.

\bibitem[Cobbe et~al.(2021)Cobbe, Kosaraju, Bavarian, Chen, Jun, Kaiser, Plappert, Tworek, Hilton, Nakano, Hesse, and Schulman]{cobbe2021verifiers}
Karl Cobbe, Vineet Kosaraju, Mohammad Bavarian, Mark Chen, Heewoo Jun, Lukasz Kaiser, Matthias Plappert, Jerry Tworek, Jacob Hilton, Reiichiro Nakano, Christopher Hesse, and John Schulman.
\newblock Training verifiers to solve math word problems.
\newblock \emph{arXiv preprint arXiv:2110.14168}, 2021.
\newblock URL \url{https://arxiv.org/abs/2110.14168}.

\bibitem[Copet et~al.(2025)Copet, Carbonneaux, Cohen, Gehring, Kahn, Kossen, Kreuk, McMilin, Meyer, Wei, Zhang, Zheng, Armengol-Estap{\'e}, Bashiri, Beck, et~al.]{copet2025cwm}
Jade Copet, Quentin Carbonneaux, Gal Cohen, Jonas Gehring, Jacob Kahn, Jannik Kossen, Felix Kreuk, Emily McMilin, Michel Meyer, Yuxiang Wei, David Zhang, Kunhao Zheng, Jordi Armengol-Estap{\'e}, Pedram Bashiri, Maximilian Beck, et~al.
\newblock {CWM}: An open-weights {LLM} for research on code generation with world models.
\newblock \emph{arXiv preprint arXiv:2510.02387}, 2025.
\newblock URL \url{https://arxiv.org/abs/2510.02387}.

\bibitem[Dainese et~al.(2024)Dainese, Merler, Alakuijala, and Marttinen]{dainese2024codewm}
Nicola Dainese, Matteo Merler, Minttu Alakuijala, and Pekka Marttinen.
\newblock Generating code world models with large language models guided by {Monte Carlo} tree search.
\newblock In \emph{Adv. Neural Inf. Process. Syst.}, volume~37, 2024.
\newblock URL \url{https://arxiv.org/abs/2405.15383}.

\bibitem[Decart et~al.(2024)Decart, Quevedo, McIntyre, Campbell, Chen, and Wachen]{oasis2024}
Decart, Julian Quevedo, Quinn McIntyre, Spruce Campbell, Xinlei Chen, and Robert Wachen.
\newblock Oasis: A universe in a transformer.
\newblock Blog post, 2024.
\newblock URL \url{https://oasis-model.github.io}.

\bibitem[Ding et~al.(2025)Ding, Zhang, Shang, Feng, Zhang, Zong, Yuan, Su, et~al.]{ding2024survey_wm}
Jingtao Ding, Yunke Zhang, Yu~Shang, Jie Feng, Yuheng Zhang, Zefang Zong, Yuan Yuan, Hongyuan Su, et~al.
\newblock Understanding world or predicting future? {A} comprehensive survey of world models.
\newblock \emph{ACM Comput. Surv.}, 2025.
\newblock URL \url{https://arxiv.org/abs/2411.14499}.

\bibitem[Dong et~al.(2025)Dong, Wu, Chen, Kong, Zhu, Hu, Zhou, Sun, He, Dai, Hauptmann, and Cheng]{dong2026uniwm}
Yifei Dong, Fengyi Wu, Guangyu Chen, Lingdong Kong, Xu~Zhu, Qiyu Hu, Yuxuan Zhou, Jingdong Sun, Jun-Yan He, Qi~Dai, Alexander~G. Hauptmann, and Zhi-Qi Cheng.
\newblock Towards unified world models for visual navigation via memory-augmented planning and foresight.
\newblock \emph{arXiv preprint arXiv:2510.08713}, 2025.
\newblock URL \url{https://arxiv.org/abs/2510.08713}.

\bibitem[Dong et~al.(2026)Dong, Wu, Dai, Kong, Chen, Zhu, Hu, Wang, et~al.]{dong2026lcvn}
Yifei Dong, Fengyi Wu, Yilong Dai, Lingdong Kong, Guangyu Chen, Xu~Zhu, Qiyu Hu, Tianyu Wang, et~al.
\newblock Language-conditioned world modeling for visual navigation.
\newblock \emph{arXiv preprint arXiv:2603.26741}, 2026.
\newblock URL \url{https://arxiv.org/abs/2603.26741}.

\bibitem[Dubois et~al.(2024)Dubois, Galambosi, Liang, and Hashimoto]{dubois2024alpacaeval}
Yann Dubois, Bal{\'a}zs Galambosi, Percy Liang, and Tatsunori~B. Hashimoto.
\newblock Length-controlled {AlpacaEval}: A simple way to debias automatic evaluators.
\newblock In \emph{Conf. Lang. Model.}, 2024.
\newblock URL \url{https://arxiv.org/abs/2404.04475}.

\bibitem[Ethayarajh et~al.(2024)Ethayarajh, Xu, Muennighoff, Jurafsky, and Kiela]{ethayarajh2024kto}
Kawin Ethayarajh, Winnie Xu, Niklas Muennighoff, Dan Jurafsky, and Douwe Kiela.
\newblock {KTO}: Model alignment as prospect theoretic optimization.
\newblock In \emph{Int. Conf. Mach. Learn.}, 2024.
\newblock URL \url{https://arxiv.org/abs/2402.01306}.

\bibitem[Fang et~al.(2025)Fang, Zhang, Zhang, Ma, Yu, Mi, and Yu]{webevolver2025}
Tianqing Fang, Hongming Zhang, Zhisong Zhang, Kaixin Ma, Wenhao Yu, Haitao Mi, and Dong Yu.
\newblock {WebEvolver}: Enhancing web agent self-improvement with coevolving world model.
\newblock In \emph{Proc. Conf. Empir. Methods Nat. Lang. Process.}, pages 8959--8975, 2025.
\newblock URL \url{https://arxiv.org/abs/2504.21024}.

\bibitem[Fei et~al.(2024)Fei, Xu, Zhang, Zhou, Yang, and He]{fei20243dgaussianera}
Ben Fei, Jingyi Xu, Rui Zhang, Qingyuan Zhou, Weidong Yang, and Ying He.
\newblock {3D} gaussian as a new era: A survey.
\newblock \emph{IEEE Trans. Vis. Comput. Graph.}, 2024.
\newblock URL \url{https://arxiv.org/abs/2402.07181}.

\bibitem[Feng et~al.(2025{\natexlab{a}})Feng, Zhang, Zhang, Lu, Liu, and Wang]{feng2025wwm}
Jichen Feng, Yifan Zhang, Chenggong Zhang, Yifu Lu, Shilong Liu, and Mengdi Wang.
\newblock Web world models.
\newblock \emph{arXiv preprint arXiv:2512.23676}, 2025{\natexlab{a}}.
\newblock URL \url{https://arxiv.org/abs/2512.23676}.

\bibitem[Feng et~al.(2025{\natexlab{b}})Feng, Wang, and Yang]{feng2025ad_wm_survey}
Tuo Feng, Wenguan Wang, and Yi~Yang.
\newblock A survey of world models for autonomous driving.
\newblock \emph{arXiv preprint arXiv:2501.11260}, 2025{\natexlab{b}}.
\newblock URL \url{https://arxiv.org/abs/2501.11260}.

\bibitem[Fridman et~al.(2023)Fridman, Abecasis, Kasten, and Dekel]{fridman2023scenescape}
Rafail Fridman, Amit Abecasis, Yoni Kasten, and Tali Dekel.
\newblock {SceneScape}: Text-driven consistent scene generation.
\newblock In \emph{Adv. Neural Inf. Process. Syst.}, 2023.
\newblock URL \url{https://arxiv.org/abs/2302.01133}.

\bibitem[Ganapavarapu and Patel(2026)]{ganapavarapu2026mcpcosmos}
Giridhar Ganapavarapu and Dhaval Patel.
\newblock {MCP-Cosmos}: World model-augmented agents for complex task execution in {MCP} environments.
\newblock \emph{arXiv preprint arXiv:2605.09131}, 2026.
\newblock URL \url{https://arxiv.org/abs/2605.09131}.

\bibitem[Gao et~al.(2024)Gao, Yang, Chen, Chitta, Qiu, Geiger, Zhang, and Li]{gao2024vista}
Shenyuan Gao, Jiazhi Yang, Li~Chen, Kashyap Chitta, Yihang Qiu, Andreas Geiger, Jun Zhang, and Hongyang Li.
\newblock Vista: A generalizable driving world model with high fidelity and versatile controllability.
\newblock In \emph{Adv. Neural Inf. Process. Syst.}, volume~37, 2024.
\newblock URL \url{https://arxiv.org/abs/2405.17398}.

\bibitem[Gao et~al.(2025)Gao, Ye, Wang, and Sang]{gao2025websynthesis}
Yifei Gao, Junhong Ye, Jiaqi Wang, and Jitao Sang.
\newblock {WebSynthesis}: World-model-guided {MCTS} for efficient {WebUI}-trajectory synthesis.
\newblock \emph{arXiv preprint arXiv:2507.04370}, 2025.
\newblock URL \url{https://arxiv.org/abs/2507.04370}.

\bibitem[Gao et~al.(2026)Gao, Wang, Zeng, Zhu, Cheng, Li, Wang, Xu, et~al.]{gao2026lingbotworld}
Zelin Gao, Qiuyu Wang, Yanhong Zeng, Jiapeng Zhu, Ka~Leong Cheng, Yixuan Li, Hanlin Wang, Yinghao Xu, et~al.
\newblock Advancing open-source world models.
\newblock \emph{arXiv preprint arXiv:2601.20540}, 2026.
\newblock URL \url{https://arxiv.org/abs/2601.20540}.

\bibitem[{Google DeepMind}(2025)]{deepmind2025genie3}
{Google DeepMind}.
\newblock Genie 3: A new frontier for world models.
\newblock DeepMind Technical Blog, 2025.
\newblock URL \url{https://deepmind.google/discover/blog/genie-3-a-new-frontier-for-world-models/}.

\bibitem[Gu et~al.(2024)Gu, Jiang, Shi, Tan, Zhai, Xu, Li, Shen, et~al.]{gu2024judge}
Jiawei Gu, Xuhui Jiang, Zhichao Shi, Hexiang Tan, Xuehao Zhai, Chengjin Xu, Wei Li, Yinghan Shen, et~al.
\newblock A survey on {LLM}-as-a-judge.
\newblock \emph{arXiv preprint arXiv:2411.15594}, 2024.
\newblock URL \url{https://arxiv.org/abs/2411.15594}.

\bibitem[Gu et~al.(2025)Gu, Zhang, Ning, Zheng, Gou, Xue, Chang, Srivastava, Xie, Qi, Sun, and Su]{gu2024webdreamer}
Yu~Gu, Kai Zhang, Yuting Ning, Boyuan Zheng, Boyu Gou, Tianci Xue, Cheng Chang, Sanjari Srivastava, Yanan Xie, Peng Qi, Huan Sun, and Yu~Su.
\newblock Is your {LLM} secretly a world model of the internet? {M}odel-based planning for web agents.
\newblock \emph{Trans. Mach. Learn. Res.}, 2025.
\newblock URL \url{https://arxiv.org/abs/2411.06559}.

\bibitem[Guan et~al.(2026)Guan, Yu, Zhang, Wang, Zhang, Li, Qiao, Qin, et~al.]{cuwm2026}
Yiming Guan, Rui Yu, John Zhang, Lu~Wang, Chaoyun Zhang, Liqun Li, Bo~Qiao, Si~Qin, et~al.
\newblock Computer-using world model.
\newblock \emph{arXiv preprint arXiv:2602.17365}, 2026.
\newblock URL \url{https://arxiv.org/abs/2602.17365}.

\bibitem[Guo et~al.(2025{\natexlab{a}})Guo, Ye, He, Wu, Jiang, Pearce, and Bian]{guo2025mineworld}
Junliang Guo, Yang Ye, Tianyu He, Haoyu Wu, Yushu Jiang, Tim Pearce, and Jiang Bian.
\newblock {MineWorld}: a real-time and open-source interactive world model on minecraft.
\newblock \emph{arXiv preprint arXiv:2504.08388}, 2025{\natexlab{a}}.
\newblock URL \url{https://arxiv.org/abs/2504.08388}.

\bibitem[Guo et~al.(2025{\natexlab{b}})Guo, Darwiche~Domingues, Avalos, Courville, and Strub]{guo2025dymo}
Shangmin Guo, Omar Darwiche~Domingues, Rapha{\"e}l Avalos, Aaron Courville, and Florian Strub.
\newblock World modelling improves language model agents.
\newblock \emph{arXiv preprint arXiv:2506.02918}, 2025{\natexlab{b}}.
\newblock URL \url{https://arxiv.org/abs/2506.02918}.

\bibitem[Guo et~al.(2026)Guo, Liang, Balogh, Lunberry, Tu, Jelasity, and Tao]{guo2026physcond}
Zhixiang Guo, Siyuan Liang, Andr{\'a}s Balogh, Noah Lunberry, Rong-Cheng Tu, M{\'a}rk Jelasity, and Dacheng Tao.
\newblock When world models dream wrong: Physical-conditioned adversarial attacks against world models.
\newblock \emph{arXiv preprint arXiv:2602.18739}, 2026.
\newblock URL \url{https://arxiv.org/abs/2602.18739}.

\bibitem[Ha and Schmidhuber(2018)]{ha2018worldmodels}
David Ha and J{\"u}rgen Schmidhuber.
\newblock Recurrent world models facilitate policy evolution.
\newblock In \emph{Adv. Neural Inf. Process. Syst.}, volume~31, 2018.
\newblock URL \url{https://arxiv.org/abs/1803.10122}.

\bibitem[Hafner et~al.(2019)Hafner, Lillicrap, Fischer, Villegas, Ha, Lee, and Davidson]{hafner2019recurrent}
Danijar Hafner, Timothy Lillicrap, Ian Fischer, Ruben Villegas, David Ha, Honglak Lee, and James Davidson.
\newblock Learning latent dynamics for planning from pixels.
\newblock In \emph{Int. Conf. Mach. Learn.}, pages 2555--2565, 2019.
\newblock URL \url{https://arxiv.org/abs/1811.04551}.

\bibitem[Hafner et~al.(2020)Hafner, Lillicrap, Ba, and Norouzi]{hafner2019dreamer}
Danijar Hafner, Timothy Lillicrap, Jimmy Ba, and Mohammad Norouzi.
\newblock Dream to control: Learning behaviors by latent imagination.
\newblock In \emph{Int. Conf. Learn. Represent.}, 2020.
\newblock URL \url{https://arxiv.org/abs/1912.01603}.

\bibitem[Hafner et~al.(2025)Hafner, Pasukonis, Ba, and Lillicrap]{hafner2023dreamerv3}
Danijar Hafner, Jurgis Pasukonis, Jimmy Ba, and Timothy Lillicrap.
\newblock Mastering diverse control tasks through world models.
\newblock \emph{Nature}, 640\penalty0 (8059):\penalty0 647--653, 2025.
\newblock URL \url{https://arxiv.org/abs/2301.04104}.

\bibitem[Hansen et~al.(2024)Hansen, Su, and Wang]{hansen2024tdmpc2}
Nicklas Hansen, Hao Su, and Xiaolong Wang.
\newblock {TD-MPC2}: Scalable, robust world models for continuous control.
\newblock In \emph{Int. Conf. Learn. Represent.}, 2024.
\newblock URL \url{https://arxiv.org/abs/2310.16828}.

\bibitem[Hao et~al.(2023)Hao, Gu, Ma, Hong, Wang, Wang, and Hu]{hao2023reasoning}
Shibo Hao, Yi~Gu, Haodi Ma, Joshua~Jiahua Hong, Zhen Wang, Daisy~Zhe Wang, and Zhiting Hu.
\newblock Reasoning with language model is planning with world model.
\newblock In \emph{Proc. Conf. Empir. Methods Nat. Lang. Process.}, pages 8154--8173, 2023.
\newblock URL \url{https://arxiv.org/abs/2305.14992}.

\bibitem[Ho et~al.(2022)Ho, Salimans, Gritsenko, Chan, Norouzi, and Fleet]{ho2022videodiffusion}
Jonathan Ho, Tim Salimans, Alexey Gritsenko, William Chan, Mohammad Norouzi, and David~J. Fleet.
\newblock Video diffusion models.
\newblock In \emph{Adv. Neural Inf. Process. Syst.}, volume~35, 2022.
\newblock URL \url{https://arxiv.org/abs/2204.03458}.

\bibitem[H{\"o}llein et~al.(2023)H{\"o}llein, Cao, Owens, Johnson, and Nie{\ss}ner]{hollein2023text2room}
Lukas H{\"o}llein, Ang Cao, Andrew Owens, Justin Johnson, and Matthias Nie{\ss}ner.
\newblock {Text2Room}: Extracting textured {3D} meshes from {2D} text-to-image models.
\newblock In \emph{IEEE/CVF Int. Conf. Comput. Vis.}, 2023.
\newblock URL \url{https://arxiv.org/abs/2303.11989}.

\bibitem[Hong et~al.(2024)Hong, Lee, and Thorne]{hong2024orpo}
Jiwoo Hong, Noah Lee, and James Thorne.
\newblock {ORPO}: Monolithic preference optimization without reference model.
\newblock In \emph{Proc. Conf. Empir. Methods Nat. Lang. Process.}, 2024.
\newblock URL \url{https://arxiv.org/abs/2403.07691}.

\bibitem[H{\"o}ppe et~al.(2022)H{\"o}ppe, Mehrjou, Bauer, Nielsen, and Dittadi]{hoeppe2022ramvid}
Tobias H{\"o}ppe, Arash Mehrjou, Stefan Bauer, Didrik Nielsen, and Andrea Dittadi.
\newblock Diffusion models for video prediction and infilling.
\newblock \emph{Trans. Mach. Learn. Res.}, 2022.
\newblock URL \url{https://arxiv.org/abs/2206.07696}.

\bibitem[Hu et~al.(2023)Hu, Russell, Yeo, Murez, Fedoseev, Kendall, Shotton, and Corrado]{hu2023gaia1}
Anthony Hu, Lloyd Russell, Hudson Yeo, Zak Murez, George Fedoseev, Alex Kendall, Jamie Shotton, and Gianluca Corrado.
\newblock {GAIA-1}: A generative world model for autonomous driving.
\newblock \emph{arXiv preprint arXiv:2309.17080}, 2023.
\newblock URL \url{https://arxiv.org/abs/2309.17080}.

\bibitem[Hu et~al.(2026{\natexlab{a}})Hu, Ai, Wen, Li, Zou, Li, Deng, Cai, Zhou, Cai, et~al.]{hu2026itercad}
Tao Hu, Jiaxin Ai, Licheng Wen, Xueheng Li, Shu Zou, Siqi Li, Nianchen Deng, Xinyu Cai, Hongbin Zhou, Pinlong Cai, et~al.
\newblock {IterCAD}: An iterative multimodal agent for visually-grounded {CAD} generation and editing.
\newblock \emph{arXiv preprint arXiv:2606.13368}, 2026{\natexlab{a}}.
\newblock URL \url{https://arxiv.org/abs/2606.13368}.

\bibitem[Hu et~al.(2026{\natexlab{b}})Hu, Gong, Kong, Mei, Ding, Zeng, Liang, Li, Zhong, and Liang]{hu2026navthinker}
Tianshuai Hu, Zeying Gong, Lingdong Kong, Xiaodong Mei, Yiyi Ding, Qi~Zeng, Ao~Liang, Rong Li, Yangyi Zhong, and Junwei Liang.
\newblock {NavThinker}: Action-conditioned world models for coupled prediction and planning in social navigation.
\newblock \emph{arXiv preprint arXiv:2603.15359}, 2026{\natexlab{b}}.
\newblock URL \url{https://arxiv.org/abs/2603.15359}.

\bibitem[Hu et~al.(2026{\natexlab{c}})Hu, Zhang, Huang, Tu, Su, Deng, Liu, Liu, Liu, and Ho]{hu2026occubench}
Xiaomeng Hu, Yinger Zhang, Fei Huang, Jianhong Tu, Yang Su, Lianghao Deng, Yuxuan Liu, Yantao Liu, Dayiheng Liu, and Tsung-Yi Ho.
\newblock {OccuBench}: Evaluating {AI} agents on real-world professional tasks via language world models.
\newblock \emph{arXiv preprint arXiv:2604.10866}, 2026{\natexlab{c}}.
\newblock URL \url{https://arxiv.org/abs/2604.10866}.

\bibitem[Huang et~al.(2025{\natexlab{a}})Huang, Chen, Zhang, Sun, and Schwager]{huang2025particleformer}
Suning Huang, Qianzhong Chen, Xiaohan Zhang, Jiankai Sun, and Mac Schwager.
\newblock {ParticleFormer}: A {3D} point cloud world model for multi-object, multi-material robotic manipulation.
\newblock \emph{arXiv preprint arXiv:2506.23126}, 2025{\natexlab{a}}.
\newblock URL \url{https://arxiv.org/abs/2506.23126}.

\bibitem[Huang et~al.(2022)Huang, Xia, Xiao, Chan, Liang, Florence, Zeng, Tompson, et~al.]{huang2022innermonologue}
Wenlong Huang, Fei Xia, Ted Xiao, Harris Chan, Jacky Liang, Pete Florence, Andy Zeng, Jonathan Tompson, et~al.
\newblock Inner monologue: Embodied reasoning through planning with language models.
\newblock In \emph{Conf. Robot Learn.}, 2022.
\newblock URL \url{https://arxiv.org/abs/2207.05608}.

\bibitem[Huang et~al.(2023)Huang, Wang, Zhang, Li, Wu, and Fei-Fei]{huang2023voxposer}
Wenlong Huang, Chen Wang, Ruohan Zhang, Yunzhu Li, Jiajun Wu, and Li~Fei-Fei.
\newblock {VoxPoser}: Composable {3D} value maps for robotic manipulation with language models.
\newblock In \emph{Conf. Robot Learn.}, 2023.
\newblock URL \url{https://arxiv.org/abs/2307.05973}.

\bibitem[Huang et~al.(2026)Huang, Chao, Mousavian, Liu, Fox, Mo, and Fei-Fei]{huang2026pointworld}
Wenlong Huang, Yu-Wei Chao, Arsalan Mousavian, Ming-Yu Liu, Dieter Fox, Kaichun Mo, and Li~Fei-Fei.
\newblock {PointWorld}: Scaling {3D} world models for in-the-wild robotic manipulation.
\newblock \emph{arXiv preprint arXiv:2601.03782}, 2026.
\newblock URL \url{https://arxiv.org/abs/2601.03782}.

\bibitem[Huang et~al.(2025{\natexlab{b}})Huang, Chen, Fei, Li, Schwaller, and Ceder]{huang2025cascade}
Xu~Huang, Junwu Chen, Yuxing Fei, Zhuohan Li, Philippe Schwaller, and Gerbrand Ceder.
\newblock {CASCADE}: Cumulative agentic skill creation through autonomous development and evolution.
\newblock \emph{arXiv preprint arXiv:2512.23880}, 2025{\natexlab{b}}.
\newblock URL \url{https://arxiv.org/abs/2512.23880}.

\bibitem[Janner et~al.(2019)Janner, Fu, Zhang, and Levine]{janner2019mbpo}
Michael Janner, Justin Fu, Marvin Zhang, and Sergey Levine.
\newblock When to trust your model: Model-based policy optimization.
\newblock In \emph{Adv. Neural Inf. Process. Syst.}, volume~32, 2019.
\newblock URL \url{https://arxiv.org/abs/1906.08253}.

\bibitem[Jimenez et~al.(2024)Jimenez, Yang, Wettig, Yao, Pei, Press, and Narasimhan]{jimenez2023swebench}
Carlos~E. Jimenez, John Yang, Alexander Wettig, Shunyu Yao, Kexin Pei, Ofir Press, and Karthik Narasimhan.
\newblock {SWE-bench}: Can language models resolve real-world {GitHub} issues?
\newblock In \emph{Int. Conf. Learn. Represent.}, 2024.
\newblock URL \url{https://arxiv.org/abs/2310.06770}.

\bibitem[Jing et~al.(2026)Jing, Hao, Zhou, and Yu]{jing2026counterscene}
Bowen Jing, Ruiyang Hao, Weitao Zhou, and Haibao Yu.
\newblock {CounterScene}: Counterfactual causal reasoning in generative world models for safety-critical closed-loop evaluation.
\newblock \emph{arXiv preprint arXiv:2603.21104}, 2026.
\newblock URL \url{https://arxiv.org/abs/2603.21104}.

\bibitem[Kaiser et~al.(2020)Kaiser, Babaeizadeh, Milos, Osinski, Campbell, Czechowski, Erhan, Finn, et~al.]{kaiser2019simple}
Lukasz Kaiser, Mohammad Babaeizadeh, Piotr Milos, Blazej Osinski, Roy~H. Campbell, K.~Czechowski, D.~Erhan, Chelsea Finn, et~al.
\newblock Model-based reinforcement learning for {Atari}.
\newblock In \emph{Int. Conf. Learn. Represent.}, 2020.
\newblock URL \url{https://arxiv.org/abs/1903.00374}.

\bibitem[Kang et~al.(2025)Kang, Yue, Lu, Lin, Zhao, Wang, Huang, and Feng]{kang2025howfar}
Bingyi Kang, Yang Yue, Rui Lu, Zhijie Lin, Yang Zhao, Kaixin Wang, Gao Huang, and Jiashi Feng.
\newblock How far is video generation from world model: A physical law perspective.
\newblock In \emph{Int. Conf. Mach. Learn.}, 2025.
\newblock URL \url{https://arxiv.org/abs/2411.02385}.

\bibitem[Kerbl et~al.(2023)Kerbl, Kopanas, Leimk{\"u}hler, and Drettakis]{kerbl20233dgs}
Bernhard Kerbl, Georgios Kopanas, Thomas Leimk{\"u}hler, and George Drettakis.
\newblock {3D} {Gaussian} splatting for real-time radiance field rendering.
\newblock \emph{ACM Trans. Graph.}, 42\penalty0 (4), 2023.
\newblock URL \url{https://arxiv.org/abs/2308.04079}.

\bibitem[Kim et~al.(2024)Kim, Shin, Cho, Jang, Longpre, Lee, Yun, Shin, Kim, Thorne, and Seo]{kim2023prometheus}
Seungone Kim, Jamin Shin, Yejin Cho, Joel Jang, Shayne Longpre, Hwaran Lee, Sangdoo Yun, Seongjin Shin, Sungdong Kim, James Thorne, and Minjoon Seo.
\newblock {Prometheus}: Inducing fine-grained evaluation capability in language models.
\newblock In \emph{Int. Conf. Learn. Represent.}, 2024.
\newblock URL \url{https://arxiv.org/abs/2310.08491}.

\bibitem[Koh et~al.(2021)Koh, Lee, Yang, Baldridge, and Anderson]{koh2021pathdreamer}
Jing~Yu Koh, Honglak Lee, Yinfei Yang, Jason Baldridge, and Peter Anderson.
\newblock {PathDreamer}: A world model for indoor navigation.
\newblock In \emph{IEEE/CVF Int. Conf. Comput. Vis.}, pages 14738--14748, 2021.
\newblock URL \url{https://arxiv.org/abs/2105.08756}.

\bibitem[Koh et~al.(2026)Koh, Han, Lee, Yun, and Shin]{gworld2026}
Woosung Koh, Sungjun Han, Segyu Lee, Se-Young Yun, and Jamin Shin.
\newblock Generative visual code mobile world models.
\newblock \emph{arXiv preprint arXiv:2602.01576}, 2026.
\newblock URL \url{https://arxiv.org/abs/2602.01576}.

\bibitem[Kojima et~al.(2022)Kojima, Gu, Reid, Matsuo, and Iwasawa]{kojima2022zeroshotcot}
Takeshi Kojima, Shixiang~Shane Gu, Machel Reid, Yutaka Matsuo, and Yusuke Iwasawa.
\newblock Large language models are zero-shot reasoners.
\newblock In \emph{Adv. Neural Inf. Process. Syst.}, 2022.
\newblock URL \url{https://arxiv.org/abs/2205.11916}.

\bibitem[Kong et~al.(2025)Kong, Yang, Mei, Liu, Liang, Zhu, Lu, Yin, et~al.]{kong2025survey_3d4d}
Lingdong Kong, Wesley Yang, Jianbiao Mei, Youquan Liu, Ao~Liang, Dekai Zhu, Dongyue Lu, Wei Yin, et~al.
\newblock {3D} and {4D} world modeling: A survey.
\newblock \emph{arXiv preprint arXiv:2509.07996}, 2025.
\newblock URL \url{https://arxiv.org/abs/2509.07996}.

\bibitem[Kong et~al.(2026)Kong, Liang, Yan, Liu, Yang, Huang, Sun, Yin, et~al.]{worldlens}
Lingdong Kong, Ao~Liang, Tianyi Yan, Hongsi Liu, Wesley Yang, Ziqi Huang, Xian Sun, Wei Yin, et~al.
\newblock Is your driving world model an all-around player?
\newblock In \emph{IEEE/CVF Conf. Comput. Vis. Pattern Recog.}, pages 36385--36399, 2026.
\newblock URL \url{https://arxiv.org/abs/2605.10858}.

\bibitem[LeCun(2022)]{lecun2022path}
Yann LeCun.
\newblock A path towards autonomous machine intelligence.
\newblock \emph{OpenReview preprint}, 2022.
\newblock URL \url{https://openreview.net/pdf?id=BZ5a1r-kVsf}.

\bibitem[Leroy et~al.(2024)Leroy, Cabon, and Revaud]{leroy2024mast3r}
Vincent Leroy, Yohann Cabon, and Jerome Revaud.
\newblock Grounding image matching in {3D} with {MASt3R}.
\newblock In \emph{Eur. Conf. Comput. Vis.}, 2024.
\newblock URL \url{https://arxiv.org/abs/2406.09756}.

\bibitem[Li et~al.(2024{\natexlab{a}})Li, Zhang, Wong, Gokmen, Srivastava, Mart{\'i}n-Mart{\'i}n, Wang, Levine, et~al.]{li2024behavior1k}
Chengshu Li, Ruohan Zhang, Josiah Wong, Cem Gokmen, Sanjana Srivastava, Roberto Mart{\'i}n-Mart{\'i}n, Chen Wang, Gabrael Levine, et~al.
\newblock {BEHAVIOR-1K}: A human-centered, embodied {AI} benchmark with 1,000 everyday activities and realistic simulation.
\newblock \emph{arXiv preprint arXiv:2403.09227}, 2024{\natexlab{a}}.
\newblock URL \url{https://arxiv.org/abs/2403.09227}.

\bibitem[Li et~al.(2025{\natexlab{a}})Li, Fang, Chen, Yang, Cao, Wong, Luo, Wang, et~al.]{fan2025worldmodelbench}
Dacheng Li, Yunhao Fang, Yukang Chen, Shuo Yang, Shiyi Cao, Justin Wong, Michael Luo, Xiaolong Wang, et~al.
\newblock {WorldModelBench}: Judging video generation models as world models.
\newblock \emph{arXiv preprint arXiv:2502.20694}, 2025{\natexlab{a}}.
\newblock URL \url{https://arxiv.org/abs/2502.20694}.

\bibitem[Li et~al.(2024{\natexlab{b}})Li, Shi, Zhang, Wu, Liao, Wang, Lee, and Zhou]{li2024dreamscene}
Haoran Li, Haolin Shi, Wenli Zhang, Wenjun Wu, Yong Liao, Lin Wang, Lik-Hang Lee, and Pengyuan Zhou.
\newblock {DreamScene}: {3D} gaussian-based text-to-{3D} scene generation via formation pattern sampling.
\newblock In \emph{Eur. Conf. Comput. Vis.}, 2024{\natexlab{b}}.
\newblock URL \url{https://arxiv.org/abs/2404.03575}.

\bibitem[Li et~al.(2025{\natexlab{b}})Li, Kallidromitis, Gokul, Kato, Kozuka, and Grover]{li2025mobileworldbench}
Shufan Li, Konstantinos Kallidromitis, Akash Gokul, Yusuke Kato, Kazuki Kozuka, and Aditya Grover.
\newblock {MobileWorldBench}: Towards semantic world modeling for mobile agents.
\newblock \emph{arXiv preprint arXiv:2512.14014}, 2025{\natexlab{b}}.
\newblock URL \url{https://arxiv.org/abs/2512.14014}.

\bibitem[Li et~al.(2026)Li, Zheng, Gao, Xia, Wang, Wang, et~al.]{li2026embodiedsafety}
Xiao Li, Xiang Zheng, Yifeng Gao, Xinyu Xia, Yixu Wang, Xin Wang, et~al.
\newblock Safety in embodied {AI}: A survey of risks, attacks, and defenses.
\newblock \emph{arXiv preprint arXiv:2605.02900}, 2026.
\newblock URL \url{https://arxiv.org/abs/2605.02900}.

\bibitem[Li et~al.(2025{\natexlab{c}})Li, He, Zhang, Wu, Li, and Liu]{li2025embodied_wm_survey}
Xinqing Li, Xin He, Le~Zhang, Min Wu, Xiaoli Li, and Yun Liu.
\newblock A comprehensive survey on world models for embodied {AI}.
\newblock \emph{arXiv preprint arXiv:2510.16732}, 2025{\natexlab{c}}.
\newblock URL \url{https://arxiv.org/abs/2510.16732}.

\bibitem[Li et~al.(2025{\natexlab{d}})Li, Tucker, Cole, Wang, Jin, Ye, Kanazawa, Holynski, and Snavely]{li2024megasam}
Zhengqi Li, Richard Tucker, Forrester Cole, Qianqian Wang, Linyi Jin, Vickie Ye, Angjoo Kanazawa, Aleksander Holynski, and Noah Snavely.
\newblock {MegaSaM}: Accurate, fast, and robust structure and motion from casual dynamic videos.
\newblock In \emph{IEEE/CVF Conf. Comput. Vis. Pattern Recog.}, 2025{\natexlab{d}}.
\newblock URL \url{https://arxiv.org/abs/2412.04463}.

\bibitem[Liang et~al.(2026)Liang, Liu, Yang, Lu, Li, Kong, Zhao, and Ooi]{liang2026lidarcrafter}
Ao~Liang, Youquan Liu, Yu~Yang, Dongyue Lu, Linfeng Li, Lingdong Kong, Huaici Zhao, and Wei~Tsang Ooi.
\newblock {LiDARCrafter}: Dynamic {4D} world modeling from {LiDAR} sequences.
\newblock In \emph{AAAI Conf. Artif. Intell.}, volume~40, pages 18406--18414, 2026.
\newblock URL \url{https://arxiv.org/abs/2508.03692}.

\bibitem[Liang et~al.(2023)Liang, Huang, Xia, Xu, Hausman, Ichter, Florence, and Zeng]{liang2022codeaspolicies}
Jacky Liang, Wenlong Huang, Fei Xia, Peng Xu, Karol Hausman, Brian Ichter, Pete Florence, and Andy Zeng.
\newblock Code as policies: Language model programs for embodied control.
\newblock In \emph{IEEE Int. Conf. Robot. Autom.}, 2023.
\newblock URL \url{https://arxiv.org/abs/2209.07753}.

\bibitem[Lightman et~al.(2024)Lightman, Kosaraju, Burda, Edwards, Baker, Lee, Leike, Schulman, Sutskever, and Cobbe]{lightman2023letsverify}
Hunter Lightman, Vineet Kosaraju, Yura Burda, Harri Edwards, Bowen Baker, Teddy Lee, Jan Leike, John Schulman, Ilya Sutskever, and Karl Cobbe.
\newblock Let's verify step by step.
\newblock In \emph{Int. Conf. Learn. Represent.}, 2024.
\newblock URL \url{https://arxiv.org/abs/2305.20050}.

\bibitem[Lin et~al.(2025)Lin, Hu, Li, Yang, Wang, Torr, and Shou]{lin2025computer}
Kevin~Qinghong Lin, Siyuan Hu, Linjie Li, Zhengyuan Yang, Lijuan Wang, Philip Torr, and Mike~Zheng Shou.
\newblock Computer-use agents as judges for generative user interface.
\newblock \emph{arXiv preprint arXiv:2511.15567}, 2025.
\newblock URL \url{https://arxiv.org/abs/2511.15567}.

\bibitem[Liu et~al.(2026)Liu, Wang, Long, Hou, Sun, Li, Yang, Peng, Liu, Jiang, Yao, and Mu]{liu2026jailwam}
Hanqing Liu, Songping Wang, Jiahuan Long, Jiacheng Hou, Jialiang Sun, Chao Li, Yang Yang, Wei Peng, Xu~Liu, Tingsong Jiang, Wen Yao, and Yao Mu.
\newblock {JailWAM}: Jailbreaking world action models in robot control.
\newblock \emph{arXiv preprint arXiv:2604.05498}, 2026.
\newblock URL \url{https://arxiv.org/abs/2604.05498}.

\bibitem[Liu et~al.(2025)Liu, Yan, Zaharia, and Abbeel]{liu2024lwm}
Hao Liu, Wilson Yan, Matei Zaharia, and Pieter Abbeel.
\newblock World model on million-length video and language with blockwise {RingAttention}.
\newblock In \emph{Int. Conf. Learn. Represent.}, 2025.
\newblock URL \url{https://arxiv.org/abs/2402.08268}.

\bibitem[Liu et~al.(2024)Liu, Yu, Zhang, Xu, Lei, Lai, Gu, et~al.]{liu2023agentbench}
Xiao Liu, Hao Yu, Hanchen Zhang, Yifan Xu, Xuanyu Lei, Hanyu Lai, Yu~Gu, et~al.
\newblock {AgentBench}: Evaluating {LLMs} as agents.
\newblock In \emph{Int. Conf. Learn. Represent.}, 2024.
\newblock URL \url{https://arxiv.org/abs/2308.03688}.

\bibitem[Liu et~al.(2023)Liu, Iter, Xu, Wang, Xu, and Zhu]{liu2023geval}
Yang Liu, Dan Iter, Yichong Xu, Shuohang Wang, Ruochen Xu, and Chenguang Zhu.
\newblock {G-Eval}: {NLG} evaluation using {GPT-4} with better human alignment.
\newblock In \emph{Proc. Conf. Empir. Methods Nat. Lang. Process.}, 2023.
\newblock URL \url{https://arxiv.org/abs/2303.16634}.

\bibitem[Lu et~al.(2025{\natexlab{a}})Lu, Jia, Li, Chen, Wang, Tang, and Huang]{lu2025gwm}
Guanxing Lu, Baoxiong Jia, Puhao Li, Yixin Chen, Ziwei Wang, Yansong Tang, and Siyuan Huang.
\newblock {GWM}: Towards scalable gaussian world models for robotic manipulation.
\newblock In \emph{IEEE/CVF Int. Conf. Comput. Vis.}, pages 9263--9274, 2025{\natexlab{a}}.
\newblock URL \url{https://arxiv.org/abs/2508.17600}.

\bibitem[Lu et~al.(2025{\natexlab{b}})Lu, Holleis, Zhang, Aumayer, Nan, Bai, Ma, Ma, Li, Yin, Wang, and Pang]{lu2024toolsandbox}
Jiarui Lu, Thomas Holleis, Yizhe Zhang, Bernhard Aumayer, Feng Nan, Felix Bai, Shuang Ma, Shen Ma, Mengyu Li, Guoli Yin, Zirui Wang, and Ruoming Pang.
\newblock {ToolSandbox}: A stateful, conversational, interactive evaluation benchmark for {LLM} tool use capabilities.
\newblock In \emph{Proc. Conf. N. Am. Chapter Assoc. Comput. Linguist.}, 2025{\natexlab{b}}.
\newblock URL \url{https://arxiv.org/abs/2408.04682}.

\bibitem[Lu et~al.(2024)Lu, Yu, Xu, Xiangli, Wang, Lin, and Dai]{lu2023scaffoldgs}
Tao Lu, Mulin Yu, Linning Xu, Yuanbo Xiangli, Limin Wang, Dahua Lin, and Bo~Dai.
\newblock {Scaffold-GS}: Structured {3D} gaussians for view-adaptive rendering.
\newblock In \emph{IEEE/CVF Conf. Comput. Vis. Pattern Recog.}, 2024.
\newblock URL \url{https://arxiv.org/abs/2312.00109}.

\bibitem[Luo et~al.(2025)Luo, Tang, Li, Papoudakis, Song, Gong, Hao, Wang, and Shao]{luo2025vimo}
Dezhao Luo, Bohan Tang, Kang Li, Georgios Papoudakis, Jifei Song, Shaogang Gong, Jianye Hao, Jun Wang, and Kun Shao.
\newblock {ViMo}: A generative visual {GUI} world model for app agents.
\newblock \emph{arXiv preprint arXiv:2504.13936}, 2025.
\newblock URL \url{https://arxiv.org/abs/2504.13936}.

\bibitem[Madaan et~al.(2023)Madaan, Tandon, Gupta, Hallinan, Gao, Wiegreffe, Alon, Dziri, et~al.]{madaan2023selfrefine}
Aman Madaan, Niket Tandon, Prakhar Gupta, Skyler Hallinan, Luyu Gao, Sarah Wiegreffe, Uri Alon, Nouha Dziri, et~al.
\newblock {Self-Refine}: Iterative refinement with self-feedback.
\newblock In \emph{Adv. Neural Inf. Process. Syst.}, 2023.
\newblock URL \url{https://arxiv.org/abs/2303.17651}.

\bibitem[Maimon et~al.(2026)Maimon, Yoran, Kreuk, Hassid, Cohen, Chambon, and Adi]{maimon2026selfexecution}
Gallil Maimon, Ori Yoran, Felix Kreuk, Michael Hassid, Gal Cohen, Pierre Chambon, and Yossi Adi.
\newblock Self-execution simulation improves coding models.
\newblock \emph{arXiv preprint arXiv:2604.03253}, 2026.
\newblock URL \url{https://arxiv.org/abs/2604.03253}.

\bibitem[Makoviychuk et~al.(2021)Makoviychuk, Wawrzyniak, Guo, Lu, Storey, Macklin, Hoeller, Rudin, Allshire, Handa, and State]{makoviychuk2021isaacgym}
Viktor Makoviychuk, Lukasz Wawrzyniak, Yunrong Guo, Michelle Lu, Kier Storey, Miles Macklin, David Hoeller, Nikita Rudin, Arthur Allshire, Ankur Handa, and Gavriel State.
\newblock {Isaac Gym}: High performance {GPU}-based physics simulation for robot learning.
\newblock In \emph{Adv. Neural Inf. Process. Syst.}, 2021.
\newblock URL \url{https://arxiv.org/abs/2108.10470}.

\bibitem[McAleese et~al.(2024)McAleese, Pokorny, Uribe, Nitishinskaya, Trebacz, and Leike]{mcaleese2024criticgpt}
Nat McAleese, Rai~Michael Pokorny, Juan Felipe~Ceron Uribe, Evgenia Nitishinskaya, Maja Trebacz, and Jan Leike.
\newblock {LLM} critics help catch {LLM} bugs.
\newblock \emph{arXiv preprint arXiv:2407.00215}, 2024.
\newblock URL \url{https://arxiv.org/abs/2407.00215}.

\bibitem[Mei et~al.(2024)Mei, Hu, Yang, Wen, Yang, Wei, Ma, Dou, Shi, and Liu]{mei2024dreamforge}
Jianbiao Mei, Tao Hu, Xuemeng Yang, Licheng Wen, Yu~Yang, Tiantian Wei, Yukai Ma, Min Dou, Botian Shi, and Yong Liu.
\newblock Dreamforge: Motion-aware autoregressive video generation for multi-view driving scenes.
\newblock \emph{arXiv preprint arXiv:2409.04003}, 2024.
\newblock URL \url{https://arxiv.org/abs/2409.04003}.

\bibitem[Mei et~al.(2025)Mei, Yang, Yang, Wen, Lv, Shi, and Liu]{mei2025vision}
Jianbiao Mei, Yu~Yang, Xuemeng Yang, Licheng Wen, Jiajun Lv, Botian Shi, and Yong Liu.
\newblock Vision-centric {4D} occupancy forecasting and planning via implicit residual world models.
\newblock \emph{arXiv preprint arXiv:2510.16729}, 2025.
\newblock URL \url{https://arxiv.org/abs/2510.16729}.

\bibitem[Meng et~al.(2024)Meng, Xia, and Chen]{meng2024simpo}
Yu~Meng, Mengzhou Xia, and Danqi Chen.
\newblock {SimPO}: Simple preference optimization with a reference-free reward.
\newblock In \emph{Adv. Neural Inf. Process. Syst.}, 2024.
\newblock URL \url{https://arxiv.org/abs/2405.14734}.

\bibitem[Micheli et~al.(2023)Micheli, Alonso, and Fleuret]{micheli2023iris}
Vincent Micheli, Eloi Alonso, and Fran{\c{c}}ois Fleuret.
\newblock Transformers are sample-efficient world models.
\newblock In \emph{Int. Conf. Learn. Represent.}, 2023.
\newblock URL \url{https://arxiv.org/abs/2209.00588}.

\bibitem[Mildenhall et~al.(2020)Mildenhall, Srinivasan, Tancik, Barron, Ramamoorthi, and Ng]{mildenhall2020nerf}
Ben Mildenhall, Pratul~P. Srinivasan, Matthew Tancik, Jonathan~T. Barron, Ravi Ramamoorthi, and Ren Ng.
\newblock {NeRF}: Representing scenes as neural radiance fields for view synthesis.
\newblock In \emph{Eur. Conf. Comput. Vis.}, 2020.
\newblock URL \url{https://arxiv.org/abs/2003.08934}.

\bibitem[Min et~al.(2024)Min, Zhao, Xiao, Zhao, Xu, Zhu, Jin, Li, et~al.]{min2024driveworld}
Chen Min, Dawei Zhao, Liang Xiao, Jian Zhao, Xinli Xu, Zheng Zhu, Lei Jin, Jianshu Li, et~al.
\newblock {DriveWorld}: 4{D} pre-trained scene understanding via world models for autonomous driving.
\newblock In \emph{IEEE/CVF Conf. Comput. Vis. Pattern Recog.}, pages 15522--15533, 2024.
\newblock URL \url{https://arxiv.org/abs/2405.04390}.

\bibitem[Moerland et~al.(2023)Moerland, Broekens, Plaat, and Jonker]{moerland2023mbrl}
Thomas~M. Moerland, Joost Broekens, Aske Plaat, and Catholijn~M. Jonker.
\newblock Model-based reinforcement learning: A survey.
\newblock \emph{Found. Trends Mach. Learn.}, 16\penalty0 (1), 2023.
\newblock URL \url{https://arxiv.org/abs/2006.16712}.

\bibitem[M{\"u}ller et~al.(2022)M{\"u}ller, Evans, Schied, and Keller]{muller2022instantngp}
Thomas M{\"u}ller, Alex Evans, Christoph Schied, and Alexander Keller.
\newblock Instant neural graphics primitives with a multiresolution hash encoding.
\newblock \emph{ACM Trans. Graph.}, 41\penalty0 (4), 2022.
\newblock URL \url{https://arxiv.org/abs/2201.05989}.

\bibitem[Nasiriany et~al.(2024)Nasiriany, Maddukuri, Zhang, Parikh, Lo, Joshi, Mandlekar, and Zhu]{nasiriany2024robocasa}
Soroush Nasiriany, Abhiram Maddukuri, Lance Zhang, Adeet Parikh, Aaron Lo, Abhishek Joshi, Ajay Mandlekar, and Yuke Zhu.
\newblock {RoboCasa}: Large-scale simulation of everyday tasks for generalist robots.
\newblock In \emph{Robot. Sci. Syst.}, 2024.
\newblock URL \url{https://arxiv.org/abs/2406.02523}.

\bibitem[Ouyang et~al.(2022)Ouyang, Wu, Jiang, Almeida, Wainwright, Mishkin, Zhang, Agarwal, et~al.]{ouyang2022instructgpt}
Long Ouyang, Jeff Wu, Xu~Jiang, Diogo Almeida, Carroll~L. Wainwright, Pamela Mishkin, Chong Zhang, Sandhini Agarwal, et~al.
\newblock Training language models to follow instructions with human feedback.
\newblock In \emph{Adv. Neural Inf. Process. Syst.}, 2022.
\newblock URL \url{https://arxiv.org/abs/2203.02155}.

\bibitem[Packer et~al.(2023)Packer, Wooders, Lin, Fang, Patil, Stoica, and Gonzalez]{packer2023memgpt}
Charles Packer, Sarah Wooders, Kevin Lin, Vivian Fang, Shishir~G. Patil, Ion Stoica, and Joseph~E. Gonzalez.
\newblock {MemGPT}: Towards {LLMs} as operating systems.
\newblock \emph{arXiv preprint arXiv:2310.08560}, 2023.
\newblock URL \url{https://arxiv.org/abs/2310.08560}.

\bibitem[Park et~al.(2023)Park, O'Brien, Cai, Morris, Liang, and Bernstein]{park2023generative}
Joon~Sung Park, Joseph~C. O'Brien, Carrie~J. Cai, Meredith~Ringel Morris, Percy Liang, and Michael~S. Bernstein.
\newblock Generative agents: Interactive simulacra of human behavior.
\newblock In \emph{Annu. ACM Symp. User Interface Softw. Technol.}, pages 1--22, 2023.
\newblock URL \url{https://arxiv.org/abs/2304.03442}.

\bibitem[Qian et~al.(2026)Qian, Acikgoz, Li, Chen, Zhang, He, Luo, Hakkani-T{\"u}r, Tur, Li, and Ji]{qian2026foresight}
Cheng Qian, Emre~Can Acikgoz, Bingxuan Li, Xiusi Chen, Yuji Zhang, Bingxiang He, Qinyu Luo, Dilek Hakkani-T{\"u}r, Gokhan Tur, Yunzhu Li, and Heng Ji.
\newblock Current agents fail to leverage world model as tool for foresight.
\newblock \emph{arXiv preprint arXiv:2601.03905}, 2026.
\newblock URL \url{https://arxiv.org/abs/2601.03905}.

\bibitem[Qiao et~al.(2024)Qiao, Fang, Zhang, Zhu, Chen, Deng, Jiang, Xie, Huang, and Chen]{qiao2024wkm}
Shuofei Qiao, Runnan Fang, Ningyu Zhang, Yuqi Zhu, Xiang Chen, Shumin Deng, Yong Jiang, Pengjun Xie, Fei Huang, and Huajun Chen.
\newblock Agent planning with world knowledge model.
\newblock In \emph{Adv. Neural Inf. Process. Syst.}, volume~37, 2024.
\newblock URL \url{https://arxiv.org/abs/2405.14205}.

\bibitem[Qin et~al.(2024)Qin, Shi, Yu, Wang, Zhou, Li, Yin, Liu, et~al.]{qin2024worldsimbench}
Yiran Qin, Zhelun Shi, Jiwen Yu, Xijun Wang, Enshen Zhou, Lijun Li, Zhenfei Yin, Xihui Liu, et~al.
\newblock {WorldSimBench}: Towards video generation models as world simulators.
\newblock \emph{arXiv preprint arXiv:2410.18072}, 2024.
\newblock URL \url{https://arxiv.org/abs/2410.18072}.

\bibitem[Rafailov et~al.(2023)Rafailov, Sharma, Mitchell, Ermon, Manning, and Finn]{rafailov2023dpo}
Rafael Rafailov, Archit Sharma, Eric Mitchell, Stefano Ermon, Christopher~D. Manning, and Chelsea Finn.
\newblock Direct preference optimization: Your language model is secretly a reward model.
\newblock In \emph{Adv. Neural Inf. Process. Syst.}, 2023.
\newblock URL \url{https://arxiv.org/abs/2305.18290}.

\bibitem[Rahmani(2026)]{rahmani2026debuggingcwm}
Babak Rahmani.
\newblock Debugging code world models.
\newblock \emph{arXiv preprint arXiv:2602.07672}, 2026.
\newblock URL \url{https://arxiv.org/abs/2602.07672}.

\bibitem[Ren et~al.(2026)Ren, Yao, Sun, Qiao, Zhang, and Chen]{ren2026aligning}
Baochang Ren, Yunzhi Yao, Rui Sun, Shuofei Qiao, Ningyu Zhang, and Huajun Chen.
\newblock Aligning agentic world models via knowledgeable experience learning.
\newblock \emph{arXiv preprint arXiv:2601.13247}, 2026.
\newblock URL \url{https://arxiv.org/abs/2601.13247}.

\bibitem[Ren et~al.(2025)Ren, Zhang, Qian, Gao, Shi, Zheng, and He]{ren2025gtm}
Zhenzhen Ren, Xinpeng Zhang, Zhenxing Qian, Yan Gao, Yu~Shi, Shuxin Zheng, and Jiyan He.
\newblock {GTM}: Simulating the world of tools for {AI} agents.
\newblock \emph{arXiv preprint arXiv:2512.04535}, 2025.
\newblock URL \url{https://arxiv.org/abs/2512.04535}.

\bibitem[Rivard et~al.(2026)Rivard, Sun, Guo, Chen, and Deng]{rivard2025neuralos}
Luke Rivard, Sun Sun, Hongyu Guo, Wenhu Chen, and Yuntian Deng.
\newblock {NeuralOS}: Towards simulating operating systems via neural generative models.
\newblock In \emph{Int. Conf. Learn. Represent.}, 2026.
\newblock URL \url{https://arxiv.org/abs/2507.08800}.

\bibitem[Russell et~al.(2025)Russell, Hu, Bertoni, Fedoseev, Shotton, Arani, and Corrado]{nvidia2025gaia2}
Lloyd Russell, Anthony Hu, Lorenzo Bertoni, George Fedoseev, Jamie Shotton, Elahe Arani, and Gianluca Corrado.
\newblock {GAIA-2}: A controllable multi-view generative world model for autonomous driving.
\newblock \emph{arXiv preprint arXiv:2503.20523}, 2025.
\newblock URL \url{https://arxiv.org/abs/2503.20523}.

\bibitem[Savva et~al.(2019)Savva, Kadian, Maksymets, Zhao, Wijmans, Jain, Straub, Liu, Koltun, Malik, Parikh, and Batra]{savva2019habitat}
Manolis Savva, Abhishek Kadian, Oleksandr Maksymets, Yili Zhao, Erik Wijmans, Bhavana Jain, Julian Straub, Jia Liu, Vladlen Koltun, Jitendra Malik, Devi Parikh, and Dhruv Batra.
\newblock Habitat: A platform for embodied {AI} research.
\newblock In \emph{IEEE/CVF Int. Conf. Comput. Vis.}, pages 9339--9347, 2019.
\newblock URL \url{https://arxiv.org/abs/1904.01201}.

\bibitem[Schick et~al.(2023)Schick, Dwivedi-Yu, Dess{\`i}, Raileanu, Lomeli, Zettlemoyer, Cancedda, and Scialom]{schick2023toolformer}
Timo Schick, Jane Dwivedi-Yu, Roberto Dess{\`i}, Roberta Raileanu, Maria Lomeli, Luke Zettlemoyer, Nicola Cancedda, and Thomas Scialom.
\newblock Toolformer: Language models can teach themselves to use tools.
\newblock In \emph{Adv. Neural Inf. Process. Syst.}, 2023.
\newblock URL \url{https://arxiv.org/abs/2302.04761}.

\bibitem[Schrittwieser et~al.(2020)Schrittwieser, Antonoglou, Hubert, Simonyan, Sifre, Schmitt, Guez, Lockhart, Hassabis, Graepel, Lillicrap, and Silver]{schrittwieser2020muzero}
Julian Schrittwieser, Ioannis Antonoglou, Thomas Hubert, Karen Simonyan, Laurent Sifre, Simon Schmitt, Arthur Guez, Edward Lockhart, Demis Hassabis, Thore Graepel, Timothy Lillicrap, and David Silver.
\newblock Mastering {A}tari, {G}o, chess and shogi by planning with a learned model.
\newblock \emph{Nature}, 588\penalty0 (7839), 2020.
\newblock URL \url{https://arxiv.org/abs/1911.08265}.

\bibitem[Schulman et~al.(2017)Schulman, Wolski, Dhariwal, Radford, and Klimov]{schulman2017ppo}
John Schulman, Filip Wolski, Prafulla Dhariwal, Alec Radford, and Oleg Klimov.
\newblock Proximal policy optimization algorithms.
\newblock \emph{arXiv preprint arXiv:1707.06347}, 2017.
\newblock URL \url{https://arxiv.org/abs/1707.06347}.

\bibitem[Shao et~al.(2024)Shao, Wang, Zhu, Xu, Song, Bi, Zhang, Zhang, Li, Wu, and Guo]{shao2024deepseekmath}
Zhihong Shao, Peiyi Wang, Qihao Zhu, Runxin Xu, Junxiao Song, Xiao Bi, Haowei Zhang, Mingchuan Zhang, Y.~K. Li, Y.~Wu, and Daya Guo.
\newblock {DeepSeekMath}: Pushing the limits of mathematical reasoning in open language models.
\newblock \emph{arXiv preprint arXiv:2402.03300}, 2024.
\newblock URL \url{https://arxiv.org/abs/2402.03300}.

\bibitem[Shen et~al.(2026)Shen, Hu, Li, Fang, Li, and Zhang]{wac2026}
Zhouzhou Shen, Xueyu Hu, Xiyun Li, Tianqing Fang, Juncheng Li, and Shengyu Zhang.
\newblock World-model-augmented web agents with action correction.
\newblock \emph{arXiv preprint arXiv:2602.15384}, 2026.
\newblock URL \url{https://arxiv.org/abs/2602.15384}.

\bibitem[Shinn et~al.(2023)Shinn, Cassano, Berman, Gopinath, Narasimhan, and Yao]{shinn2023reflexion}
Noah Shinn, Federico Cassano, Edward Berman, Ashwin Gopinath, Karthik Narasimhan, and Shunyu Yao.
\newblock Reflexion: Language agents with verbal reinforcement learning.
\newblock In \emph{Adv. Neural Inf. Process. Syst.}, 2023.
\newblock URL \url{https://arxiv.org/abs/2303.11366}.

\bibitem[Shridhar et~al.(2021)Shridhar, Manuelli, and Fox]{shridhar2021cliport}
Mohit Shridhar, Lucas Manuelli, and Dieter Fox.
\newblock {CLIPort}: What and where pathways for robotic manipulation.
\newblock In \emph{Conf. Robot Learn.}, 2021.
\newblock URL \url{https://arxiv.org/abs/2109.12098}.

\bibitem[Shridhar et~al.(2022)Shridhar, Manuelli, and Fox]{shridhar2022peract}
Mohit Shridhar, Lucas Manuelli, and Dieter Fox.
\newblock Perceiver-actor: A multi-task transformer for robotic manipulation.
\newblock In \emph{Conf. Robot Learn.}, 2022.
\newblock URL \url{https://arxiv.org/abs/2209.05451}.

\bibitem[Stiennon et~al.(2020)Stiennon, Ouyang, Wu, Ziegler, Lowe, Voss, Radford, Amodei, and Christiano]{stiennon2020summarize}
Nisan Stiennon, Long Ouyang, Jeff Wu, Daniel~M. Ziegler, Ryan Lowe, Chelsea Voss, Alec Radford, Dario Amodei, and Paul Christiano.
\newblock Learning to summarize from human feedback.
\newblock In \emph{Adv. Neural Inf. Process. Syst.}, 2020.
\newblock URL \url{https://arxiv.org/abs/2009.01325}.

\bibitem[Sumers et~al.(2024)Sumers, Yao, Narasimhan, and Griffiths]{sumers2024coala}
Theodore~R. Sumers, Shunyu Yao, Karthik Narasimhan, and Thomas~L. Griffiths.
\newblock Cognitive architectures for language agents.
\newblock \emph{Trans. Mach. Learn. Res.}, 2024.
\newblock URL \url{https://arxiv.org/abs/2309.02427}.

\bibitem[Sun et~al.(2024)Sun, Chen, Liu, Chen, Duan, Zhang, and Wang]{sun2024dimensionx}
Wenqiang Sun, Shuo Chen, Fangfu Liu, Zilong Chen, Yueqi Duan, Jun Zhang, and Yikai Wang.
\newblock {DimensionX}: Create any {3D} and {4D} scenes from a single image with controllable video diffusion.
\newblock \emph{arXiv preprint arXiv:2411.04928}, 2024.
\newblock URL \url{https://arxiv.org/abs/2411.04928}.

\bibitem[Sutton(1991)]{sutton1991dyna}
Richard~S. Sutton.
\newblock Dyna, an integrated architecture for learning, planning, and reacting.
\newblock \emph{ACM SIGART Bull.}, 2\penalty0 (4):\penalty0 160--163, 1991.
\newblock URL \url{https://dl.acm.org/doi/10.1145/122344.122377}.

\bibitem[Tancik et~al.(2023)Tancik, Weber, Ng, Li, Yi, Kerr, Wang, Kristoffersen, et~al.]{tancik2023nerfstudio}
Matthew Tancik, Ethan Weber, Evonne Ng, Ruilong Li, Brent Yi, Justin Kerr, Terrance Wang, Alexander Kristoffersen, et~al.
\newblock Nerfstudio: A modular framework for neural radiance field development.
\newblock In \emph{Proc. ACM SIGGRAPH Asia Conf.}, 2023.
\newblock URL \url{https://arxiv.org/abs/2302.04264}.

\bibitem[Tang et~al.(2024)Tang, Key, and Ellis]{tang2024worldcoder}
Hao Tang, Darren Key, and Kevin Ellis.
\newblock {WorldCoder}, a model-based {LLM} agent: Building world models by writing code and interacting with the environment.
\newblock In \emph{Adv. Neural Inf. Process. Syst.}, volume~37, 2024.
\newblock URL \url{https://arxiv.org/abs/2402.12275}.

\bibitem[Tao et~al.(2024)Tao, Xiang, Shukla, Qin, Hinrichsen, Yuan, Bao, Lin, et~al.]{tao2024maniskill3}
Stone Tao, Fanbo Xiang, Arth Shukla, Yuzhe Qin, Xander Hinrichsen, Xiaodi Yuan, Chen Bao, Xinsong Lin, et~al.
\newblock {ManiSkill3}: {GPU} parallelized robotics simulation and rendering for generalizable embodied {AI}.
\newblock \emph{arXiv preprint arXiv:2410.00425}, 2024.
\newblock URL \url{https://arxiv.org/abs/2410.00425}.

\bibitem[Todorov et~al.(2012)Todorov, Erez, and Tassa]{todorov2012mujoco}
Emanuel Todorov, Tom Erez, and Yuval Tassa.
\newblock {MuJoCo}: A physics engine for model-based control.
\newblock In \emph{IEEE/RSJ Int. Conf. Intell. Robots Syst.}, pages 5026--5033, 2012.
\newblock URL \url{https://ieeexplore.ieee.org/document/6386109}.

\bibitem[Tu et~al.(2025)Tu, Zhou, Liang, Jiang, Zhang, Li, and Bai]{tu2025wm_ad_survey}
Sifan Tu, Xin Zhou, Dingkang Liang, Xingyu Jiang, Yumeng Zhang, Xiaofan Li, and Xiang Bai.
\newblock The role of world models in shaping autonomous driving: A comprehensive survey.
\newblock \emph{arXiv preprint arXiv:2502.10498}, 2025.
\newblock URL \url{https://arxiv.org/abs/2502.10498}.

\bibitem[Uesato et~al.(2022)Uesato, Kushman, Kumar, Song, Siegel, Wang, Creswell, Irving, and Higgins]{uesato2022process}
Jonathan Uesato, Nate Kushman, Ramana Kumar, Francis Song, Noah Siegel, Lisa Wang, Antonia Creswell, Geoffrey Irving, and Irina Higgins.
\newblock Solving math word problems with process- and outcome-based feedback.
\newblock \emph{arXiv preprint arXiv:2211.14275}, 2022.
\newblock URL \url{https://arxiv.org/abs/2211.14275}.

\bibitem[Vafa et~al.(2024)Vafa, Chen, Rambachan, Kleinberg, and Mullainathan]{vafa2024evaluating}
Keyon Vafa, Justin~Y. Chen, Ashesh Rambachan, Jon Kleinberg, and Sendhil Mullainathan.
\newblock Evaluating the world model implicit in a generative model.
\newblock In \emph{Adv. Neural Inf. Process. Syst.}, volume~37, 2024.
\newblock URL \url{https://arxiv.org/abs/2406.03689}.

\bibitem[Valevski et~al.(2025)Valevski, Leviathan, Arar, and Fruchter]{valevski2025gamengin}
Dani Valevski, Yaniv Leviathan, Moab Arar, and Shlomi Fruchter.
\newblock Diffusion models are real-time game engines.
\newblock In \emph{Int. Conf. Learn. Represent.}, 2025.
\newblock URL \url{https://arxiv.org/abs/2408.14837}.

\bibitem[van~de Ven et~al.(2025)van~de Ven, Soures, and Kudithipudi]{vandeven2024continuallearning}
Gido~M. van~de Ven, Nicholas Soures, and Dhireesha Kudithipudi.
\newblock Continual learning and catastrophic forgetting.
\newblock In \emph{Learning and Memory: A Comprehensive Reference}. Academic Press, 2025.
\newblock URL \url{https://arxiv.org/abs/2403.05175}.

\bibitem[Voleti et~al.(2022)Voleti, Jolicoeur-Martineau, and Pal]{voleti2022mcvd}
Vikram Voleti, Alexia Jolicoeur-Martineau, and Christopher Pal.
\newblock {MCVD}: Masked conditional video diffusion for prediction, generation, and interpolation.
\newblock In \emph{Adv. Neural Inf. Process. Syst.}, 2022.
\newblock URL \url{https://arxiv.org/abs/2205.09853}.

\bibitem[Wang et~al.(2024{\natexlab{a}})Wang, Xie, Jiang, Mandlekar, Xiao, Zhu, Fan, and Anandkumar]{wang2023voyager}
Guanzhi Wang, Yuqi Xie, Yunfan Jiang, Ajay Mandlekar, Chaowei Xiao, Yuke Zhu, Linxi Fan, and Anima Anandkumar.
\newblock Voyager: An open-ended embodied agent with large language models.
\newblock \emph{Trans. Mach. Learn. Res.}, 2024{\natexlab{a}}.
\newblock URL \url{https://arxiv.org/abs/2305.16291}.

\bibitem[Wang and Agapito(2025)]{wang2024spann3r}
Hengyi Wang and Lourdes Agapito.
\newblock {3D} reconstruction with spatial memory.
\newblock In \emph{Int. Conf. 3D Vis.}, 2025.
\newblock URL \url{https://arxiv.org/abs/2408.16061}.

\bibitem[Wang et~al.(2025{\natexlab{a}})Wang, Chen, Karaev, Vedaldi, Rupprecht, and Novotny]{wang2025vggt}
Jianyuan Wang, Minghao Chen, Nikita Karaev, Andrea Vedaldi, Christian Rupprecht, and David Novotny.
\newblock {VGGT}: Visual geometry grounded transformer.
\newblock \emph{arXiv preprint arXiv:2503.11651}, 2025{\natexlab{a}}.
\newblock URL \url{https://arxiv.org/abs/2503.11651}.

\bibitem[Wang et~al.(2026{\natexlab{a}})Wang, Jiang, He, Sun, Zhang, He, Cao, Gan, Sun, Shao, and Yue]{wang2026mvista4d}
Jiaxu Wang, Yicheng Jiang, Tianlun He, Jingkai Sun, Qiang Zhang, Junhao He, Jiahang Cao, Zesen Gan, Mingyuan Sun, Qiming Shao, and Xiangyu Yue.
\newblock {MVISTA-4D}: View-consistent {4D} world model with test-time action inference for robotic manipulation.
\newblock \emph{arXiv preprint arXiv:2602.09878}, 2026{\natexlab{a}}.
\newblock URL \url{https://arxiv.org/abs/2602.09878}.

\bibitem[Wang et~al.(2025{\natexlab{b}})Wang, Zhang, Wang, Gao, Li, Wang, Chen, Wan, Lu, Yang, et~al.]{wang2025vagen}
Kangrui Wang, Pingyue Zhang, Zihan Wang, Yaning Gao, Linjie Li, Qineng Wang, Hanyang Chen, Chi Wan, Yiping Lu, Zhengyuan Yang, et~al.
\newblock {VAGEN}: Reinforcing world model reasoning for multi-turn {VLM} agents.
\newblock In \emph{Adv. Neural Inf. Process. Syst.}, 2025{\natexlab{b}}.
\newblock URL \url{https://arxiv.org/abs/2510.16907}.

\bibitem[Wang et~al.(2024{\natexlab{b}})Wang, Li, Shao, Xu, Dai, Li, Chen, Wu, and Sui]{wang2023mathshepherd}
Peiyi Wang, Lei Li, Zhihong Shao, R.~X. Xu, Damai Dai, Yifei Li, Deli Chen, Y.~Wu, and Zhifang Sui.
\newblock {Math-Shepherd}: Verify and reinforce {LLMs} step-by-step without human annotations.
\newblock In \emph{Annu. Meet. Assoc. Comput. Linguist.}, 2024{\natexlab{b}}.
\newblock URL \url{https://arxiv.org/abs/2312.08935}.

\bibitem[Wang et~al.(2025{\natexlab{c}})Wang, Zhang, Holynski, Efros, and Kanazawa]{wang2025cut3r}
Qianqian Wang, Yifei Zhang, Aleksander Holynski, Alexei~A. Efros, and Angjoo Kanazawa.
\newblock Continuous {3D} perception model with persistent state.
\newblock \emph{arXiv preprint arXiv:2501.12387}, 2025{\natexlab{c}}.
\newblock URL \url{https://arxiv.org/abs/2501.12387}.

\bibitem[Wang et~al.(2025{\natexlab{d}})Wang, Yin, Zhang, Zhang, Wang, Wang, Zhang, Chandrasegaran, et~al.]{mindcube}
Qineng Wang, Baiqiao Yin, Pingyue Zhang, Jianshu Zhang, Kangrui Wang, Zihan Wang, Jieyu Zhang, Keshigeyan Chandrasegaran, et~al.
\newblock {MindCube}: Spatial mental modeling from limited views.
\newblock \emph{arXiv preprint arXiv:2506.21458}, 2025{\natexlab{d}}.
\newblock URL \url{https://arxiv.org/abs/2506.21458}.

\bibitem[Wang et~al.(2024{\natexlab{c}})Wang, Leroy, Cabon, Chidlovskii, and Revaud]{wang2023dust3r}
Shuzhe Wang, Vincent Leroy, Yohann Cabon, Boris Chidlovskii, and Jerome Revaud.
\newblock {DUSt3R}: Geometric {3D} vision made easy.
\newblock In \emph{IEEE/CVF Conf. Comput. Vis. Pattern Recog.}, 2024{\natexlab{c}}.
\newblock URL \url{https://arxiv.org/abs/2312.14132}.

\bibitem[Wang et~al.(2024{\natexlab{d}})Wang, Zhu, Huang, Chen, Zhu, and Lu]{wang2024drivedreamer}
Xiaofeng Wang, Zheng Zhu, Guan Huang, Xinze Chen, Jiagang Zhu, and Jiwen Lu.
\newblock {DriveDreamer}: Towards real-world-driven world models for autonomous driving.
\newblock In \emph{Eur. Conf. Comput. Vis.}, 2024{\natexlab{d}}.
\newblock URL \url{https://arxiv.org/abs/2309.09777}.

\bibitem[Wang et~al.(2026{\natexlab{b}})Wang, Xu, Liu, Wang, Han, Yao, Yao, and He]{wang2026awm}
Zhaoyang Wang, Canwen Xu, Boyi Liu, Yite Wang, Siwei Han, Zhewei Yao, Huaxiu Yao, and Yuxiong He.
\newblock Agent world model: Infinity synthetic environments for agentic reinforcement learning.
\newblock In \emph{Int. Conf. Mach. Learn.}, 2026{\natexlab{b}}.
\newblock URL \url{https://arxiv.org/abs/2602.10090}.

\bibitem[Wang et~al.(2025{\natexlab{e}})Wang, Wang, Wang, Zhang, Li, Yang, Jin, Yu, Nguyen, Liu, et~al.]{wang2025ragen}
Zihan Wang, Kangrui Wang, Qineng Wang, Pingyue Zhang, Linjie Li, Zhengyuan Yang, Xing Jin, Kefan Yu, Minh~Nhat Nguyen, Licheng Liu, et~al.
\newblock {RAGEN}: Understanding self-evolution in {LLM} agents via multi-turn reinforcement learning.
\newblock \emph{arXiv preprint arXiv:2504.20073}, 2025{\natexlab{e}}.
\newblock URL \url{https://arxiv.org/abs/2504.20073}.

\bibitem[Wei et~al.(2022)Wei, Wang, Schuurmans, Bosma, Ichter, Xia, Chi, Le, and Zhou]{wei2022cot}
Jason Wei, Xuezhi Wang, Dale Schuurmans, Maarten Bosma, Brian Ichter, Fei Xia, Ed~Chi, Quoc~V. Le, and Denny Zhou.
\newblock Chain-of-thought prompting elicits reasoning in large language models.
\newblock In \emph{Adv. Neural Inf. Process. Syst.}, volume~35, pages 24824--24837, 2022.
\newblock URL \url{https://arxiv.org/abs/2201.11903}.

\bibitem[{World Labs Team}(2025)]{worldlabs2025marble}
{World Labs Team}.
\newblock Marble: A multimodal world model.
\newblock World Labs Technical Post, 2025.
\newblock URL \url{https://www.worldlabs.ai/blog/marble-world-model}.

\bibitem[Wu et~al.(2024{\natexlab{a}})Wu, Yi, Fang, Xie, Zhang, Wei, Liu, Tian, and Wang]{wu20244dgs}
Guanjun Wu, Taoran Yi, Jiemin Fang, Lingxi Xie, Xiaopeng Zhang, Wei Wei, Wenyu Liu, Qi~Tian, and Xinggang Wang.
\newblock {4D} gaussian splatting for real-time dynamic scene rendering.
\newblock In \emph{IEEE/CVF Conf. Comput. Vis. Pattern Recog.}, 2024{\natexlab{a}}.
\newblock URL \url{https://arxiv.org/abs/2310.08528}.

\bibitem[Wu et~al.(2024{\natexlab{b}})Wu, Yin, Feng, He, Li, Hao, and Long]{wu2024ivideogpt}
Jialong Wu, Shaofeng Yin, Ningya Feng, Xu~He, Dong Li, Jianye Hao, and Mingsheng Long.
\newblock {iVideoGPT}: Interactive {VideoGPTs} are scalable world models.
\newblock In \emph{Adv. Neural Inf. Process. Syst.}, volume~37, pages 68082--68119, 2024{\natexlab{b}}.
\newblock URL \url{https://arxiv.org/abs/2405.15223}.

\bibitem[Wu et~al.(2023)Wu, Escontrela, Hafner, Goldberg, and Abbeel]{wu2023daydreamer}
Philipp Wu, Alejandro Escontrela, Danijar Hafner, Ken Goldberg, and Pieter Abbeel.
\newblock {DayDreamer}: World models for physical robot learning.
\newblock In \emph{Conf. Robot Learn.}, 2023.
\newblock URL \url{https://arxiv.org/abs/2206.14176}.

\bibitem[Wu et~al.(2026)Wu, Fu, Wen, Yang, Zou, Mei, Wang, Zhang, Yang, Hu, et~al.]{wu2026memharness}
Rong Wu, Daocheng Fu, Licheng Wen, Xuemeng Yang, Shu Zou, Jianbiao Mei, Yuxin Wang, Hairong Zhang, Yu~Yang, Tao Hu, et~al.
\newblock Memharness: Memory is reconstructed, not replayed.
\newblock \emph{arXiv preprint arXiv:2607.28272}, 2026.
\newblock URL \url{https://arxiv.org/abs/2607.28272}.

\bibitem[Xiao et~al.(2026)Xiao, Tu, Zou, Zuo, Li, Wang, Yu, Huang, Lin, and Liu]{xiao2026webworld}
Zikai Xiao, Jianhong Tu, Chuhang Zou, Yuxin Zuo, Zhi Li, Peng Wang, Bowen Yu, Fei Huang, Junyang Lin, and Zuozhu Liu.
\newblock {WebWorld}: A large-scale world model for web agent training.
\newblock \emph{arXiv preprint arXiv:2602.14721}, 2026.
\newblock URL \url{https://arxiv.org/abs/2602.14721}.

\bibitem[Xie et~al.(2024{\natexlab{a}})Xie, Chen, Hong, and Liu]{xie2023citydreamer}
Haozhe Xie, Zhaoxi Chen, Fangzhou Hong, and Ziwei Liu.
\newblock {CityDreamer}: Compositional generative model of unbounded {3D} cities.
\newblock In \emph{IEEE/CVF Conf. Comput. Vis. Pattern Recog.}, 2024{\natexlab{a}}.
\newblock URL \url{https://arxiv.org/abs/2309.00610}.

\bibitem[Xie et~al.(2024{\natexlab{b}})Xie, Zhang, Chen, Li, Zhao, Cao, Hua, Cheng, et~al.]{xie2024osworld}
Tianbao Xie, Danyang Zhang, Jixuan Chen, Xiaochuan Li, Siheng Zhao, Ruisheng Cao, Toh~Jing Hua, Zhoujun Cheng, et~al.
\newblock {OSWorld}: Benchmarking multimodal agents for open-ended tasks in real computer environments.
\newblock In \emph{Adv. Neural Inf. Process. Syst.}, volume~37, 2024{\natexlab{b}}.
\newblock URL \url{https://arxiv.org/abs/2404.07972}.

\bibitem[Xu et~al.(2026{\natexlab{a}})Xu, Liang, Zheng, Luo, Hu, Zhang, and Tao]{xu2026ctrlattack}
Shuhan Xu, Siyuan Liang, Hongling Zheng, Yong Luo, Han Hu, Lefei Zhang, and Dacheng Tao.
\newblock {CtrlAttack}: A unified attack on world-model control in diffusion models.
\newblock \emph{arXiv preprint arXiv:2603.13435}, 2026{\natexlab{a}}.
\newblock URL \url{https://arxiv.org/abs/2603.13435}.

\bibitem[Xu et~al.(2026{\natexlab{b}})Xu, Liang, Liu, Li, Kong, Liu, and Liu]{xu2025u4d}
Xiang Xu, Ao~Liang, Youquan Liu, Linfeng Li, Lingdong Kong, Ziwei Liu, and Qingshan Liu.
\newblock {U4D}: Uncertainty-aware {4D} world modeling from {LiDAR} sequences.
\newblock In \emph{IEEE/CVF Conf. Comput. Vis. Pattern Recog.}, pages 10027--10039, 2026{\natexlab{b}}.
\newblock URL \url{https://arxiv.org/abs/2512.02982}.

\bibitem[Yan et~al.(2026{\natexlab{a}})Yan, Lin, Zhu, and Wang]{yan2026safedream}
Bo~Yan, Weikai Lin, Yada Zhu, and Song Wang.
\newblock {SafeDream}: Safety world model for proactive early jailbreak detection.
\newblock \emph{arXiv preprint arXiv:2604.16824}, 2026{\natexlab{a}}.
\newblock URL \url{https://arxiv.org/abs/2604.16824}.

\bibitem[Yan et~al.(2026{\natexlab{b}})Yan, Tang, Gui, Li, Zheng, Huang, Kong, Han, et~al.]{yan2025ad-r1}
Tianyi Yan, Tao Tang, Xingtai Gui, Yongkang Li, Jiasen Zheng, Weiyao Huang, Lingdong Kong, Wencheng Han, et~al.
\newblock {AD-R1}: Closed-loop reinforcement learning for end-to-end autonomous driving with impartial world models.
\newblock In \emph{IEEE/CVF Conf. Comput. Vis. Pattern Recog.}, pages 1085--1095, 2026{\natexlab{b}}.
\newblock URL \url{https://arxiv.org/abs/2511.20325}.

\bibitem[Yan et~al.(2021)Yan, Zhang, Abbeel, and Srinivas]{yan2021videogpt}
Wilson Yan, Yunzhi Zhang, Pieter Abbeel, and Aravind Srinivas.
\newblock {VideoGPT}: Video generation using {VQ-VAE} and transformers.
\newblock \emph{arXiv preprint arXiv:2104.10157}, 2021.
\newblock URL \url{https://arxiv.org/abs/2104.10157}.

\bibitem[Yang et~al.(2025{\natexlab{a}})Yang, Sax, Liang, Henaff, Tang, Cao, Chai, Meier, and Feiszli]{yang2025fast3r}
Jianing Yang, Alexander Sax, Kevin~J. Liang, Mikael Henaff, Hao Tang, Ang Cao, Joyce Chai, Franziska Meier, and Matt Feiszli.
\newblock {Fast3R}: Towards {3D} reconstruction of 1000+ images in one forward pass.
\newblock \emph{arXiv preprint arXiv:2501.13928}, 2025{\natexlab{a}}.
\newblock URL \url{https://arxiv.org/abs/2501.13928}.

\bibitem[Yang et~al.(2024)Yang, Du, Ghasemipour, Tompson, Kaelbling, Schuurmans, and Abbeel]{yang2024unisim}
Sherry Yang, Yilun Du, Seyed Kamyar~Seyed Ghasemipour, Jonathan Tompson, Leslie~Pack Kaelbling, Dale Schuurmans, and Pieter Abbeel.
\newblock Learning interactive real-world simulators.
\newblock In \emph{Int. Conf. Learn. Represent.}, 2024.
\newblock URL \url{https://arxiv.org/abs/2310.06114}.

\bibitem[Yang et~al.(2025{\natexlab{b}})Yang, Mei, Ma, Du, Chen, Qian, Feng, and Liu]{yang2025driving}
Yu~Yang, Jianbiao Mei, Yukai Ma, Siliang Du, Wenqing Chen, Yijie Qian, Yuxiang Feng, and Yong Liu.
\newblock Driving in the occupancy world: Vision-centric 4d occupancy forecasting and planning via world models for autonomous driving.
\newblock In \emph{AAAI Conf. Artif. Intell.}, volume~39, pages 9327--9335, 2025{\natexlab{b}}.
\newblock URL \url{https://arxiv.org/abs/2408.14197}.

\bibitem[Yang et~al.(2026{\natexlab{a}})Yang, Liang, Mei, Ma, Liu, and Lee]{yang2026x}
Yu~Yang, Alan Liang, Jianbiao Mei, Yukai Ma, Yong Liu, and Gim~Hee Lee.
\newblock X-scene: Large-scale driving scene generation with high fidelity and flexible controllability.
\newblock In \emph{Adv. Neural Inf. Process. Syst.}, volume~38, pages 104415--104451, 2026{\natexlab{a}}.
\newblock URL \url{https://arxiv.org/abs/2506.13558}.

\bibitem[Yang et~al.(2026{\natexlab{b}})Yang, Liao, Mei, Wang, Yang, Wen, Zhang, Li, Lv, Chen, et~al.]{yang2026spiral}
Yu~Yang, Yue Liao, Jianbiao Mei, Baisen Wang, Xuemeng Yang, Licheng Wen, Jiangning Zhang, Xiangtai Li, Liang Lv, Hanlin Chen, et~al.
\newblock {SPIRAL}: Self-evolving action-conditioned video generation via reflective planning agents.
\newblock \emph{arXiv preprint arXiv:2603.08403}, 2026{\natexlab{b}}.
\newblock URL \url{https://arxiv.org/abs/2603.08403}.

\bibitem[Yao et~al.(2023{\natexlab{a}})Yao, Yu, Zhao, Shafran, Griffiths, Cao, and Narasimhan]{yao2023tot}
Shunyu Yao, Dian Yu, Jeffrey Zhao, Izhak Shafran, Thomas~L. Griffiths, Yuan Cao, and Karthik Narasimhan.
\newblock Tree of thoughts: Deliberate problem solving with large language models.
\newblock In \emph{Adv. Neural Inf. Process. Syst.}, volume~36, 2023{\natexlab{a}}.
\newblock URL \url{https://arxiv.org/abs/2305.10601}.

\bibitem[Yao et~al.(2023{\natexlab{b}})Yao, Zhao, Yu, Du, Shafran, Narasimhan, and Cao]{yao2022react}
Shunyu Yao, Jeffrey Zhao, Dian Yu, Nan Du, Izhak Shafran, Karthik Narasimhan, and Yuan Cao.
\newblock {ReAct}: Synergizing reasoning and acting in language models.
\newblock In \emph{Int. Conf. Learn. Represent.}, 2023{\natexlab{b}}.
\newblock URL \url{https://arxiv.org/abs/2210.03629}.

\bibitem[Yu et~al.(2022)Yu, Fridovich-Keil, Tancik, Chen, Recht, and Kanazawa]{yu2021plenoxels}
Alex Yu, Sara Fridovich-Keil, Matthew Tancik, Qinhong Chen, Benjamin Recht, and Angjoo Kanazawa.
\newblock Plenoxels: Radiance fields without neural networks.
\newblock In \emph{IEEE/CVF Conf. Comput. Vis. Pattern Recog.}, pages 5501--5510, 2022.
\newblock URL \url{https://arxiv.org/abs/2112.05131}.

\bibitem[Yu et~al.(2025{\natexlab{a}})Yu, Duan, Herrmann, Freeman, and Wu]{yu2024wonderworld}
Hong-Xing Yu, Haoyi Duan, Charles Herrmann, William~T. Freeman, and Jiajun Wu.
\newblock {WonderWorld}: Interactive {3D} scene generation from a single image.
\newblock In \emph{IEEE/CVF Conf. Comput. Vis. Pattern Recog.}, 2025{\natexlab{a}}.
\newblock URL \url{https://arxiv.org/abs/2406.09394}.

\bibitem[Yu et~al.(2025{\natexlab{b}})Yu, Qin, Wang, Wan, Zhang, and Liu]{yu2025gamefactory}
Jiwen Yu, Yiran Qin, Xintao Wang, Pengfei Wan, Di~Zhang, and Xihui Liu.
\newblock {GameFactory}: Creating new games with generative interactive videos.
\newblock In \emph{IEEE/CVF Int. Conf. Comput. Vis.}, pages 11590--11599, 2025{\natexlab{b}}.
\newblock URL \url{https://arxiv.org/abs/2501.08325}.

\bibitem[Yu et~al.(2025{\natexlab{c}})Yu, Xu, Zhang, Chen, Zhang, He, Jiang, Zhang, Hu, and Yan]{yu2025vismem}
Xinlei Yu, Chengming Xu, Guibin Zhang, Zhangquan Chen, Yudong Zhang, Yongbo He, Peng-Tao Jiang, Jiangning Zhang, Xiaobin Hu, and Shuicheng Yan.
\newblock Vismem: Latent vision memory unlocks potential of vision-language models.
\newblock \emph{arXiv preprint arXiv:2511.11007}, 2025{\natexlab{c}}.
\newblock URL \url{https://arxiv.org/abs/2511.11007}.

\bibitem[Yu et~al.(2026)Yu, Chen, He, Fu, Dong, Yang, Xu, Ma, Hu, Cao, et~al.]{yu2026latent}
Xinlei Yu, Zhangquan Chen, Yongbo He, Tianyu Fu, Guanting Dong, Cheng Yang, Chengming Xu, Yue Ma, Xiaobin Hu, Zhe Cao, et~al.
\newblock The latent space: Foundation, evolution, mechanism, ability, and outlook.
\newblock \emph{arXiv preprint arXiv:2604.02029}, 2026.
\newblock URL \url{https://arxiv.org/abs/2604.02029}.

\bibitem[Yu et~al.(2024)Yu, Chen, Huang, Sattler, and Geiger]{yu2023mipsplatting}
Zehao Yu, Anpei Chen, Binbin Huang, Torsten Sattler, and Andreas Geiger.
\newblock {Mip-Splatting}: Alias-free {3D} gaussian splatting.
\newblock In \emph{IEEE/CVF Conf. Comput. Vis. Pattern Recog.}, pages 19447--19456, 2024.
\newblock URL \url{https://arxiv.org/abs/2311.16493}.

\bibitem[Yuan et~al.(2023)Yuan, Yuan, Tan, Wang, Huang, and Huang]{yuan2023rrhf}
Zheng Yuan, Hongyi Yuan, Chuanqi Tan, Wei Wang, Songfang Huang, and Fei Huang.
\newblock {RRHF}: Rank responses to align language models with human feedback without tears.
\newblock In \emph{Adv. Neural Inf. Process. Syst.}, 2023.
\newblock URL \url{https://arxiv.org/abs/2304.05302}.

\bibitem[Zeng et~al.(2020)Zeng, Florence, Tompson, Welker, Chien, Attarian, Armstrong, Krasin, Duong, Wahid, Sindhwani, and Lee]{zeng2020transporternets}
Andy Zeng, Pete Florence, Jonathan Tompson, Stefan Welker, Jonathan Chien, Maria Attarian, Travis Armstrong, Ivan Krasin, Dan Duong, Ayzaan Wahid, Vikas Sindhwani, and Johnny Lee.
\newblock Transporter networks: Rearranging the visual world for robotic manipulation.
\newblock In \emph{Conf. Robot Learn.}, 2020.
\newblock URL \url{https://arxiv.org/abs/2010.14406}.

\bibitem[Zeng et~al.(2024)Zeng, Zhang, Liu, Sifakis, Zhang, Liu, and Wang]{zeng2024wmsafety}
Zifan Zeng, Chongzhe Zhang, Feng Liu, Joseph Sifakis, Qunli Zhang, Shiming Liu, and Peng Wang.
\newblock World models: The safety perspective.
\newblock \emph{arXiv preprint arXiv:2411.07690}, 2024.
\newblock URL \url{https://arxiv.org/abs/2411.07690}.

\bibitem[Zha et~al.(2025)Zha, Fan, Yang, Gao, and Chen]{zha2025enable}
Jirong Zha, Yuxuan Fan, Xiao Yang, Chen Gao, and Xinlei Chen.
\newblock How to enable {LLM} with {3D} capacity? a survey of spatial reasoning in {LLM}.
\newblock In \emph{Int. Joint Conf. Artif. Intell.}, pages 10817--10825, 2025.
\newblock URL \url{https://arxiv.org/abs/2504.05786}.

\bibitem[Zhang et~al.(2025{\natexlab{a}})Zhang, Li, Lei, Wang, Liu, Yang, Li, Wang, et~al.]{zhang2025critic}
Di~Zhang, Junxian Li, Jingdi Lei, Xunzhi Wang, Yujie Liu, Zonglin Yang, Jiatong Li, Weida Wang, et~al.
\newblock {Critic-V}: {VLM} critics help catch {VLM} errors in multimodal reasoning.
\newblock In \emph{IEEE/CVF Conf. Comput. Vis. Pattern Recog.}, pages 9050--9061, 2025{\natexlab{a}}.
\newblock URL \url{https://arxiv.org/abs/2411.18203}.

\bibitem[Zhang et~al.(2024{\natexlab{a}})Zhang, Li, Wan, Wang, and Liao]{zhang2023text2nerf}
Jingbo Zhang, Xiaoyu Li, Ziyu Wan, Can Wang, and Jing Liao.
\newblock {Text2NeRF}: Text-driven {3D} scene generation with neural radiance fields.
\newblock \emph{IEEE Trans. Vis. Comput. Graph.}, 2024{\natexlab{a}}.
\newblock URL \url{https://arxiv.org/abs/2305.11588}.

\bibitem[Zhang et~al.(2025{\natexlab{b}})Zhang, Herrmann, Hur, Jampani, Darrell, Cole, Sun, and Yang]{zhang2024monst3r}
Junyi Zhang, Charles Herrmann, Junhwa Hur, Varun Jampani, Trevor Darrell, Forrester Cole, Deqing Sun, and Ming-Hsuan Yang.
\newblock {MonST3R}: A simple approach for estimating geometry in the presence of motion.
\newblock In \emph{Int. Conf. Learn. Represent.}, 2025{\natexlab{b}}.
\newblock URL \url{https://arxiv.org/abs/2410.03825}.

\bibitem[Zhang et~al.(2026{\natexlab{a}})Zhang, Chen, Liu, Xue, Liao, Liu, Wang, Ning, et~al.]{zhang2025earlyexp}
Kai Zhang, Xiangchao Chen, Bo~Liu, Tianci Xue, Zeyi Liao, Zhihan Liu, Xiyao Wang, Yuting Ning, et~al.
\newblock Agent learning via early experience.
\newblock In \emph{Int. Conf. Mach. Learn.}, 2026{\natexlab{a}}.
\newblock URL \url{https://arxiv.org/abs/2510.08558}.

\bibitem[Zhang et~al.(2024{\natexlab{b}})Zhang, Xiong, Yang, Casas, Hu, and Urtasun]{zhang2024copilot4d}
Lunjun Zhang, Yuwen Xiong, Ze~Yang, Sergio Casas, Rui Hu, and Raquel Urtasun.
\newblock {Copilot4D}: Learning unsupervised world models for autonomous driving via discrete diffusion.
\newblock In \emph{Int. Conf. Learn. Represent.}, 2024{\natexlab{b}}.
\newblock URL \url{https://arxiv.org/abs/2311.01017}.

\bibitem[Zhang et~al.(2023)Zhang, Wang, Sun, Yuan, and Huang]{zhang2023storm}
Weipu Zhang, Gang Wang, Jian Sun, Yetian Yuan, and Gao Huang.
\newblock {STORM}: Efficient stochastic transformer based world models for reinforcement learning.
\newblock In \emph{Adv. Neural Inf. Process. Syst.}, volume~36, 2023.
\newblock URL \url{https://arxiv.org/abs/2310.09615}.

\bibitem[Zhang et~al.(2026{\natexlab{b}})Zhang, He, Zhu, Wu, Yu, Chu, Zhang, Tan, and Jia]{zhang2026searchgym}
Xichen Zhang, Ziyi He, Yinghao Zhu, Sitong Wu, Shaozuo Yu, Meng Chu, Wenhu Zhang, Haoru Tan, and Jiaya Jia.
\newblock {SearchGym}: Bootstrapping real-world search agents via cost-effective and high-fidelity environment simulation.
\newblock \emph{arXiv preprint arXiv:2601.14615}, 2026{\natexlab{b}}.
\newblock URL \url{https://arxiv.org/abs/2601.14615}.

\bibitem[Zhang et~al.(2025{\natexlab{c}})Zhang, Peng, Wang, Wang, Zhu, Kang, Jiang, Gao, Li, Liu, and Zhou]{zhang2025matrixgame}
Yifan Zhang, Chunli Peng, Boyang Wang, Puyi Wang, Qingcheng Zhu, Fei Kang, Biao Jiang, Zedong Gao, Eric Li, Yang Liu, and Yahui Zhou.
\newblock {Matrix-Game}: Interactive world foundation model.
\newblock \emph{arXiv preprint arXiv:2506.18701}, 2025{\natexlab{c}}.
\newblock URL \url{https://arxiv.org/abs/2506.18701}.

\bibitem[Zhang et~al.(2025{\natexlab{d}})Zhang, Zhang, Cui, Shi, Guo, Han, Zhao, Sun, et~al.]{zhang2025robooccworld}
Zhang Zhang, Qiang Zhang, Wei Cui, Shuai Shi, Yijie Guo, Gang Han, Wen Zhao, Jingkai Sun, et~al.
\newblock Occupancy world model for robots.
\newblock \emph{arXiv preprint arXiv:2505.05512}, 2025{\natexlab{d}}.
\newblock URL \url{https://arxiv.org/abs/2505.05512}.

\bibitem[Zhen et~al.(2025)Zhen, Sun, Zhang, Li, Zhou, Du, and Gan]{zhen2025tesseract}
Haoyu Zhen, Qiao Sun, Hongxin Zhang, Junyan Li, Siyuan Zhou, Yilun Du, and Chuang Gan.
\newblock {TesserAct}: Learning {4D} embodied world models.
\newblock \emph{arXiv preprint arXiv:2504.20995}, 2025.
\newblock URL \url{https://arxiv.org/abs/2504.20995}.

\bibitem[Zheng et~al.(2023)Zheng, Chiang, Sheng, Zhuang, Wu, Zhuang, Lin, Li, et~al.]{zheng2023judge}
Lianmin Zheng, Wei-Lin Chiang, Ying Sheng, Siyuan Zhuang, Zhanghao Wu, Yonghao Zhuang, Zi~Lin, Zhuohan Li, et~al.
\newblock Judging {LLM}-as-a-judge with {MT-Bench} and chatbot arena.
\newblock In \emph{Adv. Neural Inf. Process. Syst.}, 2023.
\newblock URL \url{https://arxiv.org/abs/2306.05685}.

\bibitem[Zheng et~al.(2024)Zheng, Chen, Huang, Zhang, Duan, and Lu]{zheng2024occworld}
Wenzhao Zheng, Weiliang Chen, Yuanhui Huang, Borui Zhang, Yueqi Duan, and Jiwen Lu.
\newblock {OccWorld}: Learning a 3{D} occupancy world model for autonomous driving.
\newblock In \emph{Eur. Conf. Comput. Vis.}, 2024.
\newblock URL \url{https://arxiv.org/abs/2311.16038}.

\bibitem[Zheng et~al.(2026)Zheng, Zhong, Wang, Dai, Liu, Chu, Lv, Torr, and Lin]{zheng2026code2world}
Yuhao Zheng, Li'an Zhong, Yi~Wang, Rui Dai, Kaikui Liu, Xiangxiang Chu, Linyuan Lv, Philip Torr, and Kevin~Qinghong Lin.
\newblock {Code2World}: A {GUI} world model via renderable code generation.
\newblock \emph{arXiv preprint arXiv:2602.09856}, 2026.
\newblock URL \url{https://arxiv.org/abs/2602.09856}.

\bibitem[Zhou et~al.(2024)Zhou, Xu, Zhu, Zhou, Lo, Sridhar, Cheng, Ou, Bisk, Fried, Alon, and Neubig]{zhou2023webarena}
Shuyan Zhou, Frank~F. Xu, Hao Zhu, Xuhui Zhou, Robert Lo, Abishek Sridhar, Xianyi Cheng, Tianyue Ou, Yonatan Bisk, Daniel Fried, Uri Alon, and Graham Neubig.
\newblock {WebArena}: A realistic web environment for building autonomous agents.
\newblock In \emph{Int. Conf. Learn. Represent.}, 2024.
\newblock URL \url{https://arxiv.org/abs/2307.13854}.

\bibitem[Zhu et~al.(2025)Zhu, Wang, Zhou, Chang, Zhou, Li, Chen, Shen, Pang, and He]{zhu2025aether}
Haoyi Zhu, Yifan Wang, Jianjun Zhou, Wenzheng Chang, Yang Zhou, Zizun Li, Junyi Chen, Chunhua Shen, Jiangmiao Pang, and Tong He.
\newblock Aether: Geometric-aware unified world modeling.
\newblock In \emph{IEEE/CVF Int. Conf. Comput. Vis.}, pages 8535--8546, 2025.
\newblock URL \url{https://arxiv.org/abs/2503.18945}.

\bibitem[Zhu et~al.(2024)Zhu, Wang, Zhao, Min, Li, Deng, Dou, Wang, et~al.]{zhu2024sora_survey}
Zheng Zhu, Xiaofeng Wang, Wangbo Zhao, Chen Min, Bohan Li, Nianchen Deng, Min Dou, Yuqi Wang, et~al.
\newblock Is {Sora} a world simulator? a comprehensive survey on general world models and beyond.
\newblock \emph{arXiv preprint arXiv:2405.03520}, 2024.
\newblock URL \url{https://arxiv.org/abs/2405.03520}.

\bibitem[Ziegler et~al.(2019)Ziegler, Stiennon, Wu, Brown, Radford, Amodei, Christiano, and Irving]{ziegler2019finetuning}
Daniel~M. Ziegler, Nisan Stiennon, Jeffrey Wu, Tom~B. Brown, Alec Radford, Dario Amodei, Paul Christiano, and Geoffrey Irving.
\newblock Fine-tuning language models from human preferences.
\newblock \emph{arXiv preprint arXiv:1909.08593}, 2019.
\newblock URL \url{https://arxiv.org/abs/1909.08593}.

\end{thebibliography}
